\documentclass[reprint,
                superscriptaddress,
                aps,
                prd,
                twocolumn,
                english,
                floatfix,
                longbibliography]{revtex4-2}

\usepackage[utf8]{inputenc} 
\usepackage{amsmath,amssymb}
\usepackage{cancel}
\usepackage{graphicx}
\usepackage{dcolumn}
\usepackage{bm}
\usepackage{xspace}
\usepackage{siunitx}
\usepackage{multirow}
\usepackage{xcolor}
\usepackage{booktabs}       
\usepackage{amsfonts}       
\usepackage{nicefrac}       
\definecolor{darkblue}{RGB}{46,48,147}
\usepackage[colorlinks=true,
            linkcolor=darkblue,
            urlcolor=darkblue,
            citecolor=darkblue]{hyperref}
\usepackage[capitalise]{cleveref}
\usepackage{acronym}
\usepackage{subcaption}
\usepackage{algorithm}
\usepackage{algpseudocode}

\DeclareMathOperator{\EX}{\mathbb{E}}

\newcommand{\GeV}{\ensuremath{\,\text{Ge\hspace{-.08em}V}}\xspace}
\newcommand{\TeV}{\ensuremath{\,\text{Te\hspace{-.08em}V}}\xspace}

\newacro{ML}{machine learning}
\newacro{LHC}{CERN Large Hadron Collider}
\newacro{HEP}{high energy physics}
\newacro{DL}{deep learning}
\newacro{SGD}{stochastic gradient descent}
\newacro{BO}{Bayesian Optimization}
\newacro{GP}{Gaussian Process}
\newacro{AD}{automatic differentiation}
\newacro{KL}{Kullback-Leibler}
\newacro{MLP}{Multi-Layer Perceptron}
\newacro{TL}{Transfer Learning}
\newacro{RL}{Reinforcement Learning}
\newacro{1D NF}{one-dimensional normalizing flow}

\DeclareMathOperator*{\argmin}{arg\,min}

\def\hfilplus{\hspace{40pt}\hfil}

\begin{document}

\title{End-to-End Optimal Detector Design with Mutual Information Surrogates}

%

\author{Kinga Anna Wo\'zniak}%
\affiliation{University of Geneva}%
\author{Stephen Mulligan}%
\affiliation{University of Geneva}%
\author{Jan Kieseler}%
\affiliation{Karlsruhe Institute of Technology}%
\author{Markus Klute}%
\affiliation{Karlsruhe Institute of Technology}%
\author{Fran\c{c}ois Fleuret}%
\affiliation{University of Geneva, Meta}%
\author{Tobias Golling}%
\affiliation{University of Geneva}%

\begin{abstract}
We introduce a novel approach for end-to-end black-box optimization of \ac{HEP} detectors using local \ac{DL} surrogates. These surrogates approximate a scalar objective function that encapsulates the complex interplay of particle-matter interactions and physics analysis goals. In addition to a standard reconstruction-based metric commonly used in the field, we investigate the information-theoretic metric of mutual information. Unlike traditional methods, mutual information is inherently task-agnostic, offering a broader optimization paradigm that is less constrained by predefined targets.

We demonstrate the effectiveness of our method in a realistic physics analysis scenario: optimizing the thicknesses of calorimeter detector layers based on simulated particle interactions. The surrogate model learns to approximate objective gradients, enabling efficient optimization with respect to energy resolution.

Our findings reveal three key insights: (1) end-to-end black-box optimization using local surrogates is a practical and compelling approach for detector design, providing direct optimization of detector parameters in alignment with physics analysis goals; (2) mutual information-based optimization yields design choices that closely match those from state-of-the-art physics-informed methods, indicating that these approaches operate near optimality and reinforcing their reliability in \ac{HEP} detector design; and (3)  information-theoretic methods provide a powerful, generalizable framework for optimizing scientific instruments. By reframing the optimization process through an information-theoretic lens rather than domain-specific heuristics, mutual information enables the exploration of new avenues for discovery beyond conventional approaches.
\end{abstract}

\maketitle

\section{Introduction}\label{sec:intro}

\begin{table*}[t]
    \centering
    \renewcommand{\arraystretch}{1.2}
    \begin{tabular}{ll}
        \toprule
        \midrule
        \multicolumn{2}{c}{\textbf{\textcolor{gray}{Problem Definition}}} \\
        $\theta \in \mathbb{R}^F$ & Design parameter vector for a detector characterized by $F$ features \\
        $D_\theta$ & Simulated detector parametrized by $\theta$ \\
        $x \in \mathbb{R}^U$ & Particle with $U$ truth features \\
        $y \in \mathbb{R}^{V}$ & $V$-dimensional detector response for particle $x$ \\
        $\delta \to \mathbb{R}$ & Scalar valued objective function \\
        $R$ & Objective model implementing $\delta$ \\
        $S$ & Surrogate model approximating $\delta$ \\
        \midrule
        \multicolumn{2}{c}{\textbf{\textcolor{gray}{Iterative Optimization}}} \\
        $\theta^k \in \mathbb{R}^F$ & One of $K$ test points in the local neighborhood of $\theta$ \\
        $x^k \in \mathbb{R}^{M \times U}$ & $M$ particles generated for test point $k$ \\
        $y^k \in \mathbb{R}^{M \times V}$ & Response of a detector parametrized by $\theta^k$ to particles $x^k$ \\
        \bottomrule
    \end{tabular}
    \caption{Summary of symbols and their definitions.}
    \label{tab:nomenclature}
\end{table*}

The rapid progress of scientific research, driven by growing datasets and increasingly sophisticated ana\-lysis methods, places substantial demands on detector development in \ac{HEP}. As experiments become more complex, determining optimal detector design parameters becomes increasingly challenging. Notable examples include the forthcoming high-luminosity upgrade of the \ac{LHC} \cite{hl_lhc_osti_1767028}, where the number of collisions per bunch crossing is expected to increase by an order of magnitude, as well as the conceptual design of the Future Circular Collider (FCC) \cite{fcc_hh_7060500e92d544b297d50ba27e51dc95}, which aims to achieve significantly higher precision in the FCC-ee phase and significantly higher center-of-mass energies, reaching up to \SI{100}{\TeV} in the FCC-hh phase -- far exceeding the \SI{14}{\TeV} of the \ac{LHC}—along with enhanced luminosities for precision measurements and new physics searches. Both phases will pose significant and new challenges to the design of large-scale detectors.

Detector optimization is inherently challenging. The design parameter space is typically high-dimensional, requiring sophisticated simulation software to model detector response accurately. Moreover, the entire optimization process is stochastic, adding an additional layer of complexity.


Furthermore, the evaluation of the fitness of a set of parameters presents two major challenges. First, it incorporates detector simulations that are typically non-differentiable, or at the very least, the gradient with respect to the design parameters is not readily accessible. Second, the parameter space may exhibit multiple local and global optima that are unknown a priori. Given the high computational cost of evaluating the simulation, it is impractical to sample the design space densely enough to fully characterize its structure~\cite{Jones1998EfficientGO}.

To address these challenges, we propose an iterative \ac{ML}-based optimization framework for determining the optimal design parameters of a simulated \ac{HEP} detector. This approach enables end-to-end optimization by leveraging a differentiable \ac{DL} surrogate.\\

\subparagraph{Problem Statement} End-to-end detector development in \ac{HEP} typically involves a \textit{forward simulation} and \textit{backward reconstruction} step (see Table~\ref{tab:nomenclature} for a summary of symbol definitions). Given a set of design parameters $\theta$ (e.g., number and thickness of detector segments) and an objective metric $\delta$ (e.g. energy resolution), a forward simulation is constructed to model the interaction of a physics process $X$ (e.g., photon production) with the detector $D_{\theta}$. This simulation captures the resulting signals, such as energy deposits in calorimeter cells. The detector response to the physics process, $D_{\theta}(X)$, is then processed by a reconstruction module, which aims to recover the original event $\hat{X}$ from these recorded traces.
The primary goal of this process is to minimize the discrepancy between the true event $X$ and its reconstructed counterpart $\hat{X}$ under the chosen metric $\delta$ with respect to the optimal configuration $D_{\theta^*}$:
 \[ \theta^* = \argmin_{\theta} \delta(X,D_{\theta}(X)). \]
In the standard setup, the objective function involves a Monte Carlo simulator and is not uniquely defined, rendering the optimization problem non-differentiable with respect to $\theta$. Consequently, heuristic methods are typically employed to navigate the design space. \\
In this work, we introduce a surrogate model trained to approximate a localized region of the design parameter space $\theta$ and performance metric $\delta$. By enabling differentiation with respect to the design parameters, this surrogate model facilitates an efficient search for the optimal detector configuration $D_{\theta^*}$. \\

\subparagraph{Contribution} We implement an end-to-end optimization framework that integrates the forward detector response and backward reconstruction process into a unified objective. \\
Our innovation is twofold. First, we define the optimization objective as a scalar-valued function, enabling efficient iterative optimization while encapsulating the local optimality of the detector design for a given physics task within a single metric.
Locality, in this context, refers to surrogate models operating within a confined region of the parameter space at each iteration, a necessary strategy given the impracticality of exhaustively sampling the entire space in realistic detector optimization.
Second, we introduce and systematically study \ac{TL} to mitigate the limitations of locality. We test the hypothesis that the surrogate model can transfer knowledge about the local optimality across iterations, progressively refining and expanding the explored regions of the design parameter landscape.

Designing an effective objective function is a central challenge in experimental optimization within the fundamental sciences~\cite{dorigo2025utilityfunctionexperimentsfundamental}. In this work, we introduce two optimization framework variants, distinguished primarily by their choice of objective function. A key aspect of this study is the comparison between these two fundamentally different metrics. \\
The first variant, RECO-OPT, follows a conventional approach commonly used in high-energy physics (HEP) detector development. Here, the objective function is defined as the reconstruction loss, which quantifies the discrepancy between a particle’s true features and their approximations obtained through a detector reconstruction model. This serves as a benchmark for evaluating optimization performance, aligning with established methodologies. \\
In contrast, the second variant, MI-OPT, adopts an information-theoretic perspective, abstracting the optimization process from predefined target specifications. Instead of relying on explicit objectives, MI-OPT optimizes mutual information between a particle's true features and the detector's output signal. This task-agnostic metric eliminates the need for user-defined objectives, providing a more generalizable approach to instrument optimization. \\
Additionally, both frameworks can incorporate constraints to accommodate practical considerations, such as material limitations or financial restrictions.\\

In Section \ref{sec:related_work} we discuss related work, in Section \ref{sec:data} we present the data of our studies, in Section \ref{sec:methods} we detail the developed methods. The experimental procedures are outlined in Section \ref{sec:experiments} while the results are presented in Section \ref{sec:results}. The conclusion and outlook for further research is given in Section \ref{sec:conclusio}.

\section{Related Work}\label{sec:related_work}

In general, optimizing a non-differentiable system can be approached in three ways: transforming it into a differentiable form, applying gradient-free optimization methods, or constructing a surrogate model to approximate the system. \\

One approach to making a non-differentiable system differentiable is to leverage tools from \ac{DL}, in particular \ac{AD} techniques which allow for precise gradient computation.
Recent studies in \ac{HEP} have augmented existing forward (e.g. Geant4 simulation, particle showering simulation) and backward (e.g. jet-clustering) toolkits by algorithmic differentiation~\cite{aehle2024efficientforwardmodealgorithmicderivatives,aehle2024optimizationusingpathwisealgorithmic,kagan2023branchestreetakingderivatives,dorigo2024endtoendoptimizationswgoarray}. Others have utilized these advancements to develop fully differentiable pipelines for end-to-end optimization of detector configurations, demonstrating significant performance gains and highlighting the potential of \ac{DL}-driven optimization in future \ac{HEP} experiments~\cite{Baydin02012021,DORIGO2023100085,Strong2024TomOpt}. However, these methods require rewriting the detector simulation into a differentiable variant, which can be a significant limitation. Our approach aims to achieve the same end-to-end optimization while circumventing the need for such modifications.\\


There are also optimization techniques that operate without requiring gradient information of the objective function. Examples include \ac{RL}~\cite{qasim2024physicsinstrumentdesignreinforcement,kortus2025constrainedoptimizationchargedparticle} and genetic algorithms \cite{Golovin2017BlackBoxGenetic,Hofler2013innovativeGenetic,Geisel2013evolutionaryOpt,Rolla2023evolutionAntenna}. A drawback of \ac{RL} is its high demand for data and computational resources. While genetic algorithms are more lightweight, they often face convergence challenges, in particular for a large number of parameters. Both heuristics demand a substantial number of objective function evaluations. This becomes problematic in \ac{HEP} detector design, where each evaluation is computationally expensive.

\ac{BO} is another gradient-free technique that has gained prominence in scenarios where quantifying the uncertainty of objective function predictions is essential. This approach leverages \ac{GP} models, which enable non-parametric global optimization across diverse domains (for an overview, see~\cite{Shahriari2016revBayesOpt}). While traditionally employed for global optimization, recent advancements have introduced localized variants \cite{Eriksson2019scalableGlobal,Wu2023behaviorConvergence,Nguyen2022LocalBO}, addressing key limitations such as scalability to high-dimensional problems—making them more suitable for applications like the one considered here.
However, several limitations persist. The computational cost of \ac{BO} scales as $\mathcal{O}(N^2)$ or $\mathcal{O}(N^3)$ with the number of samples, posing challenges for \ac{HEP} applications where large datasets are common. Moreover, \ac{BO} requires problem-specific tuning, including the selection of handcrafted kernels and hyperparameters, which assumes prior knowledge of the underlying objective function or response structure \cite{Zhang2020bayesianOpt,Khatamsaz2023physicsInformed,Nicoli2023physicsInformed}. Given the complexity of our objective space and the need for a flexible framework capable of seamlessly integrating new objectives (e.g., the identification of particle types), we instead opt for a \ac{DL} surrogate approach. \\

Recently, the use of \ac{DL}-driven surrogates for black-box optimization tasks has gained significant traction~\cite{Grbcic2025AILaser,elzouka2020interpretable,Fanelli2022designDetectAi}. The choice of the surrogate model depends on the nature and domain of the problem. Notably, generative approaches have shown success in various applications (for an overview, see~\cite{dgm_surrogates_HASHEMI2024100092}).
For example, \cite{black_box_NEURIPS2020_a878dbeb} tackles detector optimization in \ac{HEP} with intractable likelihoods using local generative surrogates. By iteratively exploring the design space, the surrogate approximates the objective function’s gradient, guiding optimization toward an optimal design.
This approach is extended by \cite{schmidt2025endtoenddetectoroptimizationdiffusion} to an end-to-end framework, integrating forward simulation and backward reconstruction while reducing surrogate complexity. Building on these advancements, we propose an end-to-end optimization framework that leverages a novel information-theoretic methodology. \\

The choice of optimization metric plays a crucial role in most related studies. Information-theoretic metrics have proven effective across various domains and have been, for example, incorporated into \ac{BO} frameworks discussed earlier \cite{Neiswanger2021BayesianAE,Chitturi2024targetedMaterials}.

A well-established information-theoretic measure for capturing nonlinear statistical dependencies is the \textit{mutual information} criterion~\cite{macKay2002info,cover1999elements}. This metric has been successfully applied in diverse fields, including biology~\cite{Bitbol2018InferringIP}, neuroscience~\cite{paninski2003estimation}, speech recognition~\cite{bahl1986maxMIspeech}, and linguistics~\cite{church-hanks-1990-word}, demonstrating its versatility and effectiveness. Given these strengths, we adopt mutual information as the primary metric for our investigation and compare its performance to classic approaches discussed above, demonstrating how the new measure opens the avenue for a task-agnostic instrument assessment. \\

\section{Data}
\label{sec:data}

\begin{figure}
    \centering
    \includegraphics[width=0.34\textwidth]{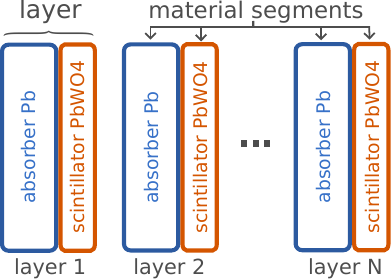}
    \caption{Calorimeter consisting of layers of interleaved lead absorber and lead-tungstate scintillator segments.}
    \label{fig:calo-diagram}
\end{figure}

In this work, we employ a dataset generated by simulating single photons incident on a calorimeter detector. These simulations are carried out using the Geant4 framework~\cite{Geant_4_0_AGOSTINELLI2003250, Geant4_1_1610988, Geant4_2_ALLISON2016186}, specifically employing the FTFP\_BERT physics list. The detector consists of multiple layers, each comprising two distinct material segments: a lead absorber and a lead-tungstate scintillator, as depicted schematically in Figure~\ref{fig:calo-diagram}. For this proof of concept, we choose PbWO$_4$ as the active scintillator material, a commonly used scintillator material in HEP ~\cite{pwo_alice_ALEKSANDROV2005169, pwo_lhc_LECOQ1995291, pwo_highly_compact_oriented_10.3389/fphy.2023.1254020}, and Pb as the absorber. During simulation, only the energy deposited in the active material is recorded, and this information is passed on to subsequent processing steps. We do not include a detailed simulation of the readout electronics.

The thickness of each material segment is treated as an optimizable design parameter. Given a nominal parameter vector $\theta \in \mathbb{R}^F$ where $F$ is the number of design features (for example $F=6$ if we want to optimize the segment-thicknesses of a detector composed of three layers), we draw a collection of $K$ perturbed parameter vectors 
\[
  \theta^k \sim \mathcal{N}(\theta, \sigma),
\]
where the width $\sigma$ is chosen to be of the order of the radiation length (approximately 1\,cm) \cite{pwo_alice_ALEKSANDROV2005169}. This strategy follows a trust-region approach similar to that outlined in~\cite{black_box_NEURIPS2020_a878dbeb, trust_region_doi:10.1137/0719026}. For each parameter vector $\theta^k$, we simulate particles incident on the detector, generating true features $x^k$, as well as measured features $y^k$.

Our inputs, $x$, as well as their progression through the detector (e.g., particle showering) and its response, $y$, are inherently stochastic.

\section{Methods \& Models: Optimal Design Surrogate}
\label{sec:methods}

The primary purpose of the optimization is to determine the optimal thicknesses for the scintillator and absorber segments of a multi-layer calorimeter (Figure~\ref{fig:calo-diagram}). The design parameters $\theta$ in our study thus correspond to thickness values in centimeters.  


We develop two local surrogate optimization pipelines that differ primarily in their choice of objective metric $\delta$ (see Section \ref{sec:intro}) and type of \textit{surrogate model} used.
The first approach, RECO-OPT, relies on a conventional reconstruction accuracy metric commonly used in detector assessment as well as a generative surrogate. RECO-OPT serves as a reference baseline typically used in the field. The second approach, MI-OPT, introduces an innovative optimization framework based on the information-theoretic metric of mutual information. This novel perspective enhances task-agnostic optimization and is invariant under homeomorphic transformations~\cite{kraskov2004estimatingMI}, offering a more robust and generalizable solution.
\begin{figure*}
    \centering
	\begin{subfigure}{0.44\textwidth}
	    \centering
		\includegraphics[width=\textwidth]{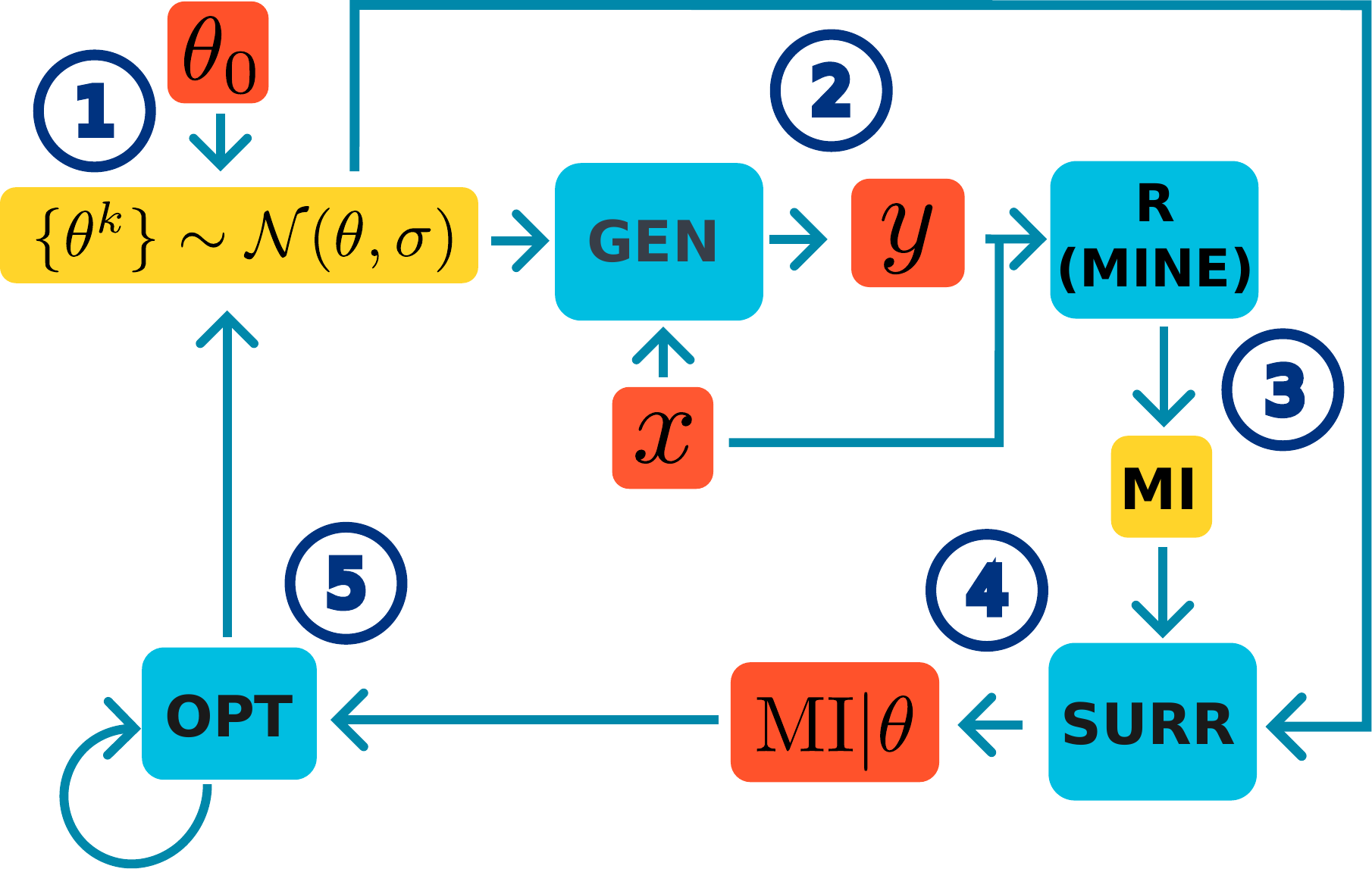}
        \label{fig:mine_pipe}
	\end{subfigure}
	\hfill
	\begin{subfigure}{0.44\textwidth}
	    \centering
		\includegraphics[width=\textwidth]{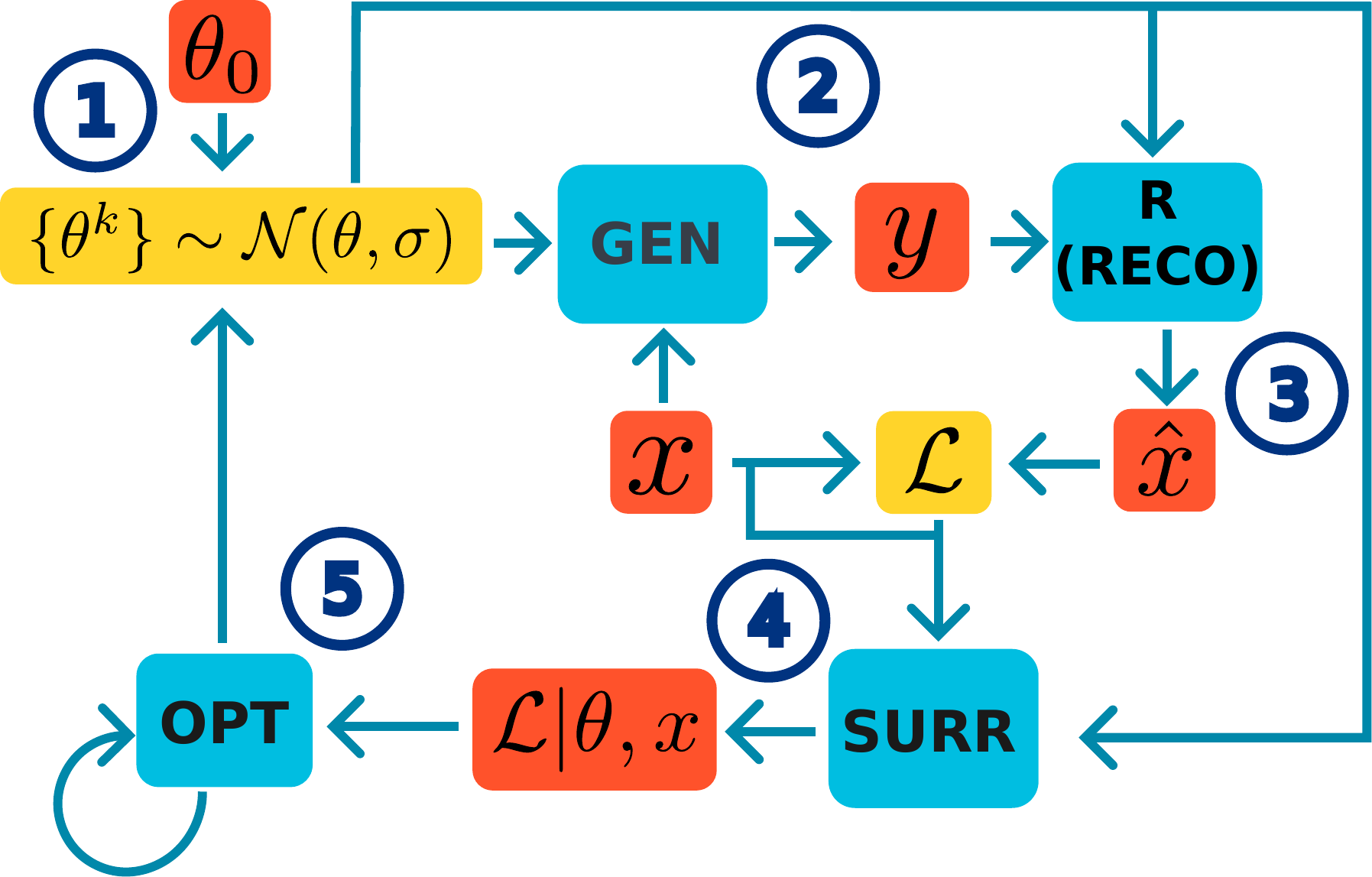}
        \label{fig:reco_pipe}
	\end{subfigure}
    \caption{Two developed optimization pipelines: MI-OPT using mutual information as the objective function (left) and REC-OPT using the loss of a reconstruction model for that purpose (right).}
    \label{fig:pipeline}
\end{figure*}
\subparagraph{\textbf{Optimization loop}} Both variants are integrated into an optimization loop framework which consists of five main parts: 
\begin{enumerate}
    \item Sampling of a set of test points from a local neighborhood in the design parameter space.
    \item Simulation of a physics process and detector response (\textsc{Geant4}, see Section~\ref{sec:data}).
    \item Training of an objective model $R$ which computes the objective function $\delta$.
    \item Training of a surrogate model $S$ to approximate the relationship between objective function $\delta$ and parameters $\theta$.
    \item Gradient descent on the surrogate's approximated objective with respect to the design parameters $\frac{\partial S}{\partial \theta} \approx \frac{\partial \delta}{\partial \theta}$.
\end{enumerate}
The framework for both variants is illustrated in Figure~\ref{fig:pipeline} (MI-OPT on the left, RECO-OPT on the right) and described in detail below. \\

At the start of each optimization cycle, MI-OPT and RECO-OPT follow an identical procedure. Each iteration begins with an initial nominal parameter, $\theta$. Based on this nominal value, a set of test points is generated from its local neighborhood (Figure~\ref{fig:pipeline}, step~\raisebox{.5pt}{\textcircled{\raisebox{-.9pt} {1}}}). Specifically, $K$ neighboring candidates, $\theta^k$ are sampled from a normal distribution centered at nominal $\theta$, with a standard deviation $\sigma$ that defines the neighborhood size: $\{\theta^k\}_{k=1}^K \sim  \mathcal{N}(\theta, \sigma)$. The locality parameter $\sigma$ is set to~\SI{1.5}{\cm}. 

For each parameter vector $\theta^k \in \mathbb{R}^F$, we simulate $M$ photons incident on the detector, generating $U$ true features per particle $x^k \in \mathbb{R}^{M \times U}$, as well as their corresponding $V$ measured calorimeter energy deposits $y^k  \in \mathbb{R}^{M \times V}$ (Figure~\ref{fig:pipeline}, step~\raisebox{.5pt}{\textcircled{\raisebox{-.9pt} {2}}}).

In each iteration of the pipeline, the triplets $\{\theta^k, x^k, y^k\}_{k=1}^K$ are used as inputs to the objective \ac{ML} models. The objective model $R$ implements the objective criterion and computes a corresponding metric $\delta^k$ for each triplet (Figure~\ref{fig:pipeline}, step~\raisebox{.5pt}{\textcircled{\raisebox{-.9pt} {3}}}). Based on the quadruplet  $\{x^k,y^k,\theta^k,\delta^k\}_{k=1}^K$, a surrogate model $S$ learns the relationship between the inputs and the target objective (Figure~\ref{fig:pipeline}, step~\raisebox{.5pt}{\textcircled{\raisebox{-1.0pt} {4}}}). Parts \raisebox{.5pt}{\textcircled{\raisebox{-.9pt} {3}}} and \raisebox{.5pt}{\textcircled{\raisebox{-.9pt} {4}}} of the pipeline differ between MI-OPT and RECO-OPT in terms of objective models, surrogate models and objective function. These differences are discussed in detail in the dedicated paragraphs below.

After training the surrogate on $K$ new candidate points, \ac{SGD} is applied to identify the next optimal design parameters. By following the gradient $-\frac{\partial S}{\partial \theta}$, the optimizer searches for the most promising parameter set based on the learned approximation of the objective function. This search is confined to a local neighborhood around the current nominal $\theta$ and is halted if it exceeds predefined boundaries. Once a new optimal value is determined, the algorithm updates the nominal $\theta$ accordingly and proceeds to the next iteration. The process repeats from step~\raisebox{.5pt}{\textcircled{\raisebox{-.9pt} {1}}}, continuing until a set number of iterations is completed. The entire end-to-end optimization procedure is summarized in the framework Algorithm~\ref{alg:end_to_end}.
\begin{algorithm}[H]
    \caption{End-to-end Optimization Loop}\label{alg:end_to_end}
        \begin{algorithmic}[1]
            \Require number K of $\theta$ candidate points, number $M$ of particles $x$, size of the local region $\epsilon$, number of iterations I, measure of locality $d$
            \State Choose initial nominal $\theta = \theta_{0}$
            \While{$i \leq \text{I}$}
                \State Sample $\mathbf{\theta}^{k} \sim \mathcal{N}(\mathbf{\theta}, \sigma)$, $k = 1, \ldots , K$
                \State For each $\mathbf{\theta}^k$, sample $M$ truth inputs $x^k \sim q(x)$ 
                \State Run detector simulation $y^k = D(x^k,\theta^k)$
                \State Obtain surrogate \\
                \hspace{0.6cm}$\nabla_{\theta} S \gets$ \Call{RECO-OPT}{$\{x^k,y^k,\theta^k\}$} or \\
                \hspace{0.6cm}$\nabla_{\theta} S \gets$ \Call{MI-OPT}{$\{x^k,y^k,\theta^k\}$}
                \While{$d(\theta^*, \theta) < \epsilon$}
                    \State $\theta^* \gets \text{SGD}(\theta, \nabla_{\theta} S)$
                \EndWhile
                \State $\theta \gets \theta^*$
                \State $ i = i+1$
            \EndWhile
        \end{algorithmic}
\end{algorithm}

The next two subsections describe objective models and surrogates of RECO-OPT and MI-OPT in more detail.

\subparagraph{\textbf{RECO-OPT}}
For the standard, explicit reconstruction based optimization, the objective function is taken to be a normalized mean squared error loss between the true energy $x^k$ and some reconstructed energy. The objective model $R(y^k | \theta^k)$ is trained to reconstruct the true energy $x^k$ of the incident photon from the energy deposited in the scintillators $y^k$. The model is implemented as a \ac{MLP}.

A \textit{generative} surrogate model $S_{\omega}(z, x^k, \theta^k)$ is then trained to approximate the simulation and reconstruction step, directly predicting the loss of the reconstruction model $\delta^k$, for a given $\{x^k, \theta^k\}$ pair.

Normalizing flows \cite{normalising_flows_0_rezende2016variationalinferencenormalizingflows,normalising_flows_1_Kobyzev_2021,normalizing_flows_papamakarios_JMLR:v22:19-1028}, one form of generative model that can be used as a surrogate, model the distribution $p_{w}$ of some data $\mathbf{w}$ as a bijective transformation $f$ from a known noise distribution $\mathbf{z} \sim p_z(z)$, and can be used for conditional distributions $p_{w}(\mathbf{w|c})$ \cite{conditional_nf_winkler2023learninglikelihoodsconditionalnormalizing}. 

Since the end-to-end approach maps to one output scalar, we introduce the \ac{1D NF} model, which uses a piecewise linear transformation as our bijective function $f$.
This is similar to the piecewise linear coupling transforms used in \cite{linear_splines_müller2019neuralimportancesampling} except this is \textit{not} structured as a coupling transform, given that our input is scalar valued.

A piecewise linear transform $f$ of an input $w$ is defined between some $w_{min}$ and $w_{max}$, where the interval between the two values is divided based on a preselected number of bins, $N_{bins}$. 
Values below $w_{min}$ are mapped to a constant value $\alpha$ while values above $w_{max}$ are mapped to the value of the highest bin.
For a given input, an MLP that takes the conditional values $\mathbf{c}$ as input is used to predict the slopes $m_i$ in each linear piece. The output of the MLP is then used to parameterize the piecewise linear transformation and calculate the output $f(w)$ as:

\begin{equation}
    f(w) = \begin{cases}
        \alpha & \text{if } w < w_{min} \\
        m_i (w - w_{\text{offset}}) + b_i & \text{if } w \in \text{bin}_i
    \end{cases}
\end{equation}

where $m_i$ and $b_i$ are the slope and intercept of the linear transformation in the $i$-th bin, and $w_{\text{offset}}$ the lower bound of some given bin.

The \ac{1D NF} surrogate is trained to transform between a normal distribution $\mathcal{N}(0, 1)$ and the logarithm of the distribution of the reconstruction loss, conditioned on the true energy $x$ and detector parameters $\mathbf{\theta}$. The logarithm of the reconstruction loss is chosen as it is close to a normal distribution. For evaluation, we consider the exponent of the surrogate output.
The surrogate is trained by minimizing a negative log-likelihood loss.
When the surrogate is trained, given an energy $x$, a set of detector parameters $\theta$, and some sampled $z$, the learned distribution of reconstruction losses for these conditional values can be sampled from, estimating the loss for a single event.
These sample outputs are differentiable with respect to $\theta$ and can be used to optimize the surrogate parameters via gradient descent.
A description of the procedure is given in surrogate Algorithm \ref{alg:reco_opt_iter}.
\begin{algorithm}[H]
    \caption{RECO-OPT Surrogate Variant}\label{alg:reco_opt_iter}
        \begin{algorithmic}[1]
            \Require Surrogate generative model $S_{\omega}$
            \State Train objective model $R_{\psi}(y|\theta)$
            \State Fix weights $\psi$ of objective model
            \State Collect objective scores $\delta$ for each sample associated with a parameter value $\{x, \delta, \theta \}$   
            \State Train $S_{\omega}(z, \delta;x,\theta)$, where $z \sim \mathcal{N}(0, 1)$
            \State Fix weights of surrogate model $\omega$
            \State Sample $\hat{\delta} = S_{\omega}(z, x;\theta)$, $z \sim \mathcal{N}(0,1)$
            \State $\nabla_{\theta}\delta \approx \nabla_{\theta}\EX[\hat{\delta}] \gets \frac{1}{M}\sum\limits_{m=1}^{M}\frac{\partial S_{\omega}(z, x,\theta)}{\partial \theta}$
        \end{algorithmic}
\end{algorithm}

In this work we study two generative approaches: a diffusion model based surrogate as in \cite{schmidt2025endtoenddetectoroptimizationdiffusion} and the \ac{1D NF} described above.
Initially the use of the diffusion model was explored, but it was found that the computational cost of training was high and did not produce noticeable gains in performance in this setting. 
As such the \ac{1D NF} model is primarily used here. 

\subparagraph{\textbf{MI-OPT}}
The MI-OPT framework variant employs mutual information as its objective function. This metric captures the interdependence between two random variables by evaluating their joint distribution, eliminating the need for explicit domain-specific modeling.\\

Mutual information between two random variables $A$ and $B$ is expressed as \[ MI(A,B) = H(A) - H(A|B)\] and describes how much information $B$ conveys about $A$ \cite{macKay2002info,cover1999elements}.
Given two known distributions of $A$ and $B$, it is computed as 
\[
I(A;B) = \int_{A \times B} \log \left( \frac{dP_{AB}}{dP_A \otimes P_B} \right) dP_{AB}.
\] \\
The metric can be expressed as the \ac{KL} divergence between the joint distribution of the random variables and the product of their marginals
\[
I(A;B) = D_{\text{KL}}(P_{AB} \,||\, P_A \otimes P_B).
\]
As demonstrated by~\cite{belghazi2018mutual}, this \ac{KL} formulation allows for the implementation of a mutual information approximator as a neural network, although simpler divergences can also be used~\cite{hjelm2018learning}.
This is made possible by its dual formulation, known as the Donsker-Varadhan representation
\[
D_{\text{KL}}(P \,||\, Q) = \sup_{T : \Omega \to \mathbb{R}} \left( \mathbb{E}_P[T] - \log\left(\mathbb{E}_Q\left[e^T\right]\right) \right)
\]
where $T$ is a family of parametrizeable functions. Then $T_\omega$ is chosen to be a feed forward neural network with learnable parameters $\omega$ such that
\[
\begin{split}
D_{\text{KL}}(P(A,B) \,||\, P(A) \otimes P(B)) \\
= \sup\limits_{\omega \in \Omega}  \mathbb{E}_{P_{AB}}[T_\omega] - \log (\mathbb{E}_{ P(A) \otimes P(B)}[e^{T_\omega}]).
\end{split}
\]

Mutual information can serve as a powerful metric for evaluating how effectively a given detector design retains information about a particle’s true properties as it traverses the apparatus. Here, the two random variables represent a key particle characteristic and the corresponding detector response. In our study, $A$ corresponds to the true energy of the particle, while $B$ represents the set of calorimeter energy deposits. A high mutual information value between these variables indicates that the detector preserves the essential physics information from the particle’s interaction to its recorded response. By maximizing mutual information, we can therefore drive the optimization toward detector designs that enhance information retention and overall performance. \\
The metric bears three significant advantages: First, it enables us to distill the quality of a detector design into a single value that reflects its performance across multiple physics goals. Second, the metric is task-agnostic, freeing the user of the need to hand-craft an appropriate objective function. Third, due to its structure, it can naturally incorporate multi-objective scenarios. Adding a new goal to the overall objective amounts to expanding the inputs by one dimension. The disadvantages are related: since the metric is computed over an ensemble of samples, it is prone to the curse of dimensionality, generally requiring more samples than traditional methods. Second, weighting of goals is not straight-forward and would have to be addressed by future work. 

The MI-OPT branch utilizes the mutual information neural estimator~\cite{belghazi2018mutual} as its objective model $R$. This estimator receives the particle truth and the detector response (calorimeter hits) as inputs and computes mutual information as the objective metric: $\delta^k = \textrm{MI}^k = R(x^k,y^k)$. In contrast to RECO-OPT, where each of the $M$ samples in $\{x^k,y^k\}$ is assigned an individual objective metric value, $\delta^k \in \mathbb{R}^M$, MI-OPT projects the ensemble onto a single scalar quantity that characterizes the entire distribution, $\delta^k \in \mathbb{R}^1$.

Since mutual information serves as a summary metric over random variables, the surrogate model of the MI-OPT approach does not require a stochastic formulation. Consequently, we employ a \ac{MLP} regression model to approximate the metric effectively. The \ac{MLP} surrogate model $S$ is trained to predict this mutual information metric from the detector parameters: $\hat{\delta^k} = S(\theta^k)$. The procedure is outlined in the surrogate Algorithm~\ref{alg:mi_opt_iter}.
\begin{algorithm}[H]
    \caption{MI-OPT Surrogate Variant}\label{alg:mi_opt_iter}
        \begin{algorithmic}[1]
            \Require Objective model $R_{\phi}$, Surrogate model $S_{\omega}$
            \ForAll{$k \in K$}
                \State Compute MI with objective model: $\delta^k = R(x^k,y^k)$
            \EndFor
            \State Train surrogate model $\hat{\delta^k} = S_{\omega}(\theta^k)$
            \State Fix weights of surrogate model $S_\omega$
            \State Return $\nabla_{\theta} S_{\omega}$
        \end{algorithmic}
\end{algorithm}
Note that for the \ac{TL} studies (Section~\ref{sec:experiments}) of MI-OPT, only the the state of the surrogate model is preserved across iterations. As the mutual information module serves as an approximator for a predefined distribution, it is re-initialized at every iteration.

\section{Experiments}\label{sec:experiments}

We evaluate MI-OPT and RECO-OPT in the optimization of a simplified calorimeter detector composed of multiple layers, simulated using the physics simulation package GEANT4~\cite{Geant_4_0_AGOSTINELLI2003250, Geant4_1_1610988, Geant4_2_ALLISON2016186}, as detailed in Section \ref{sec:data}. The optimization parameters, denoted as $\theta$, correspond to the thicknesses (in centimeters) of the interleaved absorber and scintillator segments (see Figure~\ref{fig:calo-diagram}). \\

The following three main experiment configurations are defined:
\begin{enumerate}
    \item A set of \textbf{base studies} to establish the baseline behaviour of the framework with one, two, and three layers of scintillator-absorber pairs and 700 events of a single photon shot at the detector with an energy range of \qtyrange[range-units=single,range-phrase=-]{1}{20}{\GeV} per parameter set candidate. The nominal parameter neighborhood is covered by 30 candidates for RECO-OPT and 117 candidates for MI-OPT. 
     \item A series of studies that examines the role of \textbf{transfer learning} in the optimization process~\cite{transfer_learning_DBLP:journals/corr/abs-1911-02685,mokhtar2025finetuningmachinelearnedparticleflowreconstruction}, with a particular focus on the effect of reduced dataset sizes. A surrogate model is said to employ \ac{TL} if its weights are \textit{not} reset between iterations, allowing it to retain knowledge from previous training. Conversely, it is said to \textit{not} use \ac{TL} if its weights are reinitialized at each iteration. \\
     The impact of \ac{TL} is assessed by reducing the number of events per iteration, thereby increasing the reliance on information learned in previous iterations for effective parameter space exploration. This study primarily investigates its feasibility, impact, and limitations within the optimization framework.
    \item A set of \textbf{energy studies} that investigates the change in evolution of the detector parameters in the case where the maximum possible energy of a particle is significantly increased. In this case, as before, a single photon is shot at the detector but its energy range is set to vary between \qtyrange[range-units=single,range-phrase=-]{1}{100}{\GeV}. The remaining experiment parameters are equal to the base study.
\end{enumerate}

In each of these settings we compare the mutual information based metric to the use of the simpler reconstruction loss as the surrogate objective. Initial segment thicknesses are \SI{1.0}{\cm} for both scintillators and absorbers and the maximal thickness of the sum of all segments is constrained to \SI{25}{\cm}. This constraint corresponds to roughly 28 radiation lengths in $\text{PbWO}_{4}$ scintillators, approximately the same maximum depth of scintillator used in the electromagnetic calorimeter (ECAL) in the Complex Muon Solenoid (CMS) detector at the LHC \cite{cms_pwo_scint_constraint_de_Fatis_2012}. All results are presented as the mean evolution with standard deviation, computed from three experimental runs per setting. The specific values of the experiment variables are detailed in Table~\ref{tab:var_vals}.

\begin{table}[t]
    \centering
    \renewcommand{\arraystretch}{1.3}
    \begin{tabular}{l@{\hskip 10pt}c@{\hskip 10pt}c}
        \toprule
        \toprule
        \multicolumn{3}{c}{\textbf{\textcolor{gray}{Experiment Settings}}} \vspace{3pt}\\
        \textbf{Variables} & RECO-OPT & MI-OPT \\ 
        \midrule
        $F$: parameter $\mathrm{dim}(\theta)$ & \multicolumn{2}{c}{2,4,6} \\
        $U$: truth feature $\mathrm{dim}(x)$ & \multicolumn{2}{c}{1}\\
        $V$: detector output $\mathrm{dim}(y)$ & \multicolumn{2}{c}{1, 2, 3} \\
        $K$: number candidates $|\theta|$ & 30 & 117 \\
        $M$: number events per $K$ $|\{x,y\}|$& 700, 50, 5 & 700, 50 \\
        \bottomrule
    \end{tabular}
    \caption{Experiment variable values}
    \label{tab:var_vals}
\end{table}

\section{Results}
\label{sec:results}
\begin{figure*}
\centering
	\begin{subfigure}{0.41\textwidth}
	    \centering
		\includegraphics[width=0.98\textwidth]{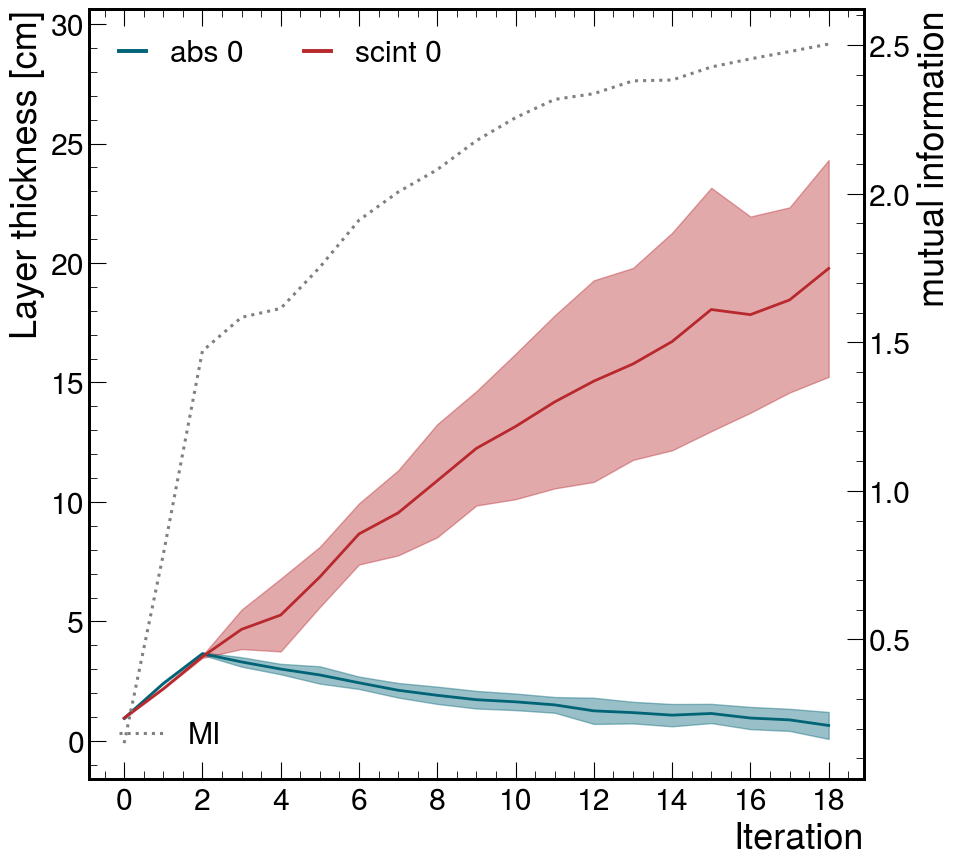}
            \caption{MI-OPT, 1 layer}
            \label{fig:evo_base_1L_mi}
	\end{subfigure}%
	\hfilplus
	\begin{subfigure}{0.41\textwidth}
	    \centering
		\includegraphics[width=\textwidth]{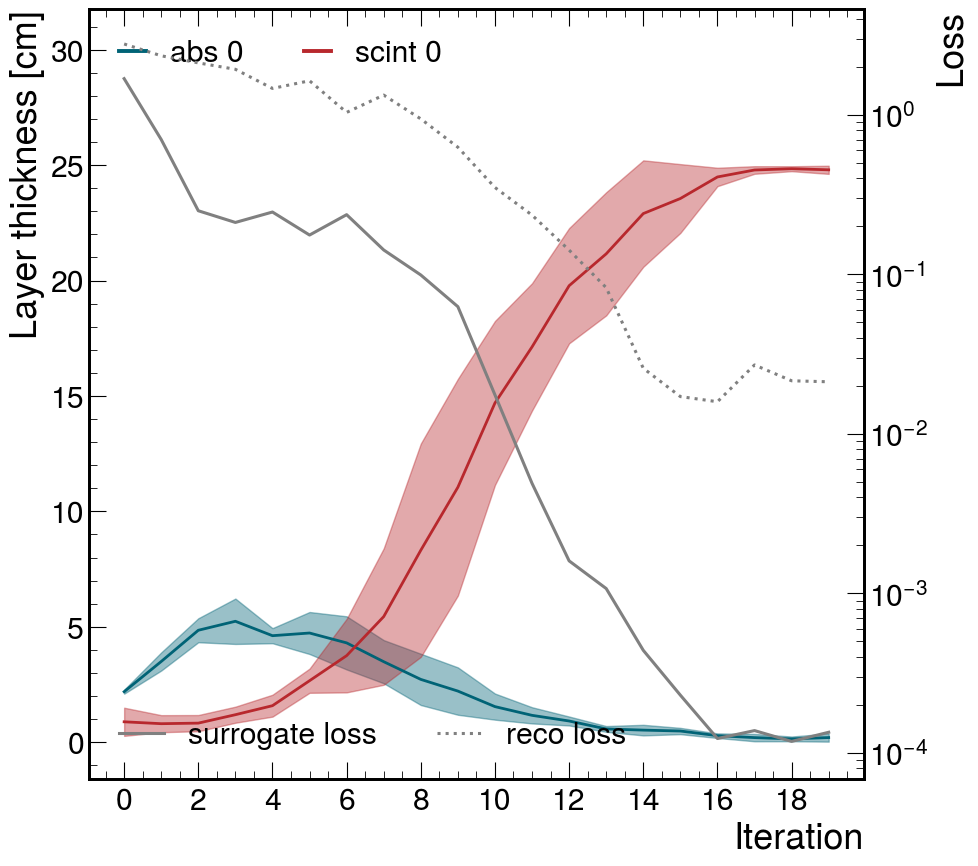}
        \caption{RECO-OPT, 1 layer}
        \label{fig:evo_base_1L_reco}
	\end{subfigure}
    \begin{subfigure}{0.41\textwidth}
	    \centering
		\includegraphics[width=\textwidth]{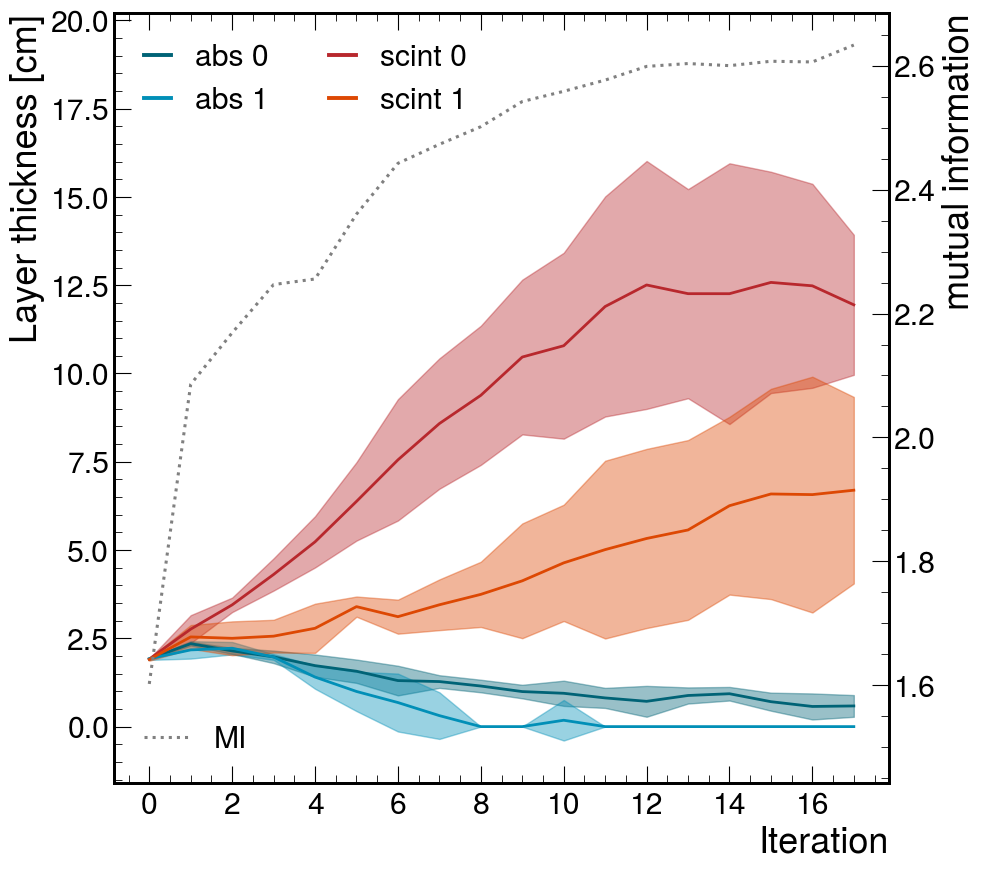}
            \caption{MI-OPT, 2 layers}\label{fig:evo_base_2L_mi}
	\end{subfigure}
	\hfilplus
	\begin{subfigure}{0.41\textwidth}
	    \centering
		\includegraphics[width=0.99\textwidth]{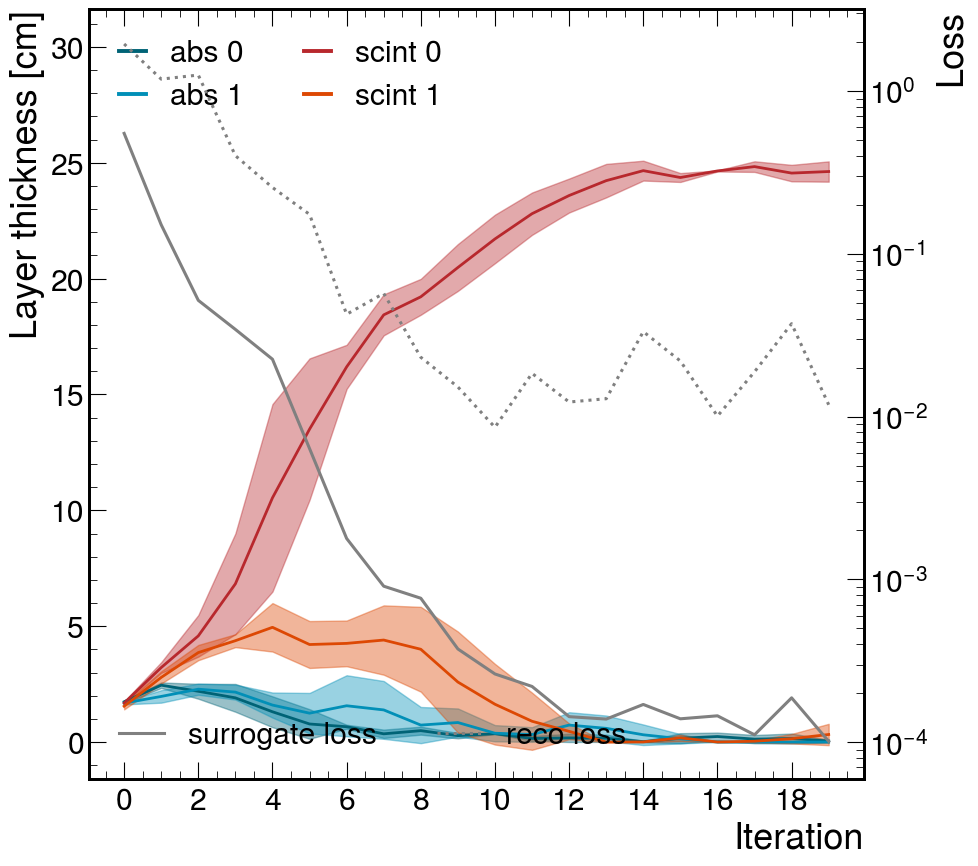}
            \caption{RECO-OPT, 2 layers}\label{fig:evo_base_2L_reco}
	\end{subfigure}
    \begin{subfigure}{0.41\textwidth}
	    \centering
		\includegraphics[width=0.98\textwidth]{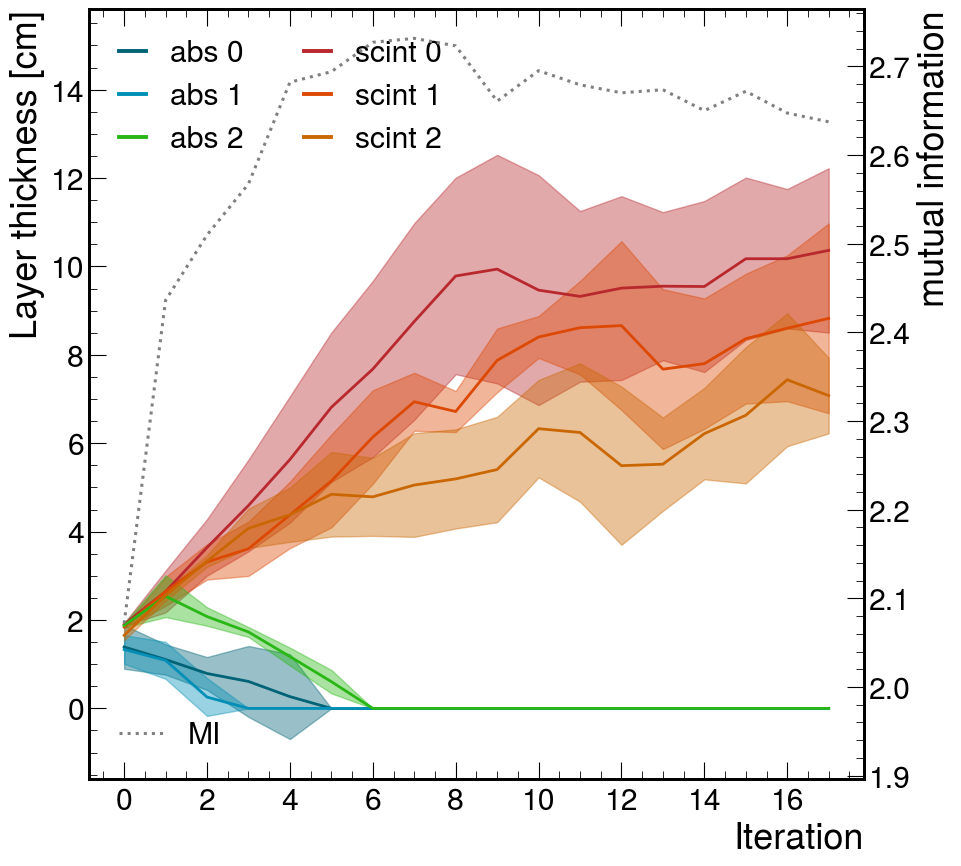}
            \caption{MI-OPT, 3 layers}\label{fig:evo_base_3L_mi}
	\end{subfigure}
	\hfilplus
	\begin{subfigure}{0.41\textwidth}
	    \centering
		\includegraphics[width=\textwidth]{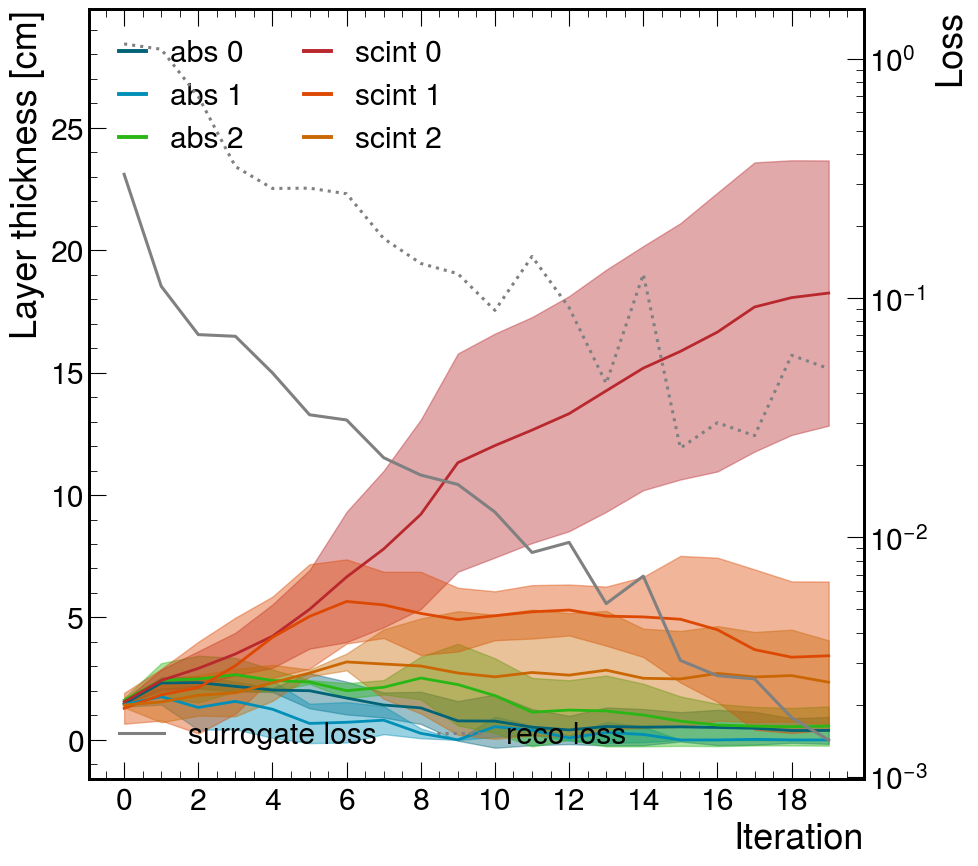}
            \caption{RECO-OPT, 3 layers}
            \label{fig:evo_base_3L_reco}
	\end{subfigure}
\caption{BASE Study: Layer thickness evolution (solid lines) for MI-OPT (left, mutual information maximization in gray dashed) and RECO-OPT (right, reconstruction loss in gray solid, surrogate approximated loss in gray dashed), averaged over three runs for 1-layer (top), 2-layer (middle), and 3-layer (bottom) studies.}\label{fig:evo_base}
\end{figure*}

\subsection{Base Studies}

The results of the \textbf{base studies} are shown in Figure~\ref{fig:evo_base}. The left column displays data for the information theoretic approach MI-OPT while the right column displays data for the standard reconstruction based variant RECO-OPT.

In general, we observe that - as expected - both approaches maximize the scintillator segments while decreasing absorbers. In all tested settings, the absorbers are reduced either to zero or slightly above zero thicknesses, while the total gauge of scintillators increases beyond \SI{20}{\cm}. Additionally, the scalar surrogate metric converges towards an optimum throughout the optimization iterations: mutual information increases in the MI-OPT studies (left column, dotted line), while the surrogate's approximated reconstruction loss decreases in the RECO-OPT studies (right column, dotted line).

While MI-OPT increases the thickness of all available scintillators, RECO-OPT predominantly enhances only the first scintillator segment. This divergence can arise from the two approaches converging to distinct but functionally equivalent optima. 
As the optimization progresses and absorbers are effectively reduced to near-zero, the problem can be considered degenerate as segmentation of the scintillators becomes negligible if their total thicknesses sum to similar amounts (see Appendix~\ref{appendix:summation}). \\
At the initial stages of optimization, both methods maintain at least one absorber with a nonzero thickness while the total scintillator thickness remains below a certain threshold. Absorbing material is discarded only once this limit is exceeded. This behavior reflects the necessity for sufficient energy containment within the calorimeter, ensuring maximal capture of the electromagnetic shower generated by the incident particle~\cite{pwo_highly_compact_oriented_10.3389/fphy.2023.1254020}. \\
As the total gauge of the calorimeter is constrained (by adding a penalization term to the optimization of $\theta$ in the surrogate), we observe how both loss metrics plateau in RECO-OPT as the first scintillator thickness approaches \SI{25}{cm}, while the mutual information metric plateaus in the three-layered study (Figure \ref{fig:evo_base_3L_mi}) alongside scintillator thicknesses whose sum exceeds \SI{25}{\cm}. 

In conclusion, the results of the baseline experiment demonstrate that both optimization approaches successfully identify viable calorimeter designs that minmise or maximise their metric of choice across all case studies—ranging from single-layer to three-layer configurations—relying only on a scalar-valued objective function. The variations in the solutions are attributed to the existence of multiple optima in the objective function. This outcome highlights that an information-theoretic metric, despite offering a fundamentally different perspective on the problem, produces design choices closely aligned with those from traditional methods, confirming that state-of-the-art approaches are indeed operating near optimality.

The base studies were also performed for RECO-OPT using the diffusion based surrogate. A comparison of the single and three layer case to that of the \ac{1D NF} can be found in Appendix \ref{appendix:1d_vs_diff}.



\subsection{Transfer Learning}
\begin{figure*}
\centering
	\begin{subfigure}{0.41\textwidth}
	    \centering
		\includegraphics[width=\textwidth]{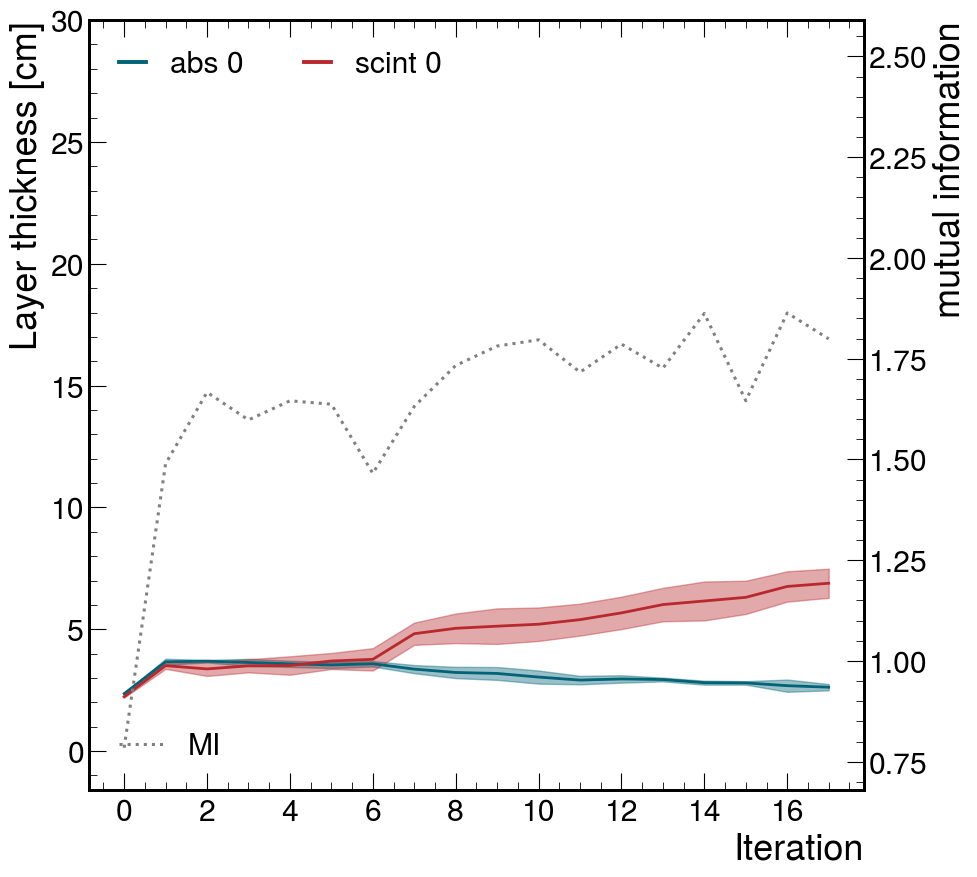}
        \caption{MI-OPT, 1 layer, no TL, 700 events}
        \label{fig:evo_transfer_layers1_700_noTL_mi}
	\end{subfigure}%
	\hfilplus
	\begin{subfigure}{0.41\textwidth}
	    \centering
		\includegraphics[width=\textwidth]{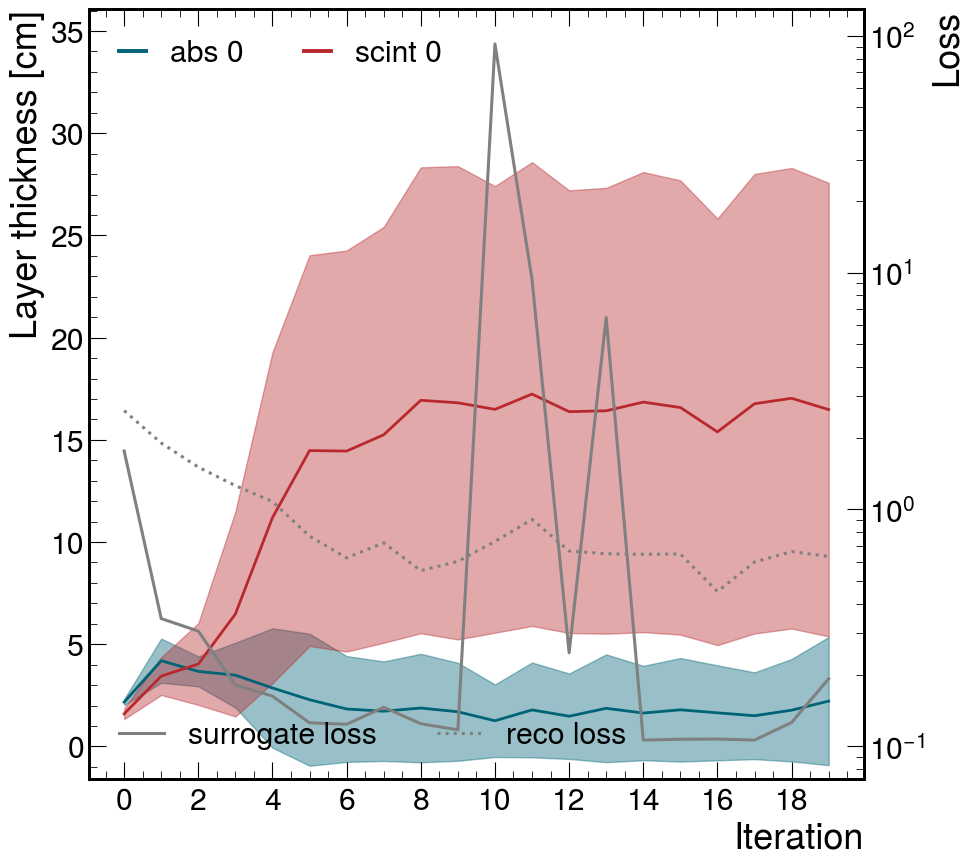}
        \caption{RECO-OPT, 1 layer, no TL, 700 events}
        \label{fig:evo_transfer_layers1_700_noTL_reco}
	\end{subfigure}%
\caption{TRANSFER study (700): Re-initialized learning (no transfer) with full sample size and a single layer calorimeter. Comparison of thickness evolutions for MI-OPT (left) and RECO-OPT (right) approaches.\label{fig:evo_transfer_notl_700ev_1layer}}
\end{figure*}

\begin{figure*}
\centering
	\begin{subfigure}{0.41\textwidth}
	    \centering
		\includegraphics[width=\textwidth]{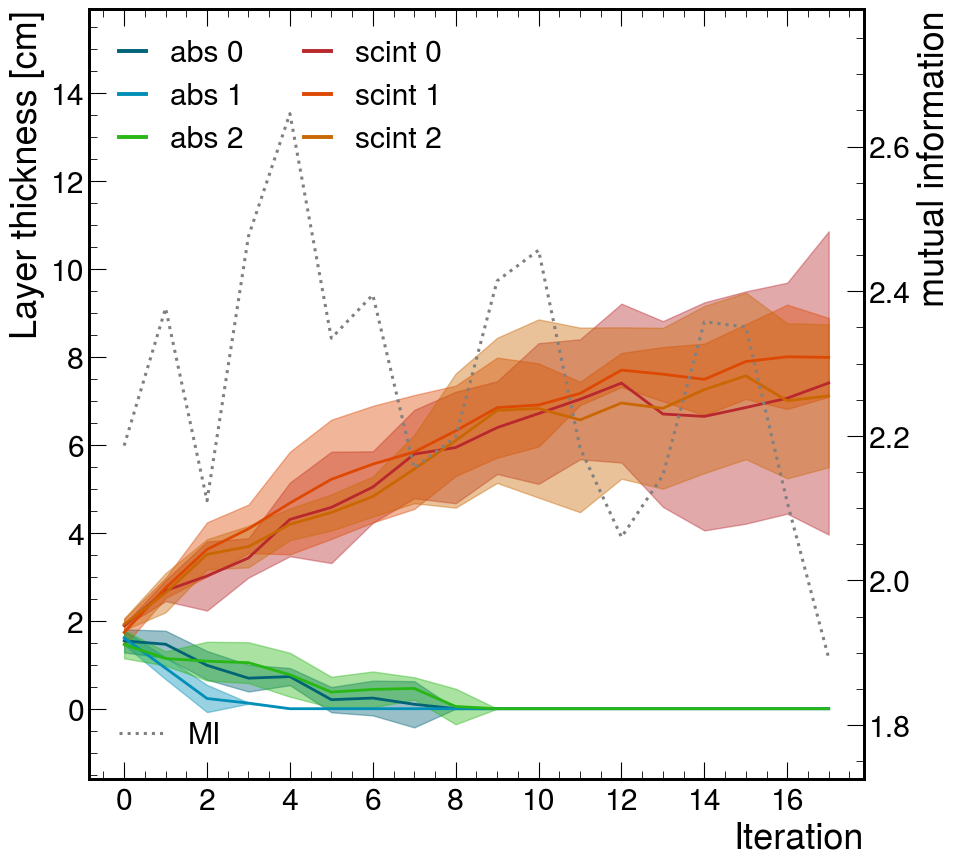}
        \caption{MI-OPT, 3 layers, no TL, 700 events}
        \label{fig:evo_transfer_layers_3_700_noTL_mi}
	\end{subfigure}%
	\hfilplus
	\begin{subfigure}{0.43\textwidth}
	    \centering
		\includegraphics[width=1.01\textwidth]{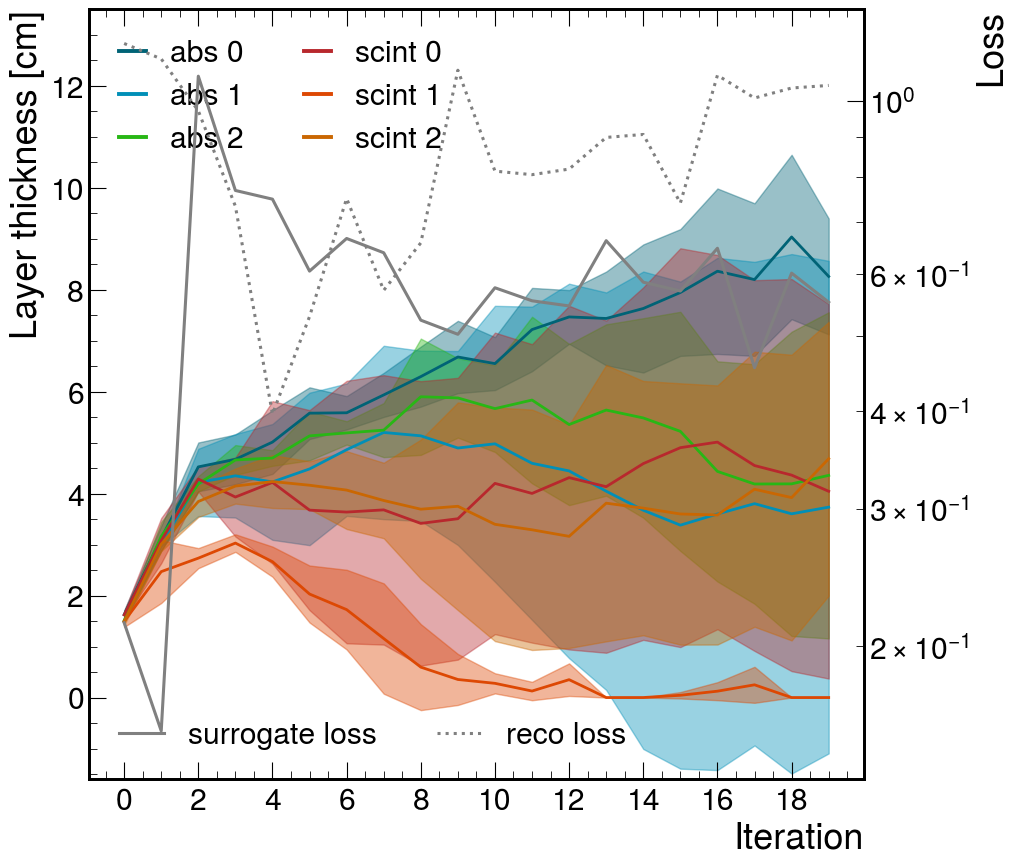}
        \caption{RECO-OPT, 3 layers, no TL, 700 events}
        \label{fig:evo_transfer_layers3_700_noTL_reco}
	\end{subfigure}%
\caption{TRANSFER study (700): Re-initialized learning (no transfer) with full sample size and three layers. Comparison of thickness evolutions for MI-OPT (left) and RECO-OPT (right) approaches.\label{fig:evo_transfer_notl_700ev_3layer}}
\end{figure*}

The impact of knowledge transfer with \textbf{\ac{TL}} is illustrated in Figures \ref{fig:evo_transfer_notl_700ev_1layer} and \ref{fig:evo_transfer_notl_700ev_3layer}, which present results for one- and three-layered calorimeters, respectively. 
Unlike the base studies discussed earlier, this experiment involves reinitializing the surrogate models (along with the reconstruction model in the case of RECO-OPT) at each iteration. Consequently, these models are exclusively trained on data in a local area of the parameter space, akin to the approach in \cite{black_box_NEURIPS2020_a878dbeb}.

A key observation is the increased fluctuation in the objective function (depicted by the grey dotted/solid lines). This behavior is expected, as the surrogates cannot accumulate knowledge to construct a smooth mapping of the $\theta-\delta$ landscape.
In RECO-OPT, the losses stagnate or even increase, indicating a breakdown in the optimization process. 

Without \ac{TL}, RECO-OPT shows significantly increased variation in the evolution of one-layer studies and fails to converge to a solution for the three-layer problem (individual run plots are shown in the Appendix~\ref{appendix:notl_failure}). 
The deviation from results reported by~\cite{black_box_NEURIPS2020_a878dbeb} are attributed to the use of a fixed metric function which might contribute to stability in that study. \\ 
For MI-OPT, the effect of turning off \ac{TL} leads to stagnation: the scintillator thickness plateaus at a lower value compared to the case where \ac{TL} is enabled. For example, in the single-layer scenario (Figure \ref{fig:evo_transfer_layers1_700_noTL_mi}), the scintillator thickness remains below \SI{10}{cm} and absorber compression requires more iterations. \\
These findings highlight the effectiveness of \ac{TL} in stabilizing the optimization process and guiding the search towards more optimal configurations, corroborating our hypothesis that a local surrogate can map out consecutive patches of the design parameter space in our framework setup. \\

\begin{figure*}
    \centering
	\begin{subfigure}{0.41\textwidth}
		\includegraphics[width=\textwidth]{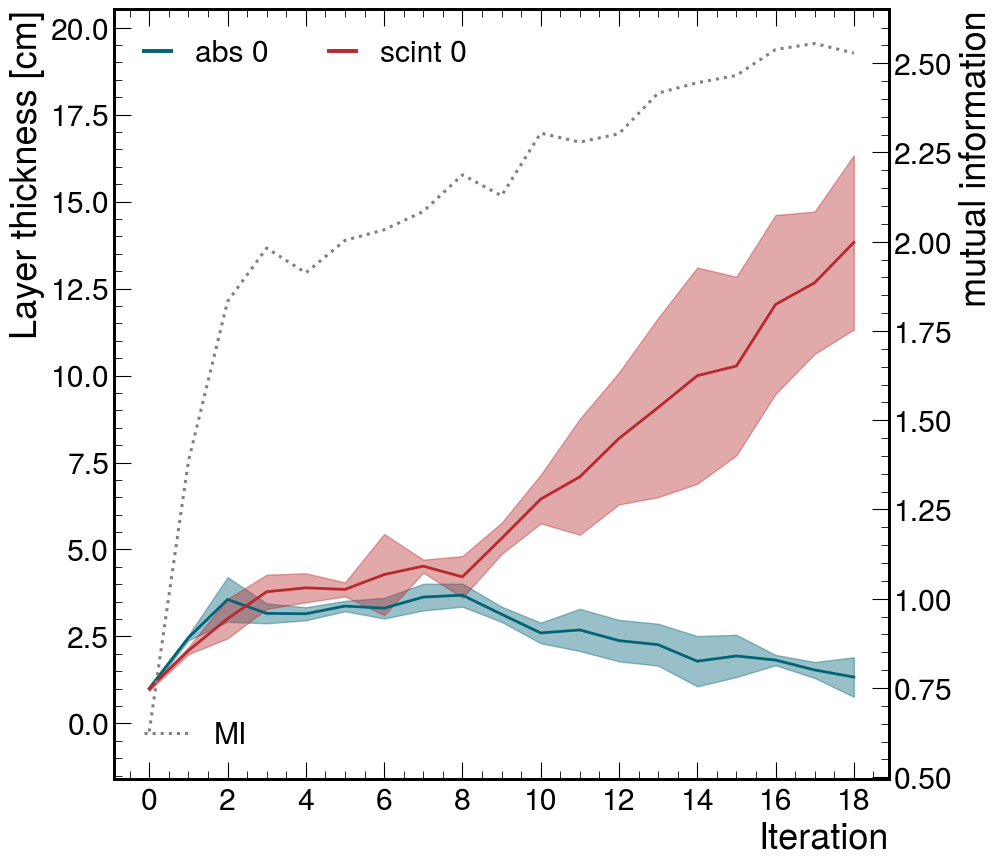}
        \caption{MI-OPT, 1 layer, TL, 50 events}
        \label{fig:tl_mi_50_1L}
	\end{subfigure}%
    \hfilplus
	\begin{subfigure}{0.41\textwidth}
		\includegraphics[width=0.975\textwidth]{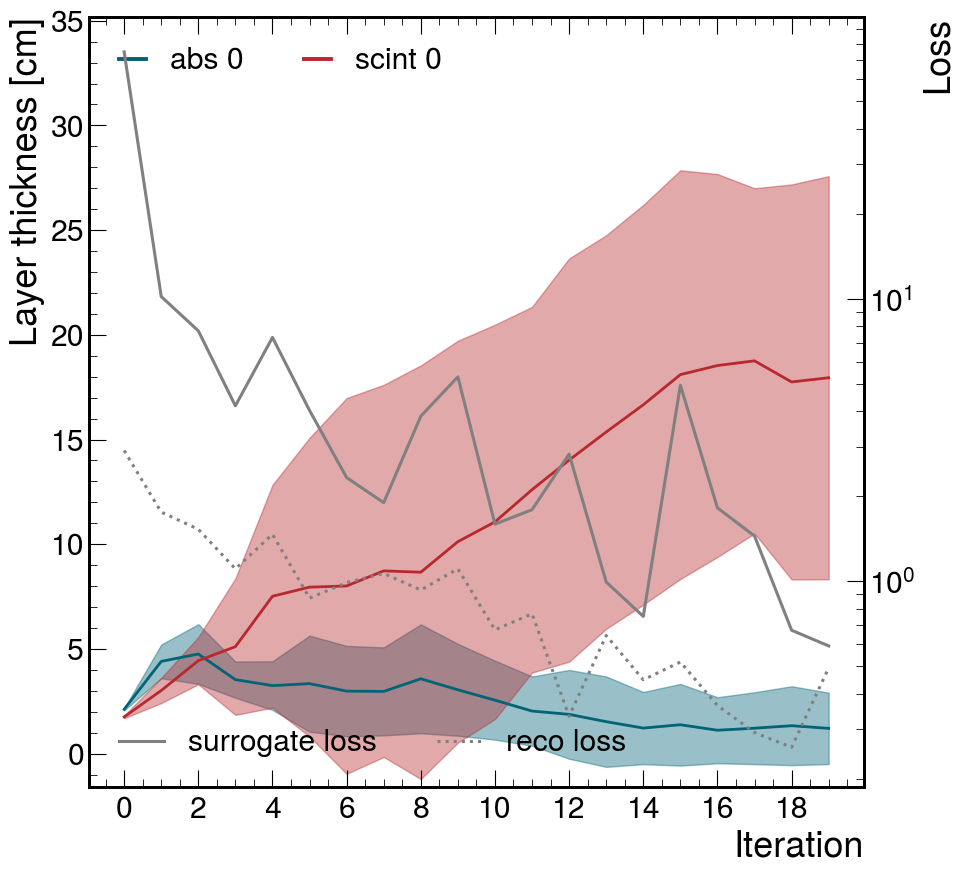}
        \caption{RECO-OPT, 1 layer, TL, 50 events}
        \label{fig:tl_reco_50_1L}
	\end{subfigure}
	\begin{subfigure}{0.41\textwidth}
		\includegraphics[width=\textwidth]{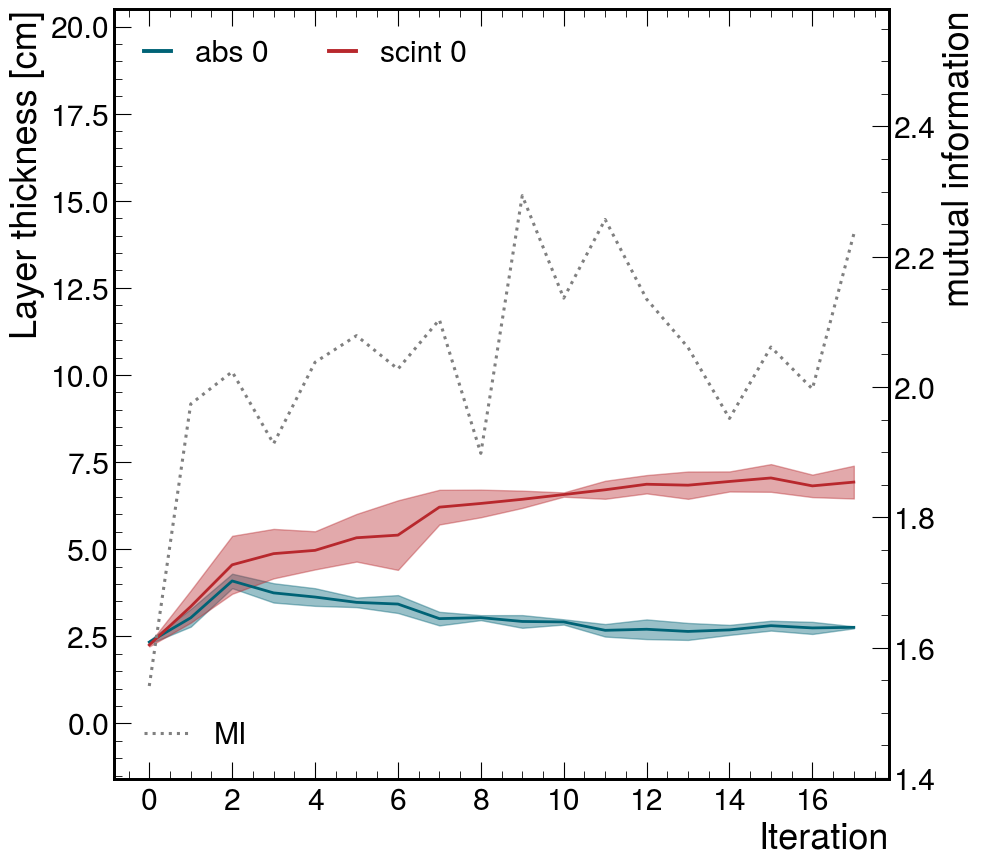}
        \caption{MI-OPT, 1 layer, no TL, 50 events}
        \label{fig:notl_mi_50}
	\end{subfigure}%
    \hfilplus
	\begin{subfigure}{0.41\textwidth}
		\includegraphics[width=0.975\textwidth]{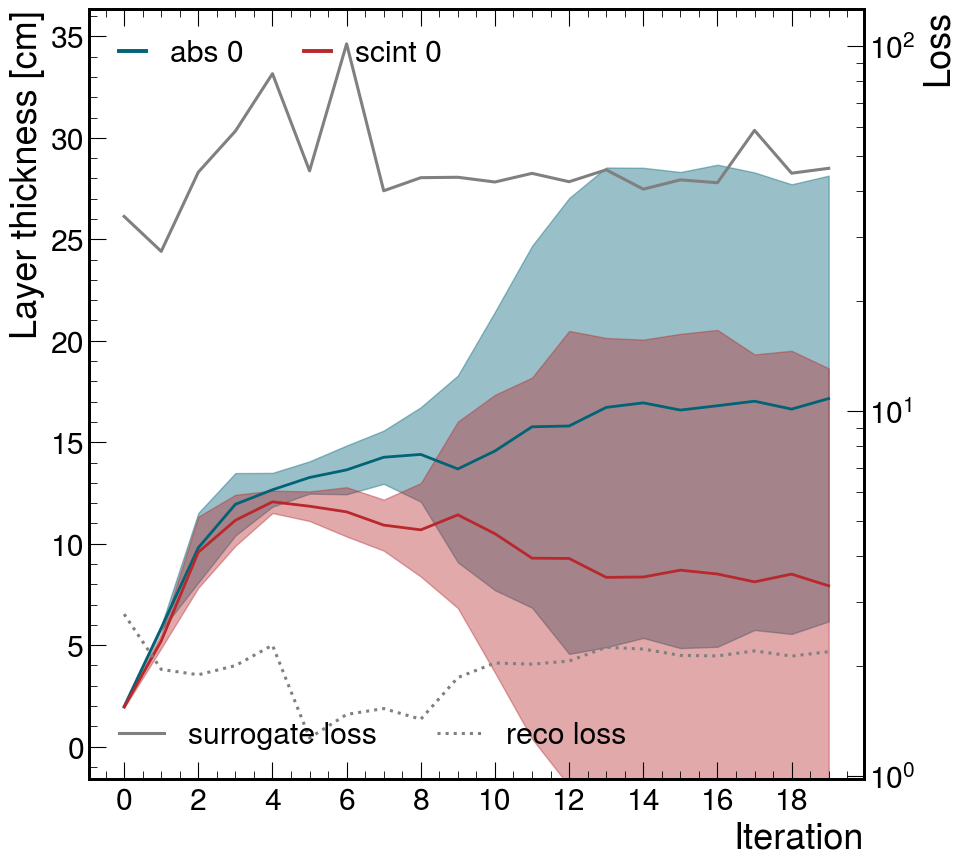}
        \caption{RECO-OPT, 1 layer, no TL, 50 events}
        \label{fig:notl_reco_50_1L}
	\end{subfigure}
\caption{TRANSFER study (50): Transfer learning studies with reduced sample size and one layer. Comparison of optimization evolutions with (top row) and without (bottom row) \ac{TL} for mutual information (left column) and reconstruction (right column) approaches, with $50$ events per tested design parameter set.}
\label{fig:evo_transfer_50ev_1layer}
\end{figure*}

\begin{figure*}
\centering
	\begin{subfigure}{0.41\textwidth}
		\includegraphics[width=0.985\textwidth]{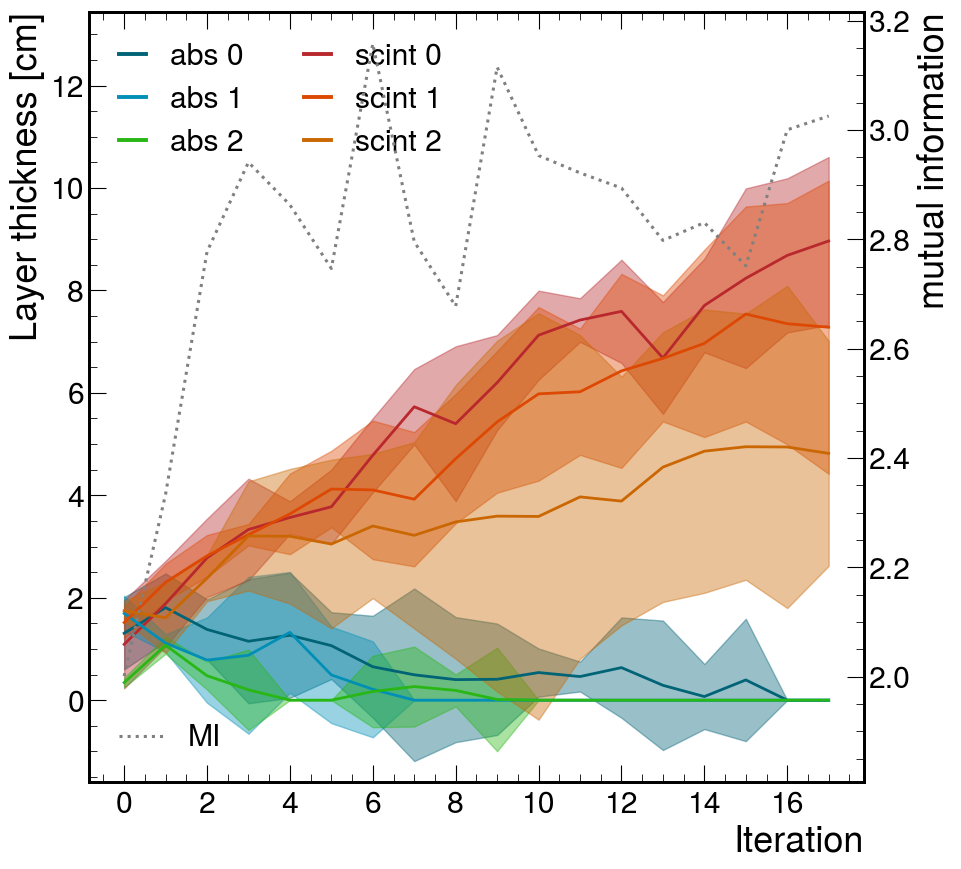}
        \caption{MI-OPT, 3 layers, TL, 50 events}
        \label{fig:tl_50_3L_mi}
	\end{subfigure}%
        \hfilplus
	\begin{subfigure}{0.41\textwidth}
		\includegraphics[width=\textwidth]{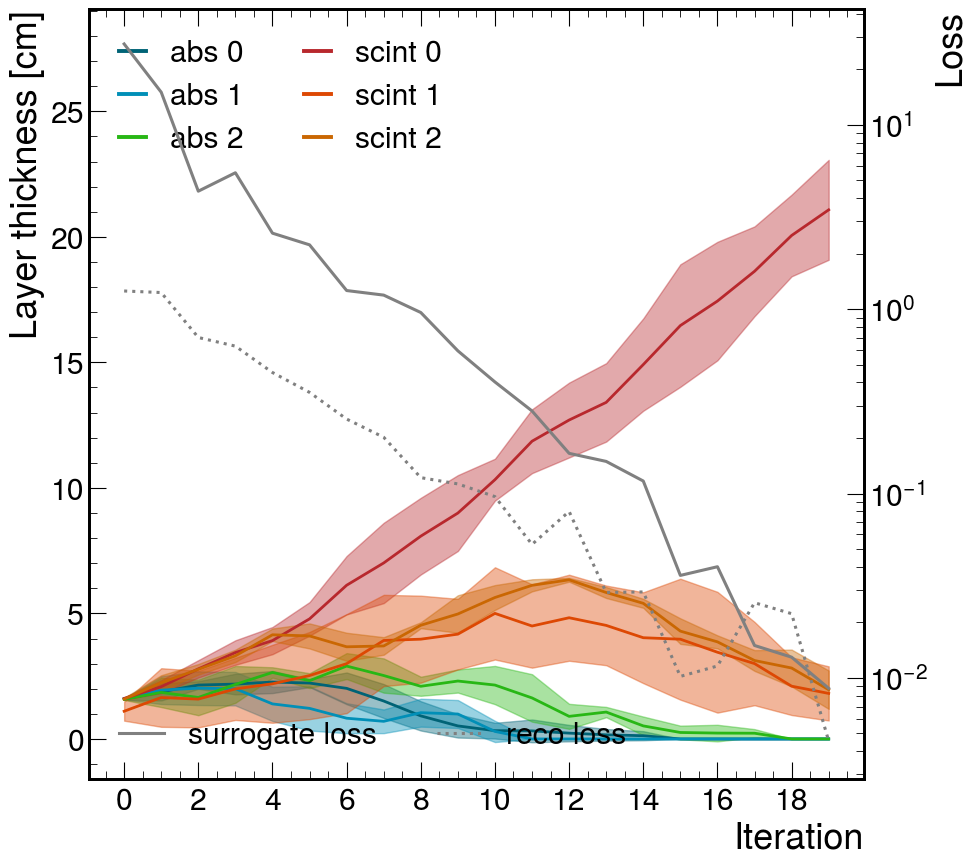}
            \caption{RECO-OPT, 3 layers, TL, 50 events}
            \label{fig:tl_mi_50_3L}
	\end{subfigure}
	\begin{subfigure}{0.41\textwidth}
		\includegraphics[width=0.985\textwidth]{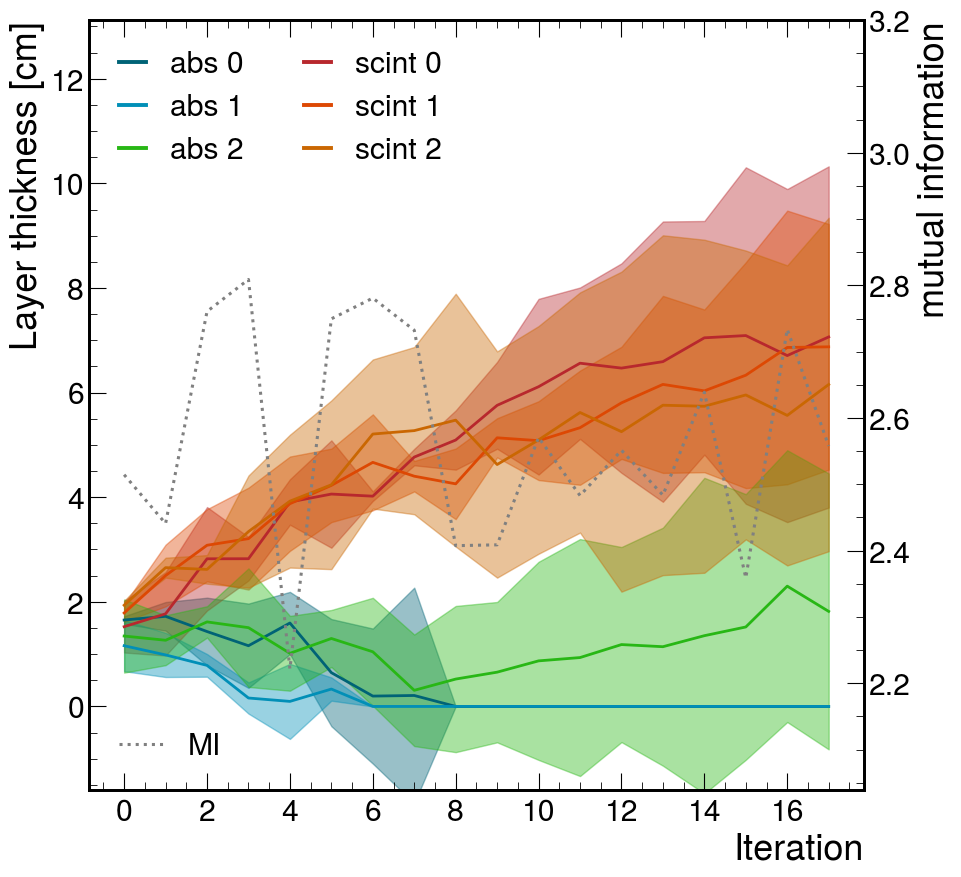}
            \caption{MI-OPT, 3 layers, no TL, 50 events}
            \label{fig:notl_reco_50_3L}
	\end{subfigure}%
	\hfilplus
        \begin{subfigure}{0.41\textwidth}
		\includegraphics[width=\textwidth]{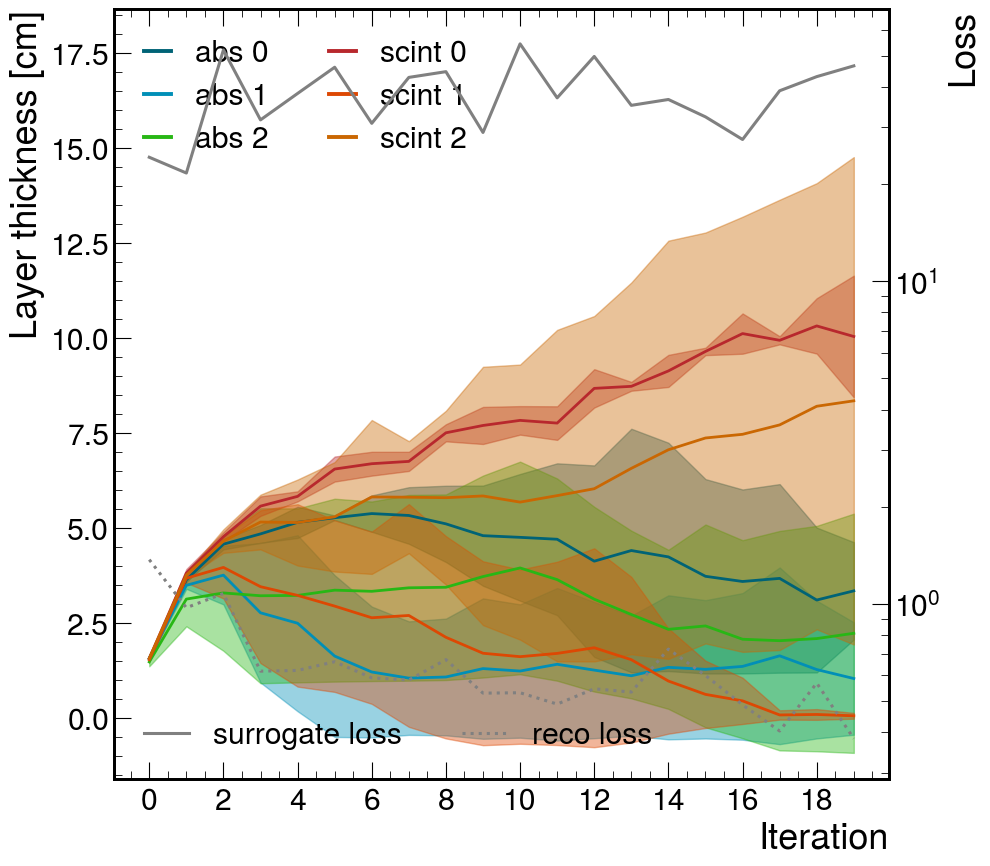}
            \caption{RECO-OPT, 3 layers, no TL, 50 events}
            \label{fig:tl_reco_50}
	\end{subfigure}
	\caption{TRANSFER study (50): Transfer learning studies with reduced sample size and three layers. Comparison of optimization evolutions with (top row) and without (bottom row) \ac{TL} for mutual information (left column) and reconstruction (right column) approaches, with 50 events per parameter candidate.\label{fig:evo_transfer_50ev_3layer}}
\end{figure*}

To assess the relationship between \ac{TL} and data sample size, we conducted experiments using just 50 simulated collision events per design parameter set. Note that the parameter neighborhood is sampled more densely in MI-OPT ($K=117$) than in RECO-OPT ($K=30$), such that the latter has access to a systematically smaller number of total samples. Preliminary studies were conducted showing that MI-OPT produces similar results with $K=50$, but in general, the information-theoretic metric relies more on adequate coverage of the true distributions of particle and sensor energies than the classic reconstruction approach. \\
We observe that, even with this limited data, both optimization approaches effectively increase scintillator thickness and reduce absorber thickness in a single-layer (Figure~\ref{fig:evo_transfer_50ev_1layer}) and three-layer (Figure~\ref{fig:evo_transfer_50ev_3layer}) calorimeter when \ac{TL} is activated.

The results are similar to the ones obtained with 700 events: For MI-OPT \ac{TL} not only yields a much smoother mutual information metric but also leads to a different optimization trajectory. Both thickness values stagnate at a single-digit level when \ac{TL} is not available.
For RECO-OPT, we observe again a breakdown of optimization for the no-\ac{TL} runs and a slower, but viable convergence when \ac{TL} is present.

\begin{figure*}
    \centering
    \begin{subfigure}{0.41\textwidth}
        \includegraphics[width=\textwidth]{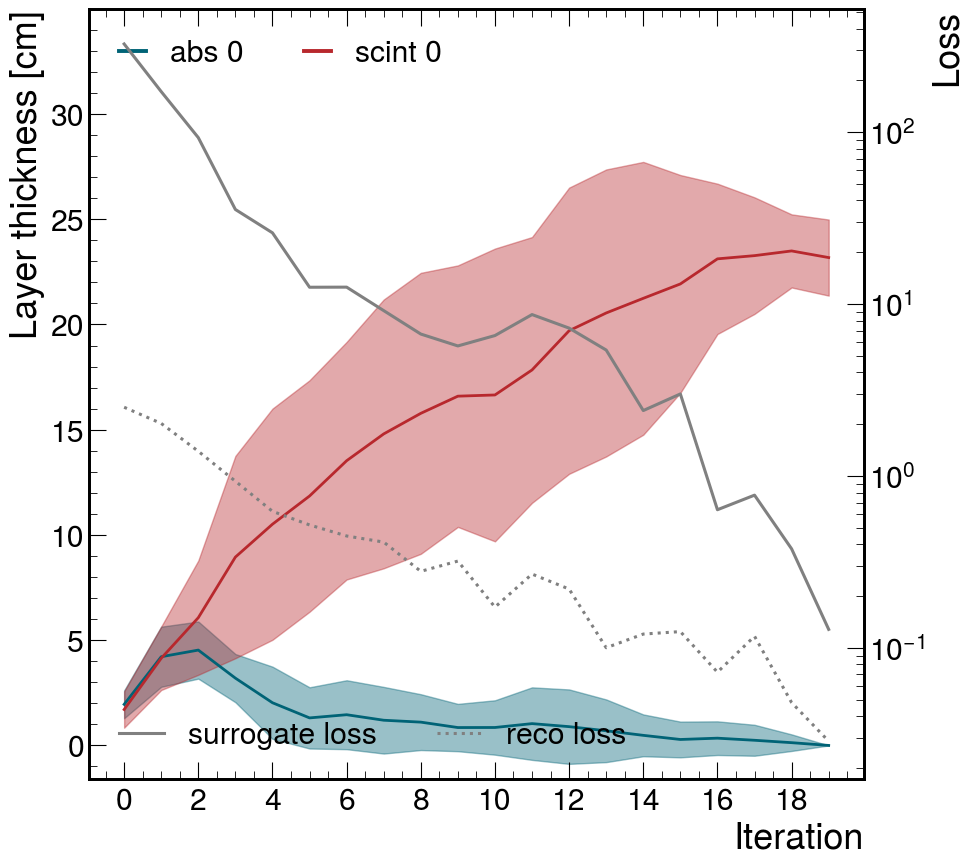}
        \caption{RECO-OPT with \ac{TL}, 5 events}
        \label{fig:tl_reco_5}
    \end{subfigure}%
    \hfilplus
    \begin{subfigure}{0.41\textwidth}
        \includegraphics[width=\textwidth]{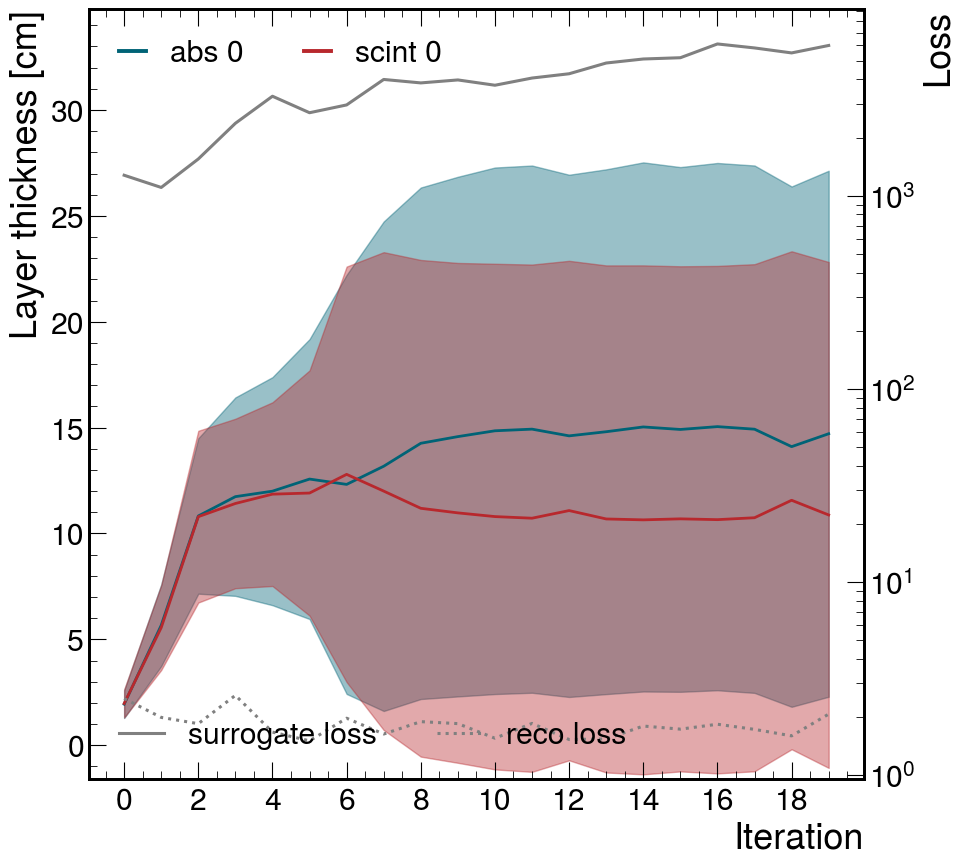}
        \caption{RECO-OPT without \ac{TL}, 5 events}
        \label{fig:notl_reco_5}
    \end{subfigure}
\caption{TRANSFER study (5): Transfer learning studies with reduced sample size and one layer. Comparison of optimization evolutions with (left) and without (right) \ac{TL} for reconstruction metric, with $5$ events per tested design parameter set. Averaged over 10 runs.}
\label{fig:tl_5_events}
\end{figure*}

With RECO-OPT successfully converging with 50 events (Figure~\ref{fig:evo_transfer_50ev_1layer}), we further reduce the event count to~5 (Figure~\ref{fig:tl_5_events}), where optimization remains effective with \ac{TL} but fails without it.

It is notable how well the framework performs with \ac{TL} at such low event numbers. 
Even when simulating $\mathcal{O}(10^1)$ and $\mathcal{O}(10^2)$ less events per variable the framework is still able to consistently converge to a similar final state as the base studies (Figure~\ref{fig:evo_base}). 
Overall, these findings highlight the threshold at which sample size reduction begins to impede optimization performance and delineate the limits of \ac{TL}. These observations have significant practical implications when extending the approaches to more complex, and slower simulation. \\


\begin{figure*}
    \begin{subfigure}{0.41\textwidth}
        \centering
        \includegraphics[width=\textwidth]{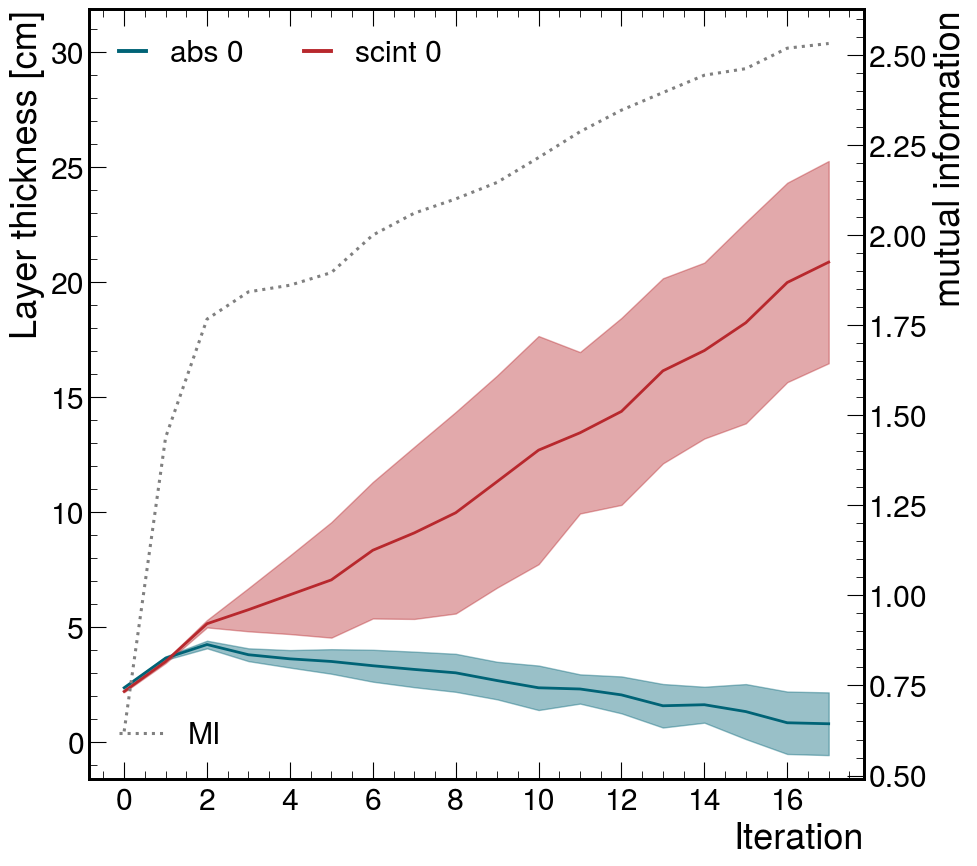}
        \caption{MI-OPT, 1 layer, 700 events, \qtyrange[range-units=single,range-phrase=-]{1}{100}{\GeV}}
        \label{fig:results_energy_mi_1L}
    \end{subfigure}
    \hfilplus
    \begin{subfigure}{0.41\textwidth}
        \centering
        \includegraphics[width=0.99\textwidth]{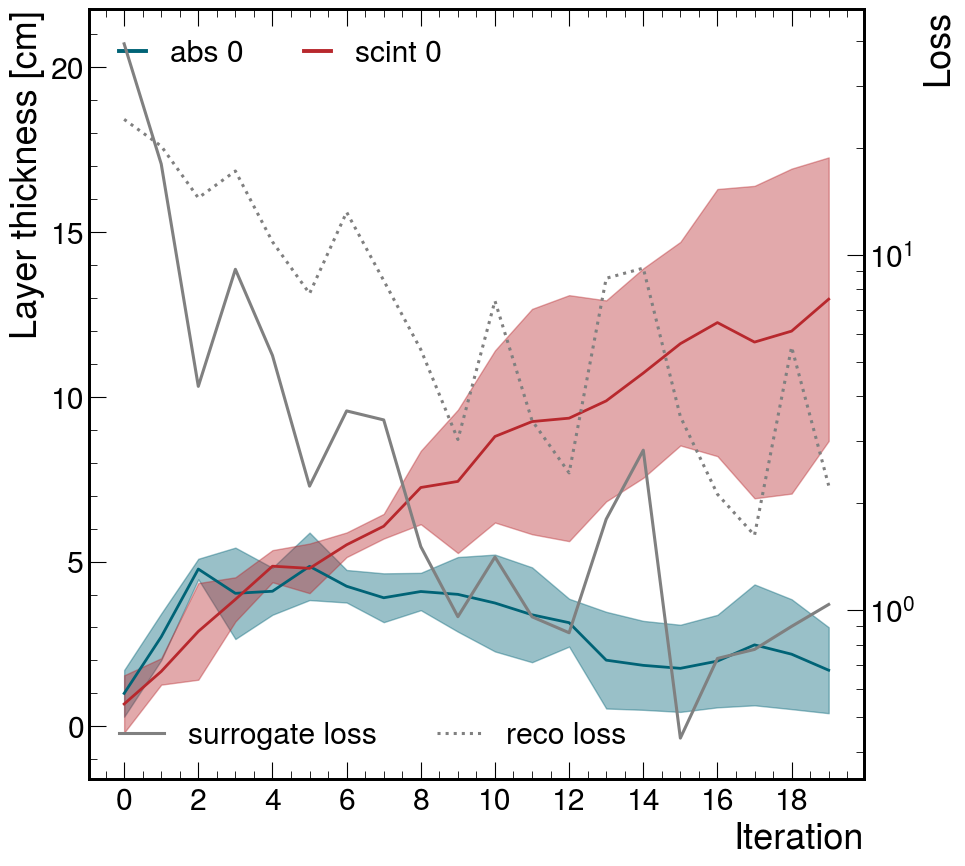}
        \caption{RECO-OPT, 1 layer, 700 events, \qtyrange[range-units=single,range-phrase=-]{1}{100}{\GeV}}
        \label{fig:results_energy_reco_1L}
    \end{subfigure}
    \begin{subfigure}{0.41\textwidth}
        \centering
        \includegraphics[width=\textwidth]{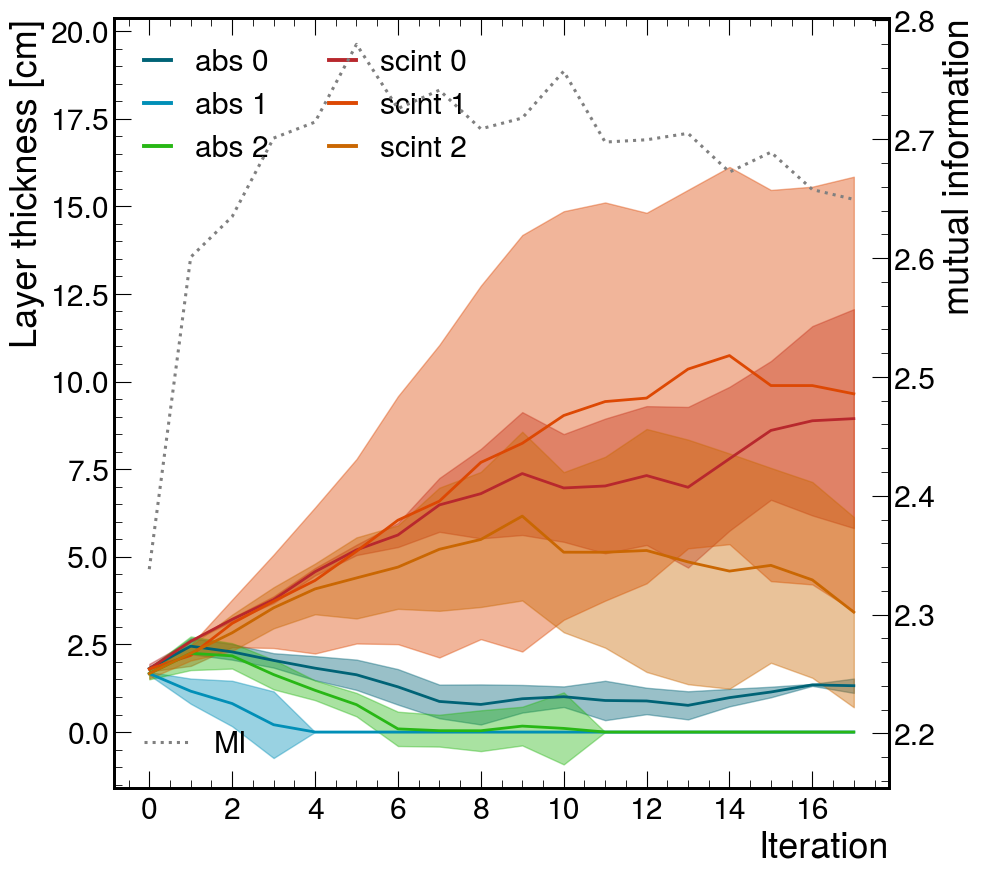}
        \caption{MI-OPT, 3 layers, 700 events, \qtyrange[range-units=single,range-phrase=-]{1}{100}{\GeV}}
        \label{fig:results_energy_mi_3L}
    \end{subfigure}
    \hfilplus
    \begin{subfigure}{0.41\textwidth}
        \centering
        \includegraphics[width=\textwidth]{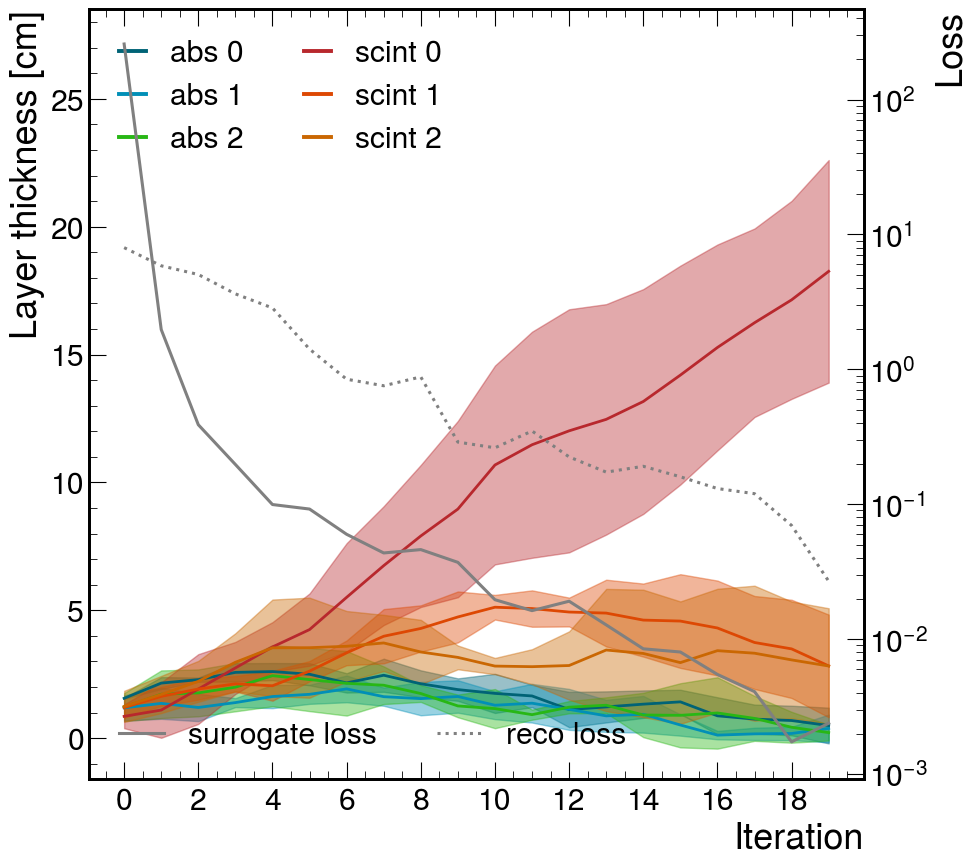}
        \caption{RECO-OPT, 3 layers, 700 events, \qtyrange[range-units=single,range-phrase=-]{1}{100}{\GeV}}
        \label{fig:results_energy_reco_3L}
    \end{subfigure}
    \caption{ENERGY study: Single and triple-paired calorimeter thickness evolution (solid lines) for MI-OPT (left, mutual information maximization shown by dotted line) and RECO-OPT (right, reconstruction loss minimization shown by dotted line), averaged over three runs for a photon with center of mass between \qtyrange[range-units=single,range-phrase=-]{1}{100}{\GeV}.}
    \label{fig:evo_energy}
\end{figure*}

\subsection{Energy Studies}

The results for the \textbf{energy studies}, where the energy range of the photon is increased to \SI{100}{\GeV}, are shown in Figure \ref{fig:evo_energy}.

We observe, that the absorber thickness, both in terms of its average value and variation bounds, remains consistently above that of the baseline scenario. For both algorithms and both layer settings, at least one absorber remains consistently above zero gauge. This outcome is expected, as a greater calorimeter thickness is necessary to effectively capture and measure the trajectories of higher-energy particles. Furthermore, the variability in scintillator thickness is consistently greater than in the baseline studies conducted with photon energies up to \SI{20}{\GeV}. Also, notably, in the case of MI-OPT, the prioritization of layers is reversed, with the second scintillator exhibiting the most significant extension.

With increased energy, RECO-OPT continues to focus on increasing the width of the first scintillator primarily, though evolution proceeds at a lower rate. 
The scintillator thickness does not reach the \SI{25}{cm} constraint for neither the one nor three layer cases (Figures \ref{fig:results_energy_reco_1L} and~\ref{fig:results_energy_reco_3L}), though it is trending towards that. 
With a significant number of higher energy events, a significantly lower fraction of the total energy of a given event is deposited in the calorimeter scintillators, making it difficult for the reconstruction network to learn and the surrogate as a whole to optimize the detector parameters. 

\section{Discussion}\label{sec:conclusio}
In this work, we presented a \ac{DL}-based local surrogate framework for optimizing \ac{HEP} detector designs using a scalar objective function. The optimization encompasses the forward as well as the inverse detector problem, yielding an end-to-end result. 

Alongside a traditional reconstruction-based approach, we introduced a novel method leveraging the information-theoretic metric of mutual information as an optimization measure.

To evaluate the two frameworks, we conducted studies on a toy calorimeter model with one, two, and three layers, optimizing the thicknesses of absorber and scintillator components. Our results demonstrate that both approaches successfully optimize detector parameters, increasing scintillator thickness while reducing absorber thickness. This confirms that end-to-end, black-box \ac{DL}-based optimization is feasible with only a single scalar objective function. In the case of our mutual information based optimization MI-OPT, this scalar objective function does not have to be specified explicitly, thus making it task-agnostic and user independent.

Our findings also support the effectiveness of progressive learning with \ac{TL} in this context. The surrogate model accumulates knowledge about the design parameter space and the objective function landscape over the course of optimization. This is particularly evident in reduced data-sample settings, where strong optimization performance is achieved with as few as 50 events per parameter candidate for both frameworks, and 5 events per parameter candidate for the standard optimization RECO-OPT. In contrast, without \ac{TL}, optimization results remain suboptimal or the optimization process completely fails, highlighting the benefits of leveraging prior knowledge in iterative design processes. The ability to achieve comparable optimization performance for orders of magnitude fewer simulation calls is desirable as the simulation step is often a bottleneck for optimization tasks.

For future work, incorporating both energy resolution and particle classification accuracy as an additional objective, the framework could optimize detector designs for multi-objective tasks, reflecting the real-world challenges of \ac{HEP} detector development.

Our findings show that mutual information produces design choices closely aligned with state-of-the-art physics-informed methods, suggesting these approaches operate near optimality. Beyond validating existing methods, mutual information enables broader exploration and generalization of the optimization process, as it is inherently problem-agnostic, though at the cost of increased computational demands. By shifting the focus from domain-specific heuristics to an information-theoretic formulation, this approach opens new avenues for exploring design spaces beyond conventional methodologies.

\section*{Acknowledgements}
We thank Atul Sinha and Yusong Tian for their contributions to this project. JK is supported by the Alexander-von-Humboldt foundation.

\bibliography{references}

\begin{thebibliography}{58}%
\makeatletter
\providecommand \@ifxundefined [1]{%
 \@ifx{#1\undefined}
}%
\providecommand \@ifnum [1]{%
 \ifnum #1\expandafter \@firstoftwo
 \else \expandafter \@secondoftwo
 \fi
}%
\providecommand \@ifx [1]{%
 \ifx #1\expandafter \@firstoftwo
 \else \expandafter \@secondoftwo
 \fi
}%
\providecommand \natexlab [1]{#1}%
\providecommand \enquote  [1]{``#1''}%
\providecommand \bibnamefont  [1]{#1}%
\providecommand \bibfnamefont [1]{#1}%
\providecommand \citenamefont [1]{#1}%
\providecommand \href@noop [0]{\@secondoftwo}%
\providecommand \href [0]{\begingroup \@sanitize@url \@href}%
\providecommand \@href[1]{\@@startlink{#1}\@@href}%
\providecommand \@@href[1]{\endgroup#1\@@endlink}%
\providecommand \@sanitize@url [0]{\catcode `\\12\catcode `\$12\catcode `\&12\catcode `\#12\catcode `\^12\catcode `\_12\catcode `\%12\relax}%
\providecommand \@@startlink[1]{}%
\providecommand \@@endlink[0]{}%
\providecommand \url  [0]{\begingroup\@sanitize@url \@url }%
\providecommand \@url [1]{\endgroup\@href {#1}{\urlprefix }}%
\providecommand \urlprefix  [0]{URL }%
\providecommand \Eprint [0]{\href }%
\providecommand \doibase [0]{https://doi.org/}%
\providecommand \selectlanguage [0]{\@gobble}%
\providecommand \bibinfo  [0]{\@secondoftwo}%
\providecommand \bibfield  [0]{\@secondoftwo}%
\providecommand \translation [1]{[#1]}%
\providecommand \BibitemOpen [0]{}%
\providecommand \bibitemStop [0]{}%
\providecommand \bibitemNoStop [0]{.\EOS\space}%
\providecommand \EOS [0]{\spacefactor3000\relax}%
\providecommand \BibitemShut  [1]{\csname bibitem#1\endcsname}%
\let\auto@bib@innerbib\@empty
\bibitem [{\citenamefont {Apollinari}\ \emph {et~al.}(2017)\citenamefont {Apollinari} \emph {et~al.}}]{hl_lhc_osti_1767028}%
  \BibitemOpen
  \bibfield  {author} {\bibinfo {author} {\bibfnamefont {G.}~\bibnamefont {Apollinari}} \emph {et~al.},\ }\href {https://doi.org/10.23731/CYRM-2017-004} {\emph {\bibinfo {title} {High-Luminosity Large Hadron Collider (HL-LHC). Technical Design Report V. 0.1}}},\ \bibinfo {type} {Tech. Rep.}\ (\bibinfo  {institution} {Fermi National Accelerator Lab. (FNAL), Batavia, IL (United States)},\ \bibinfo {year} {2017})\BibitemShut {NoStop}%
\bibitem [{\citenamefont {{The FCC Collaboration}}(2019)}]{fcc_hh_7060500e92d544b297d50ba27e51dc95}%
  \BibitemOpen
  \bibfield  {author} {\bibinfo {author} {\bibnamefont {{The FCC Collaboration}}},\ }\bibfield  {title} {\bibinfo {title} {Fcc-hh: The hadron collider: Future circular collider conceptual design report volume 3},\ }\href {https://doi.org/10.1140/epjst/e2019-900087-0} {\bibfield  {journal} {\bibinfo  {journal} {European Physical Journal Special Topics}\ }\textbf {\bibinfo {volume} {228}},\ \bibinfo {pages} {755} (\bibinfo {year} {2019})}\BibitemShut {NoStop}%
\bibitem [{\citenamefont {Jones}\ \emph {et~al.}(1998)\citenamefont {Jones}, \citenamefont {Schonlau},\ and\ \citenamefont {Welch}}]{Jones1998EfficientGO}%
  \BibitemOpen
  \bibfield  {author} {\bibinfo {author} {\bibfnamefont {D.~R.}\ \bibnamefont {Jones}}, \bibinfo {author} {\bibfnamefont {M.}~\bibnamefont {Schonlau}},\ and\ \bibinfo {author} {\bibfnamefont {W.~J.}\ \bibnamefont {Welch}},\ }\bibfield  {title} {\bibinfo {title} {Efficient global optimization of expensive black-box functions},\ }\href {https://api.semanticscholar.org/CorpusID:263864014} {\bibfield  {journal} {\bibinfo  {journal} {Journal of Global Optimization}\ }\textbf {\bibinfo {volume} {13}},\ \bibinfo {pages} {455} (\bibinfo {year} {1998})}\BibitemShut {NoStop}%
\bibitem [{\citenamefont {Dorigo}\ \emph {et~al.}(2025)\citenamefont {Dorigo}, \citenamefont {Doro}, \citenamefont {Aehle}, \citenamefont {Gauger}, \citenamefont {Awais}, \citenamefont {Izbicki}, \citenamefont {Kieseler}, \citenamefont {Masserano}, \citenamefont {Nardi},\ and\ \citenamefont {Vergara}}]{dorigo2025utilityfunctionexperimentsfundamental}%
  \BibitemOpen
  \bibfield  {author} {\bibinfo {author} {\bibfnamefont {T.}~\bibnamefont {Dorigo}}, \bibinfo {author} {\bibfnamefont {M.}~\bibnamefont {Doro}}, \bibinfo {author} {\bibfnamefont {M.}~\bibnamefont {Aehle}}, \bibinfo {author} {\bibfnamefont {N.~R.}\ \bibnamefont {Gauger}}, \bibinfo {author} {\bibfnamefont {M.}~\bibnamefont {Awais}}, \bibinfo {author} {\bibfnamefont {R.}~\bibnamefont {Izbicki}}, \bibinfo {author} {\bibfnamefont {J.}~\bibnamefont {Kieseler}}, \bibinfo {author} {\bibfnamefont {L.}~\bibnamefont {Masserano}}, \bibinfo {author} {\bibfnamefont {F.}~\bibnamefont {Nardi}},\ and\ \bibinfo {author} {\bibfnamefont {L.~R.}\ \bibnamefont {Vergara}},\ }\href {https://arxiv.org/abs/2501.13544} {\bibinfo {title} {On the utility function of experiments in fundamental science}} (\bibinfo {year} {2025}),\ \Eprint {https://arxiv.org/abs/2501.13544} {arXiv:2501.13544 [hep-ex]} \BibitemShut {NoStop}%
\bibitem [{\citenamefont {Aehle}\ \emph {et~al.}(2024{\natexlab{a}})\citenamefont {Aehle}, \citenamefont {Nguyen}, \citenamefont {Novák}, \citenamefont {Dorigo}, \citenamefont {Gauger}, \citenamefont {Kieseler}, \citenamefont {Klute},\ and\ \citenamefont {Vassilev}}]{aehle2024efficientforwardmodealgorithmicderivatives}%
  \BibitemOpen
  \bibfield  {author} {\bibinfo {author} {\bibfnamefont {M.}~\bibnamefont {Aehle}}, \bibinfo {author} {\bibfnamefont {X.~T.}\ \bibnamefont {Nguyen}}, \bibinfo {author} {\bibfnamefont {M.}~\bibnamefont {Novák}}, \bibinfo {author} {\bibfnamefont {T.}~\bibnamefont {Dorigo}}, \bibinfo {author} {\bibfnamefont {N.~R.}\ \bibnamefont {Gauger}}, \bibinfo {author} {\bibfnamefont {J.}~\bibnamefont {Kieseler}}, \bibinfo {author} {\bibfnamefont {M.}~\bibnamefont {Klute}},\ and\ \bibinfo {author} {\bibfnamefont {V.}~\bibnamefont {Vassilev}},\ }\href {https://arxiv.org/abs/2407.02966} {\bibinfo {title} {Efficient forward-mode algorithmic derivatives of geant4}} (\bibinfo {year} {2024}{\natexlab{a}}),\ \Eprint {https://arxiv.org/abs/2407.02966} {arXiv:2407.02966 [physics.comp-ph]} \BibitemShut {NoStop}%
\bibitem [{\citenamefont {Aehle}\ \emph {et~al.}(2024{\natexlab{b}})\citenamefont {Aehle}, \citenamefont {Novák}, \citenamefont {Vassilev}, \citenamefont {Gauger}, \citenamefont {Heinrich}, \citenamefont {Kagan},\ and\ \citenamefont {Lange}}]{aehle2024optimizationusingpathwisealgorithmic}%
  \BibitemOpen
  \bibfield  {author} {\bibinfo {author} {\bibfnamefont {M.}~\bibnamefont {Aehle}}, \bibinfo {author} {\bibfnamefont {M.}~\bibnamefont {Novák}}, \bibinfo {author} {\bibfnamefont {V.}~\bibnamefont {Vassilev}}, \bibinfo {author} {\bibfnamefont {N.~R.}\ \bibnamefont {Gauger}}, \bibinfo {author} {\bibfnamefont {L.}~\bibnamefont {Heinrich}}, \bibinfo {author} {\bibfnamefont {M.}~\bibnamefont {Kagan}},\ and\ \bibinfo {author} {\bibfnamefont {D.}~\bibnamefont {Lange}},\ }\href {https://arxiv.org/abs/2405.07944} {\bibinfo {title} {Optimization using pathwise algorithmic derivatives of electromagnetic shower simulations}} (\bibinfo {year} {2024}{\natexlab{b}}),\ \Eprint {https://arxiv.org/abs/2405.07944} {arXiv:2405.07944 [physics.comp-ph]} \BibitemShut {NoStop}%
\bibitem [{\citenamefont {Kagan}\ and\ \citenamefont {Heinrich}(2023)}]{kagan2023branchestreetakingderivatives}%
  \BibitemOpen
  \bibfield  {author} {\bibinfo {author} {\bibfnamefont {M.}~\bibnamefont {Kagan}}\ and\ \bibinfo {author} {\bibfnamefont {L.}~\bibnamefont {Heinrich}},\ }\href {https://arxiv.org/abs/2308.16680} {\bibinfo {title} {Branches of a tree: Taking derivatives of programs with discrete and branching randomness in high energy physics}} (\bibinfo {year} {2023}),\ \Eprint {https://arxiv.org/abs/2308.16680} {arXiv:2308.16680 [stat.ML]} \BibitemShut {NoStop}%
\bibitem [{\citenamefont {Dorigo}\ \emph {et~al.}(2024)\citenamefont {Dorigo}, \citenamefont {Aehle}, \citenamefont {Arcaro}, \citenamefont {Awais}, \citenamefont {Bergamaschi}, \citenamefont {Donini}, \citenamefont {Doro}, \citenamefont {Gauger}, \citenamefont {Izbicki}, \citenamefont {Kieseler}, \citenamefont {Lee}, \citenamefont {Masserano}, \citenamefont {Nardi}, \citenamefont {Rajesh}, \citenamefont {Vergara},\ and\ \citenamefont {Shen}}]{dorigo2024endtoendoptimizationswgoarray}%
  \BibitemOpen
  \bibfield  {author} {\bibinfo {author} {\bibfnamefont {T.}~\bibnamefont {Dorigo}}, \bibinfo {author} {\bibfnamefont {M.}~\bibnamefont {Aehle}}, \bibinfo {author} {\bibfnamefont {C.}~\bibnamefont {Arcaro}}, \bibinfo {author} {\bibfnamefont {M.}~\bibnamefont {Awais}}, \bibinfo {author} {\bibfnamefont {F.}~\bibnamefont {Bergamaschi}}, \bibinfo {author} {\bibfnamefont {J.}~\bibnamefont {Donini}}, \bibinfo {author} {\bibfnamefont {M.}~\bibnamefont {Doro}}, \bibinfo {author} {\bibfnamefont {N.~R.}\ \bibnamefont {Gauger}}, \bibinfo {author} {\bibfnamefont {R.}~\bibnamefont {Izbicki}}, \bibinfo {author} {\bibfnamefont {J.}~\bibnamefont {Kieseler}}, \bibinfo {author} {\bibfnamefont {A.}~\bibnamefont {Lee}}, \bibinfo {author} {\bibfnamefont {L.}~\bibnamefont {Masserano}}, \bibinfo {author} {\bibfnamefont {F.}~\bibnamefont {Nardi}}, \bibinfo {author} {\bibfnamefont {R.}~\bibnamefont {Rajesh}}, \bibinfo {author} {\bibfnamefont {L.~R.}\ \bibnamefont {Vergara}},\ and\ \bibinfo {author} {\bibfnamefont {A.}~\bibnamefont
  {Shen}},\ }\href {https://arxiv.org/abs/2310.01857} {\bibinfo {title} {Toward the end-to-end optimization of the swgo array layout}} (\bibinfo {year} {2024}),\ \Eprint {https://arxiv.org/abs/2310.01857} {arXiv:2310.01857 [astro-ph.IM]} \BibitemShut {NoStop}%
\bibitem [{\citenamefont {Baydin}\ \emph {et~al.}(2021)\citenamefont {Baydin} \emph {et~al.}}]{Baydin02012021}%
  \BibitemOpen
  \bibfield  {author} {\bibinfo {author} {\bibfnamefont {A.~G.}\ \bibnamefont {Baydin}} \emph {et~al.},\ }\bibfield  {title} {\bibinfo {title} {Toward machine learning optimization of experimental design},\ }\href {https://doi.org/10.1080/10619127.2021.1881364} {\bibfield  {journal} {\bibinfo  {journal} {Nuclear Physics News}\ }\textbf {\bibinfo {volume} {31}},\ \bibinfo {pages} {25} (\bibinfo {year} {2021})},\ \Eprint {https://arxiv.org/abs/https://doi.org/10.1080/10619127.2021.1881364} {https://doi.org/10.1080/10619127.2021.1881364} \BibitemShut {NoStop}%
\bibitem [{\citenamefont {Dorigo}\ \emph {et~al.}(2023)\citenamefont {Dorigo} \emph {et~al.}}]{DORIGO2023100085}%
  \BibitemOpen
  \bibfield  {author} {\bibinfo {author} {\bibfnamefont {T.}~\bibnamefont {Dorigo}} \emph {et~al.},\ }\bibfield  {title} {\bibinfo {title} {Toward the end-to-end optimization of particle physics instruments with differentiable programming},\ }\href {https://doi.org/https://doi.org/10.1016/j.revip.2023.100085} {\bibfield  {journal} {\bibinfo  {journal} {Reviews in Physics}\ }\textbf {\bibinfo {volume} {10}},\ \bibinfo {pages} {100085} (\bibinfo {year} {2023})}\BibitemShut {NoStop}%
\bibitem [{\citenamefont {Strong}\ \emph {et~al.}(2024)\citenamefont {Strong} \emph {et~al.}}]{Strong2024TomOpt}%
  \BibitemOpen
  \bibfield  {author} {\bibinfo {author} {\bibfnamefont {G.}~\bibnamefont {Strong}} \emph {et~al.},\ }\bibfield  {title} {\bibinfo {title} {Tomopt: differential optimisation for task- and constraint-aware design of particle detectors in the context of muon tomography},\ }\href {https://doi.org/10.1088/2632-2153/ad52e7} {\bibfield  {journal} {\bibinfo  {journal} {Machine Learning: Science and Technology}\ }\textbf {\bibinfo {volume} {5}},\ \bibinfo {pages} {035002} (\bibinfo {year} {2024})}\BibitemShut {NoStop}%
\bibitem [{\citenamefont {Qasim}\ \emph {et~al.}(2024)\citenamefont {Qasim}, \citenamefont {Owen},\ and\ \citenamefont {Serra}}]{qasim2024physicsinstrumentdesignreinforcement}%
  \BibitemOpen
  \bibfield  {author} {\bibinfo {author} {\bibfnamefont {S.~R.}\ \bibnamefont {Qasim}}, \bibinfo {author} {\bibfnamefont {P.}~\bibnamefont {Owen}},\ and\ \bibinfo {author} {\bibfnamefont {N.}~\bibnamefont {Serra}},\ }\href {https://arxiv.org/abs/2412.10237} {\bibinfo {title} {Physics instrument design with reinforcement learning}} (\bibinfo {year} {2024}),\ \Eprint {https://arxiv.org/abs/2412.10237} {arXiv:2412.10237 [physics.ins-det]} \BibitemShut {NoStop}%
\bibitem [{\citenamefont {Kortus}\ \emph {et~al.}(2025)\citenamefont {Kortus}, \citenamefont {Keidel}, \citenamefont {Gauger},\ and\ \citenamefont {Kieseler}}]{kortus2025constrainedoptimizationchargedparticle}%
  \BibitemOpen
  \bibfield  {author} {\bibinfo {author} {\bibfnamefont {T.}~\bibnamefont {Kortus}}, \bibinfo {author} {\bibfnamefont {R.}~\bibnamefont {Keidel}}, \bibinfo {author} {\bibfnamefont {N.~R.}\ \bibnamefont {Gauger}},\ and\ \bibinfo {author} {\bibfnamefont {J.}~\bibnamefont {Kieseler}},\ }\href {https://arxiv.org/abs/2501.05113} {\bibinfo {title} {Constrained optimization of charged particle tracking with multi-agent reinforcement learning}} (\bibinfo {year} {2025}),\ \Eprint {https://arxiv.org/abs/2501.05113} {arXiv:2501.05113 [physics.comp-ph]} \BibitemShut {NoStop}%
\bibitem [{\citenamefont {Golovin}\ \emph {et~al.}(2017)\citenamefont {Golovin}, \citenamefont {Kochanski},\ and\ \citenamefont {Karro}}]{Golovin2017BlackBoxGenetic}%
  \BibitemOpen
  \bibfield  {author} {\bibinfo {author} {\bibfnamefont {D.}~\bibnamefont {Golovin}}, \bibinfo {author} {\bibfnamefont {G.}~\bibnamefont {Kochanski}},\ and\ \bibinfo {author} {\bibfnamefont {J.~E.}\ \bibnamefont {Karro}},\ }\bibfield  {title} {\bibinfo {title} {Black box optimization via a bayesian-optimized genetic algorithm},\ }in\ \href@noop {} {\emph {\bibinfo {booktitle} {Advances in Neural Information Processing Systems 30 (NIPS 2017)}}}\ (\bibinfo {year} {2017})\ \bibinfo {note} {to be submitted to Opt2017 Optimization for Machine Learning, at NIPS 2017.}\BibitemShut {Stop}%
\bibitem [{\citenamefont {Hofler}\ \emph {et~al.}(2013)\citenamefont {Hofler}, \citenamefont {Terzi\ifmmode~\acute{c}\else \'{c}\fi{}}, \citenamefont {Kramer}, \citenamefont {Zvezdin}, \citenamefont {Morozov}, \citenamefont {Roblin}, \citenamefont {Lin},\ and\ \citenamefont {Jarvis}}]{Hofler2013innovativeGenetic}%
  \BibitemOpen
  \bibfield  {author} {\bibinfo {author} {\bibfnamefont {A.}~\bibnamefont {Hofler}}, \bibinfo {author} {\bibfnamefont {B.~c.~v.}\ \bibnamefont {Terzi\ifmmode~\acute{c}\else \'{c}\fi{}}}, \bibinfo {author} {\bibfnamefont {M.}~\bibnamefont {Kramer}}, \bibinfo {author} {\bibfnamefont {A.}~\bibnamefont {Zvezdin}}, \bibinfo {author} {\bibfnamefont {V.}~\bibnamefont {Morozov}}, \bibinfo {author} {\bibfnamefont {Y.}~\bibnamefont {Roblin}}, \bibinfo {author} {\bibfnamefont {F.}~\bibnamefont {Lin}},\ and\ \bibinfo {author} {\bibfnamefont {C.}~\bibnamefont {Jarvis}},\ }\bibfield  {title} {\bibinfo {title} {Innovative applications of genetic algorithms to problems in accelerator physics},\ }\href {https://doi.org/10.1103/PhysRevSTAB.16.010101} {\bibfield  {journal} {\bibinfo  {journal} {Phys. Rev. ST Accel. Beams}\ }\textbf {\bibinfo {volume} {16}},\ \bibinfo {pages} {010101} (\bibinfo {year} {2013})}\BibitemShut {NoStop}%
\bibitem [{\citenamefont {Geisel}\ \emph {et~al.}(2013)\citenamefont {Geisel} \emph {et~al.}}]{Geisel2013evolutionaryOpt}%
  \BibitemOpen
  \bibfield  {author} {\bibinfo {author} {\bibfnamefont {I.}~\bibnamefont {Geisel}} \emph {et~al.},\ }\bibfield  {title} {\bibinfo {title} {Evolutionary optimization of an experimental apparatus},\ }\href {https://doi.org/10.1063/1.4808213} {\bibfield  {journal} {\bibinfo  {journal} {Applied Physics Letters}\ }\textbf {\bibinfo {volume} {102}},\ \bibinfo {pages} {214105} (\bibinfo {year} {2013})},\ \Eprint {https://arxiv.org/abs/https://pubs.aip.org/aip/apl/article-pdf/doi/10.1063/1.4808213/14276514/214105\_1\_online.pdf} {https://pubs.aip.org/aip/apl/article-pdf/doi/10.1063/1.4808213/14276514/214105\_1\_online.pdf} \BibitemShut {NoStop}%
\bibitem [{\citenamefont {Rolla}\ \emph {et~al.}(2023)\citenamefont {Rolla}, \citenamefont {Machtay}, \citenamefont {Patton}, \citenamefont {Banzhaf}, \citenamefont {Connolly}, \citenamefont {Debolt}, \citenamefont {Deer}, \citenamefont {Fahimi}, \citenamefont {Ferstle}, \citenamefont {Kuzma}, \citenamefont {Pfendner}, \citenamefont {Sipe}, \citenamefont {Staats},\ and\ \citenamefont {Wissel}}]{Rolla2023evolutionAntenna}%
  \BibitemOpen
  \bibfield  {author} {\bibinfo {author} {\bibfnamefont {J.}~\bibnamefont {Rolla}}, \bibinfo {author} {\bibfnamefont {A.}~\bibnamefont {Machtay}}, \bibinfo {author} {\bibfnamefont {A.}~\bibnamefont {Patton}}, \bibinfo {author} {\bibfnamefont {W.}~\bibnamefont {Banzhaf}}, \bibinfo {author} {\bibfnamefont {A.}~\bibnamefont {Connolly}}, \bibinfo {author} {\bibfnamefont {R.}~\bibnamefont {Debolt}}, \bibinfo {author} {\bibfnamefont {L.}~\bibnamefont {Deer}}, \bibinfo {author} {\bibfnamefont {E.}~\bibnamefont {Fahimi}}, \bibinfo {author} {\bibfnamefont {E.}~\bibnamefont {Ferstle}}, \bibinfo {author} {\bibfnamefont {P.}~\bibnamefont {Kuzma}}, \bibinfo {author} {\bibfnamefont {C.}~\bibnamefont {Pfendner}}, \bibinfo {author} {\bibfnamefont {B.}~\bibnamefont {Sipe}}, \bibinfo {author} {\bibfnamefont {K.}~\bibnamefont {Staats}},\ and\ \bibinfo {author} {\bibfnamefont {S.~A.}\ \bibnamefont {Wissel}} (\bibinfo {collaboration} {GENETIS Collaboration}),\ }\bibfield  {title} {\bibinfo {title} {Using evolutionary algorithms
  to design antennas with greater sensitivity to ultrahigh energy neutrinos},\ }\href {https://doi.org/10.1103/PhysRevD.108.102002} {\bibfield  {journal} {\bibinfo  {journal} {Phys. Rev. D}\ }\textbf {\bibinfo {volume} {108}},\ \bibinfo {pages} {102002} (\bibinfo {year} {2023})}\BibitemShut {NoStop}%
\bibitem [{\citenamefont {Shahriari}(2016)}]{Shahriari2016revBayesOpt}%
  \BibitemOpen
  \bibfield  {author} {\bibinfo {author} {\bibfnamefont {B.~e.~a.}\ \bibnamefont {Shahriari}},\ }\bibfield  {title} {\bibinfo {title} {Taking the human out of the loop: A review of bayesian optimization},\ }\href {https://doi.org/10.1109/JPROC.2015.2494218} {\bibfield  {journal} {\bibinfo  {journal} {Proceedings of the IEEE}\ }\textbf {\bibinfo {volume} {104}},\ \bibinfo {pages} {148} (\bibinfo {year} {2016})}\BibitemShut {NoStop}%
\bibitem [{\citenamefont {Eriksson}\ \emph {et~al.}(2019)\citenamefont {Eriksson}, \citenamefont {Pearce}, \citenamefont {Gardner}, \citenamefont {Turner},\ and\ \citenamefont {Poloczek}}]{Eriksson2019scalableGlobal}%
  \BibitemOpen
  \bibfield  {author} {\bibinfo {author} {\bibfnamefont {D.}~\bibnamefont {Eriksson}}, \bibinfo {author} {\bibfnamefont {M.}~\bibnamefont {Pearce}}, \bibinfo {author} {\bibfnamefont {J.}~\bibnamefont {Gardner}}, \bibinfo {author} {\bibfnamefont {R.~D.}\ \bibnamefont {Turner}},\ and\ \bibinfo {author} {\bibfnamefont {M.}~\bibnamefont {Poloczek}},\ }\bibfield  {title} {\bibinfo {title} {Scalable global optimization via local bayesian optimization},\ }in\ \href {https://proceedings.neurips.cc/paper_files/paper/2019/file/6c990b7aca7bc7058f5e98ea909e924b-Paper.pdf} {\emph {\bibinfo {booktitle} {Advances in Neural Information Processing Systems}}},\ Vol.~\bibinfo {volume} {32},\ \bibinfo {editor} {edited by\ \bibinfo {editor} {\bibfnamefont {H.}~\bibnamefont {Wallach}}, \bibinfo {editor} {\bibfnamefont {H.}~\bibnamefont {Larochelle}}, \bibinfo {editor} {\bibfnamefont {A.}~\bibnamefont {Beygelzimer}}, \bibinfo {editor} {\bibfnamefont {F.}~\bibnamefont {d\textquotesingle Alch\'{e}-Buc}}, \bibinfo {editor}
  {\bibfnamefont {E.}~\bibnamefont {Fox}},\ and\ \bibinfo {editor} {\bibfnamefont {R.}~\bibnamefont {Garnett}}}\ (\bibinfo  {publisher} {Curran Associates, Inc.},\ \bibinfo {year} {2019})\BibitemShut {NoStop}%
\bibitem [{\citenamefont {Wu}\ \emph {et~al.}(2023)\citenamefont {Wu}, \citenamefont {Kim}, \citenamefont {Garnett},\ and\ \citenamefont {Gardner}}]{Wu2023behaviorConvergence}%
  \BibitemOpen
  \bibfield  {author} {\bibinfo {author} {\bibfnamefont {K.}~\bibnamefont {Wu}}, \bibinfo {author} {\bibfnamefont {K.}~\bibnamefont {Kim}}, \bibinfo {author} {\bibfnamefont {R.}~\bibnamefont {Garnett}},\ and\ \bibinfo {author} {\bibfnamefont {J.~R.}\ \bibnamefont {Gardner}},\ }\bibfield  {title} {\bibinfo {title} {The behavior and convergence of local bayesian optimization},\ }in\ \href@noop {} {\emph {\bibinfo {booktitle} {Proceedings of the 37th International Conference on Neural Information Processing Systems}}},\ \bibinfo {series and number} {NIPS '23}\ (\bibinfo  {publisher} {Curran Associates Inc.},\ \bibinfo {address} {Red Hook, NY, USA},\ \bibinfo {year} {2023})\BibitemShut {NoStop}%
\bibitem [{\citenamefont {Nguyen}\ \emph {et~al.}(2022)\citenamefont {Nguyen}, \citenamefont {Wu}, \citenamefont {Gardner},\ and\ \citenamefont {Garnett}}]{Nguyen2022LocalBO}%
  \BibitemOpen
  \bibfield  {author} {\bibinfo {author} {\bibfnamefont {Q.}~\bibnamefont {Nguyen}}, \bibinfo {author} {\bibfnamefont {K.}~\bibnamefont {Wu}}, \bibinfo {author} {\bibfnamefont {J.~R.}\ \bibnamefont {Gardner}},\ and\ \bibinfo {author} {\bibfnamefont {R.}~\bibnamefont {Garnett}},\ }\bibfield  {title} {\bibinfo {title} {Local bayesian optimization via maximizing probability of descent},\ }\href {https://api.semanticscholar.org/CorpusID:253080324} {\bibfield  {journal} {\bibinfo  {journal} {Advances in neural information processing systems}\ }\textbf {\bibinfo {volume} {abs/2210.11662}} (\bibinfo {year} {2022})}\BibitemShut {NoStop}%
\bibitem [{\citenamefont {Zhang}\ \emph {et~al.}(2020)\citenamefont {Zhang}, \citenamefont {Apley},\ and\ \citenamefont {Chen}}]{Zhang2020bayesianOpt}%
  \BibitemOpen
  \bibfield  {author} {\bibinfo {author} {\bibfnamefont {Y.}~\bibnamefont {Zhang}}, \bibinfo {author} {\bibfnamefont {D.}~\bibnamefont {Apley}},\ and\ \bibinfo {author} {\bibfnamefont {W.}~\bibnamefont {Chen}},\ }\bibfield  {title} {\bibinfo {title} {Bayesian optimization for materials design with mixed quantitative and qualitative variables},\ }\href {https://doi.org/10.1038/s41598-020-60652-9} {\bibfield  {journal} {\bibinfo  {journal} {Scientific Reports}\ }\textbf {\bibinfo {volume} {10}} (\bibinfo {year} {2020})}\BibitemShut {NoStop}%
\bibitem [{\citenamefont {Khatamsaz}\ \emph {et~al.}(2023)\citenamefont {Khatamsaz} \emph {et~al.}}]{Khatamsaz2023physicsInformed}%
  \BibitemOpen
  \bibfield  {author} {\bibinfo {author} {\bibfnamefont {D.}~\bibnamefont {Khatamsaz}} \emph {et~al.},\ }\bibfield  {title} {\bibinfo {title} {A physics informed bayesian optimization approach for material design: application to niti shape memory alloys},\ }\href {https://doi.org/10.1038/s41524-023-01173-7} {\bibfield  {journal} {\bibinfo  {journal} {npj Computational Materials}\ }\textbf {\bibinfo {volume} {9}} (\bibinfo {year} {2023})}\BibitemShut {NoStop}%
\bibitem [{\citenamefont {Nicoli}\ \emph {et~al.}(2023)\citenamefont {Nicoli} \emph {et~al.}}]{Nicoli2023physicsInformed}%
  \BibitemOpen
  \bibfield  {author} {\bibinfo {author} {\bibfnamefont {K.~A.}\ \bibnamefont {Nicoli}} \emph {et~al.},\ }\bibfield  {title} {\bibinfo {title} {Physics-informed bayesian optimization of variational quantum circuits},\ }in\ \href@noop {} {\emph {\bibinfo {booktitle} {Proceedings of the 37th International Conference on Neural Information Processing Systems}}},\ \bibinfo {series and number} {NIPS '23}\ (\bibinfo  {publisher} {Curran Associates Inc.},\ \bibinfo {address} {Red Hook, NY, USA},\ \bibinfo {year} {2023})\BibitemShut {NoStop}%
\bibitem [{\citenamefont {Grbcic}\ \emph {et~al.}(2025)\citenamefont {Grbcic} \emph {et~al.}}]{Grbcic2025AILaser}%
  \BibitemOpen
  \bibfield  {author} {\bibinfo {author} {\bibfnamefont {L.}~\bibnamefont {Grbcic}} \emph {et~al.},\ }\bibfield  {title} {\bibinfo {title} {Artificial intelligence driven laser parameter search: Inverse design of photonic surfaces using greedy surrogate-based optimization},\ }\href {https://doi.org/https://doi.org/10.1016/j.engappai.2024.109971} {\bibfield  {journal} {\bibinfo  {journal} {Engineering Applications of Artificial Intelligence}\ }\textbf {\bibinfo {volume} {143}},\ \bibinfo {pages} {109971} (\bibinfo {year} {2025})}\BibitemShut {NoStop}%
\bibitem [{\citenamefont {Elzouka}\ \emph {et~al.}(2020)\citenamefont {Elzouka} \emph {et~al.}}]{elzouka2020interpretable}%
  \BibitemOpen
  \bibfield  {author} {\bibinfo {author} {\bibfnamefont {M.}~\bibnamefont {Elzouka}} \emph {et~al.},\ }\bibfield  {title} {\bibinfo {title} {Interpretable forward and inverse design of particle spectral emissivity using common machine-learning models},\ }\href@noop {} {\bibfield  {journal} {\bibinfo  {journal} {Cell Reports Physical Science}\ }\textbf {\bibinfo {volume} {1}} (\bibinfo {year} {2020})}\BibitemShut {NoStop}%
\bibitem [{\citenamefont {Fanelli}(2022)}]{Fanelli2022designDetectAi}%
  \BibitemOpen
  \bibfield  {author} {\bibinfo {author} {\bibfnamefont {C.}~\bibnamefont {Fanelli}},\ }\bibfield  {title} {\bibinfo {title} {Design of detectors at the electron ion collider with artificial intelligence},\ }\href {https://doi.org/10.1088/1748-0221/17/04/C04038} {\bibfield  {journal} {\bibinfo  {journal} {Journal of Instrumentation}\ }\textbf {\bibinfo {volume} {17}}\bibinfo  {number} { (04)},\ \bibinfo {pages} {C04038}}\BibitemShut {NoStop}%
\bibitem [{\citenamefont {Hashemi}\ and\ \citenamefont {Krause}(2024)}]{dgm_surrogates_HASHEMI2024100092}%
  \BibitemOpen
\bibfield  {number} {  }\bibfield  {author} {\bibinfo {author} {\bibfnamefont {B.}~\bibnamefont {Hashemi}}\ and\ \bibinfo {author} {\bibfnamefont {C.}~\bibnamefont {Krause}},\ }\bibfield  {title} {\bibinfo {title} {Deep generative models for detector signature simulation: A taxonomic review},\ }\href {https://doi.org/https://doi.org/10.1016/j.revip.2024.100092} {\bibfield  {journal} {\bibinfo  {journal} {Reviews in Physics}\ }\textbf {\bibinfo {volume} {12}},\ \bibinfo {pages} {100092} (\bibinfo {year} {2024})}\BibitemShut {NoStop}%
\bibitem [{\citenamefont {Shirobokov}\ \emph {et~al.}(2020)\citenamefont {Shirobokov}, \citenamefont {Belavin}, \citenamefont {Kagan}, \citenamefont {Ustyuzhanin},\ and\ \citenamefont {Baydin}}]{black_box_NEURIPS2020_a878dbeb}%
  \BibitemOpen
  \bibfield  {author} {\bibinfo {author} {\bibfnamefont {S.}~\bibnamefont {Shirobokov}}, \bibinfo {author} {\bibfnamefont {V.}~\bibnamefont {Belavin}}, \bibinfo {author} {\bibfnamefont {M.}~\bibnamefont {Kagan}}, \bibinfo {author} {\bibfnamefont {A.}~\bibnamefont {Ustyuzhanin}},\ and\ \bibinfo {author} {\bibfnamefont {A.~G.}\ \bibnamefont {Baydin}},\ }\bibfield  {title} {\bibinfo {title} {Black-box optimization with local generative surrogates},\ }in\ \href {https://proceedings.neurips.cc/paper_files/paper/2020/file/a878dbebc902328b41dbf02aa87abb58-Paper.pdf} {\emph {\bibinfo {booktitle} {Advances in Neural Information Processing Systems}}},\ Vol.~\bibinfo {volume} {33},\ \bibinfo {editor} {edited by\ \bibinfo {editor} {\bibfnamefont {H.}~\bibnamefont {Larochelle}}, \bibinfo {editor} {\bibfnamefont {M.}~\bibnamefont {Ranzato}}, \bibinfo {editor} {\bibfnamefont {R.}~\bibnamefont {Hadsell}}, \bibinfo {editor} {\bibfnamefont {M.}~\bibnamefont {Balcan}},\ and\ \bibinfo {editor} {\bibfnamefont {H.}~\bibnamefont
  {Lin}}}\ (\bibinfo  {publisher} {Curran Associates, Inc.},\ \bibinfo {year} {2020})\ pp.\ \bibinfo {pages} {14650--14662}\BibitemShut {NoStop}%
\bibitem [{\citenamefont {Schmidt}\ \emph {et~al.}(2025)\citenamefont {Schmidt}, \citenamefont {Kota}, \citenamefont {Kieseler}, \citenamefont {Vita}, \citenamefont {Klute}, \citenamefont {Abhishek}, \citenamefont {Aehle}, \citenamefont {Awais}, \citenamefont {Breccia}, \citenamefont {Carroccio}, \citenamefont {Chen}, \citenamefont {Dorigo}, \citenamefont {Gauger}, \citenamefont {Lupi}, \citenamefont {Nardi}, \citenamefont {Nguyen}, \citenamefont {Sandin}, \citenamefont {Willmore},\ and\ \citenamefont {Vischia}}]{schmidt2025endtoenddetectoroptimizationdiffusion}%
  \BibitemOpen
  \bibfield  {author} {\bibinfo {author} {\bibfnamefont {K.}~\bibnamefont {Schmidt}}, \bibinfo {author} {\bibfnamefont {N.}~\bibnamefont {Kota}}, \bibinfo {author} {\bibfnamefont {J.}~\bibnamefont {Kieseler}}, \bibinfo {author} {\bibfnamefont {A.~D.}\ \bibnamefont {Vita}}, \bibinfo {author} {\bibfnamefont {M.}~\bibnamefont {Klute}}, \bibinfo {author} {\bibnamefont {Abhishek}}, \bibinfo {author} {\bibfnamefont {M.}~\bibnamefont {Aehle}}, \bibinfo {author} {\bibfnamefont {M.}~\bibnamefont {Awais}}, \bibinfo {author} {\bibfnamefont {A.}~\bibnamefont {Breccia}}, \bibinfo {author} {\bibfnamefont {R.}~\bibnamefont {Carroccio}}, \bibinfo {author} {\bibfnamefont {L.}~\bibnamefont {Chen}}, \bibinfo {author} {\bibfnamefont {T.}~\bibnamefont {Dorigo}}, \bibinfo {author} {\bibfnamefont {N.~R.}\ \bibnamefont {Gauger}}, \bibinfo {author} {\bibfnamefont {E.}~\bibnamefont {Lupi}}, \bibinfo {author} {\bibfnamefont {F.}~\bibnamefont {Nardi}}, \bibinfo {author} {\bibfnamefont {X.~T.}\ \bibnamefont {Nguyen}}, \bibinfo {author}
  {\bibfnamefont {F.}~\bibnamefont {Sandin}}, \bibinfo {author} {\bibfnamefont {J.}~\bibnamefont {Willmore}},\ and\ \bibinfo {author} {\bibfnamefont {P.}~\bibnamefont {Vischia}},\ }\href {https://arxiv.org/abs/2502.02152} {\bibinfo {title} {End-to-end detector optimization with diffusion models: A case study in sampling calorimeters}} (\bibinfo {year} {2025}),\ \Eprint {https://arxiv.org/abs/2502.02152} {arXiv:2502.02152 [physics.ins-det]} \BibitemShut {NoStop}%
\bibitem [{\citenamefont {Neiswanger}\ \emph {et~al.}(2021)\citenamefont {Neiswanger}, \citenamefont {Wang},\ and\ \citenamefont {Ermon}}]{Neiswanger2021BayesianAE}%
  \BibitemOpen
  \bibfield  {author} {\bibinfo {author} {\bibfnamefont {W.}~\bibnamefont {Neiswanger}}, \bibinfo {author} {\bibfnamefont {K.~A.}\ \bibnamefont {Wang}},\ and\ \bibinfo {author} {\bibfnamefont {S.}~\bibnamefont {Ermon}},\ }\bibfield  {title} {\bibinfo {title} {Bayesian algorithm execution: Estimating computable properties of black-box functions using mutual information},\ }in\ \href@noop {} {\emph {\bibinfo {booktitle} {International Conference on Machine Learning (ICML)}}}\ (\bibinfo {year} {2021})\BibitemShut {NoStop}%
\bibitem [{\citenamefont {Chitturi}\ \emph {et~al.}(2024)\citenamefont {Chitturi}, \citenamefont {Ramdas}, \citenamefont {Wu}, \citenamefont {Rohr}, \citenamefont {Ermon}, \citenamefont {Dionne}, \citenamefont {Jornada}, \citenamefont {Dunne}, \citenamefont {Tassone}, \citenamefont {Neiswanger},\ and\ \citenamefont {Ratner}}]{Chitturi2024targetedMaterials}%
  \BibitemOpen
  \bibfield  {author} {\bibinfo {author} {\bibfnamefont {S.~R.}\ \bibnamefont {Chitturi}}, \bibinfo {author} {\bibfnamefont {A.}~\bibnamefont {Ramdas}}, \bibinfo {author} {\bibfnamefont {Y.}~\bibnamefont {Wu}}, \bibinfo {author} {\bibfnamefont {B.}~\bibnamefont {Rohr}}, \bibinfo {author} {\bibfnamefont {S.}~\bibnamefont {Ermon}}, \bibinfo {author} {\bibfnamefont {J.}~\bibnamefont {Dionne}}, \bibinfo {author} {\bibfnamefont {F.~H.~d.}\ \bibnamefont {Jornada}}, \bibinfo {author} {\bibfnamefont {M.}~\bibnamefont {Dunne}}, \bibinfo {author} {\bibfnamefont {C.}~\bibnamefont {Tassone}}, \bibinfo {author} {\bibfnamefont {W.}~\bibnamefont {Neiswanger}},\ and\ \bibinfo {author} {\bibfnamefont {D.}~\bibnamefont {Ratner}},\ }\bibfield  {title} {\bibinfo {title} {Targeted materials discovery using bayesian algorithm execution},\ }\bibfield  {journal} {\bibinfo  {journal} {npj Computational Materials}\ }\textbf {\bibinfo {volume} {10}},\ \href {https://doi.org/10.1038/s41524-024-01326-2} {10.1038/s41524-024-01326-2}
  (\bibinfo {year} {2024})\BibitemShut {NoStop}%
\bibitem [{\citenamefont {MacKay}(2002)}]{macKay2002info}%
  \BibitemOpen
  \bibfield  {author} {\bibinfo {author} {\bibfnamefont {D.~J.~C.}\ \bibnamefont {MacKay}},\ }\href@noop {} {\emph {\bibinfo {title} {Information Theory, Inference \& Learning Algorithms}}}\ (\bibinfo  {publisher} {Cambridge University Press},\ \bibinfo {address} {USA},\ \bibinfo {year} {2002})\BibitemShut {NoStop}%
\bibitem [{\citenamefont {Cover}(1999)}]{cover1999elements}%
  \BibitemOpen
  \bibfield  {author} {\bibinfo {author} {\bibfnamefont {T.~M.}\ \bibnamefont {Cover}},\ }\href@noop {} {\emph {\bibinfo {title} {Elements of information theory}}}\ (\bibinfo  {publisher} {John Wiley \& Sons},\ \bibinfo {year} {1999})\BibitemShut {NoStop}%
\bibitem [{\citenamefont {Bitbol}(2018)}]{Bitbol2018InferringIP}%
  \BibitemOpen
  \bibfield  {author} {\bibinfo {author} {\bibfnamefont {A.-F.}\ \bibnamefont {Bitbol}},\ }\bibfield  {title} {\bibinfo {title} {Inferring interaction partners from protein sequences using mutual information},\ }\href {https://api.semanticscholar.org/CorpusID:51714422} {\bibfield  {journal} {\bibinfo  {journal} {PLoS Computational Biology}\ }\textbf {\bibinfo {volume} {14}} (\bibinfo {year} {2018})}\BibitemShut {NoStop}%
\bibitem [{\citenamefont {Paninski}(2003)}]{paninski2003estimation}%
  \BibitemOpen
  \bibfield  {author} {\bibinfo {author} {\bibfnamefont {L.}~\bibnamefont {Paninski}},\ }\bibfield  {title} {\bibinfo {title} {Estimation of entropy and mutual information},\ }\href {https://doi.org/10.1162/089976603321780272} {\bibfield  {journal} {\bibinfo  {journal} {Neural Computation}\ }\textbf {\bibinfo {volume} {15}},\ \bibinfo {pages} {1191} (\bibinfo {year} {2003})},\ \Eprint {https://arxiv.org/abs/https://direct.mit.edu/neco/article-pdf/15/6/1191/815550/089976603321780272.pdf} {https://direct.mit.edu/neco/article-pdf/15/6/1191/815550/089976603321780272.pdf} \BibitemShut {NoStop}%
\bibitem [{\citenamefont {Bahl}\ \emph {et~al.}(1986)\citenamefont {Bahl}, \citenamefont {Brown}, \citenamefont {Souza},\ and\ \citenamefont {Mercer}}]{bahl1986maxMIspeech}%
  \BibitemOpen
  \bibfield  {author} {\bibinfo {author} {\bibfnamefont {L.}~\bibnamefont {Bahl}}, \bibinfo {author} {\bibfnamefont {P.}~\bibnamefont {Brown}}, \bibinfo {author} {\bibfnamefont {P.}~\bibnamefont {Souza}},\ and\ \bibinfo {author} {\bibfnamefont {R.}~\bibnamefont {Mercer}},\ }\bibfield  {title} {\bibinfo {title} {Maximum mutual information estimation of hidden markov parameters for speech recognition},\ }in\ \href {https://doi.org/10.1109/ICASSP.1986.1169179} {\emph {\bibinfo {booktitle} {ICASSP, IEEE International Conference on Acoustics, Speech and Signal Processing - Proceedings}}},\ Vol.~\bibinfo {volume} {11}\ (\bibinfo {year} {1986})\ pp.\ \bibinfo {pages} {49 -- 52}\BibitemShut {NoStop}%
\bibitem [{\citenamefont {Church}\ and\ \citenamefont {Hanks}(1990)}]{church-hanks-1990-word}%
  \BibitemOpen
  \bibfield  {author} {\bibinfo {author} {\bibfnamefont {K.~W.}\ \bibnamefont {Church}}\ and\ \bibinfo {author} {\bibfnamefont {P.}~\bibnamefont {Hanks}},\ }\bibfield  {title} {\bibinfo {title} {Word association norms, mutual information, and lexicography},\ }\href {https://aclanthology.org/J90-1003/} {\bibfield  {journal} {\bibinfo  {journal} {Computational Linguistics}\ }\textbf {\bibinfo {volume} {16}},\ \bibinfo {pages} {22} (\bibinfo {year} {1990})}\BibitemShut {NoStop}%
\bibitem [{\citenamefont {Agostinelli}\ \emph {et~al.}(2003)\citenamefont {Agostinelli} \emph {et~al.}}]{Geant_4_0_AGOSTINELLI2003250}%
  \BibitemOpen
  \bibfield  {author} {\bibinfo {author} {\bibfnamefont {S.}~\bibnamefont {Agostinelli}} \emph {et~al.},\ }\bibfield  {title} {\bibinfo {title} {Geant4—a simulation toolkit},\ }\href {https://doi.org/https://doi.org/10.1016/S0168-9002(03)01368-8} {\bibfield  {journal} {\bibinfo  {journal} {Nuclear Instruments and Methods in Physics Research Section A: Accelerators, Spectrometers, Detectors and Associated Equipment}\ }\textbf {\bibinfo {volume} {506}},\ \bibinfo {pages} {250} (\bibinfo {year} {2003})}\BibitemShut {NoStop}%
\bibitem [{\citenamefont {Allison}\ \emph {et~al.}(2006)\citenamefont {Allison} \emph {et~al.}}]{Geant4_1_1610988}%
  \BibitemOpen
  \bibfield  {author} {\bibinfo {author} {\bibfnamefont {J.}~\bibnamefont {Allison}} \emph {et~al.},\ }\bibfield  {title} {\bibinfo {title} {Geant4 developments and applications},\ }\href {https://doi.org/10.1109/TNS.2006.869826} {\bibfield  {journal} {\bibinfo  {journal} {IEEE Transactions on Nuclear Science}\ }\textbf {\bibinfo {volume} {53}},\ \bibinfo {pages} {270} (\bibinfo {year} {2006})}\BibitemShut {NoStop}%
\bibitem [{\citenamefont {Allison}\ \emph {et~al.}(2016)\citenamefont {Allison} \emph {et~al.}}]{Geant4_2_ALLISON2016186}%
  \BibitemOpen
  \bibfield  {author} {\bibinfo {author} {\bibfnamefont {J.}~\bibnamefont {Allison}} \emph {et~al.},\ }\bibfield  {title} {\bibinfo {title} {Recent developments in geant4},\ }\href {https://doi.org/https://doi.org/10.1016/j.nima.2016.06.125} {\bibfield  {journal} {\bibinfo  {journal} {Nuclear Instruments and Methods in Physics Research Section A: Accelerators, Spectrometers, Detectors and Associated Equipment}\ }\textbf {\bibinfo {volume} {835}},\ \bibinfo {pages} {186} (\bibinfo {year} {2016})}\BibitemShut {NoStop}%
\bibitem [{\citenamefont {Aleksandrov}\ \emph {et~al.}(2005)\citenamefont {Aleksandrov} \emph {et~al.}}]{pwo_alice_ALEKSANDROV2005169}%
  \BibitemOpen
  \bibfield  {author} {\bibinfo {author} {\bibfnamefont {D.}~\bibnamefont {Aleksandrov}} \emph {et~al.},\ }\bibfield  {title} {\bibinfo {title} {A high resolution electromagnetic calorimeter based on lead-tungstate crystals},\ }\href {https://doi.org/https://doi.org/10.1016/j.nima.2005.03.174} {\bibfield  {journal} {\bibinfo  {journal} {Nuclear Instruments and Methods in Physics Research Section A: Accelerators, Spectrometers, Detectors and Associated Equipment}\ }\textbf {\bibinfo {volume} {550}},\ \bibinfo {pages} {169} (\bibinfo {year} {2005})}\BibitemShut {NoStop}%
\bibitem [{\citenamefont {Lecoq}\ \emph {et~al.}(1995)\citenamefont {Lecoq} \emph {et~al.}}]{pwo_lhc_LECOQ1995291}%
  \BibitemOpen
  \bibfield  {author} {\bibinfo {author} {\bibfnamefont {P.}~\bibnamefont {Lecoq}} \emph {et~al.},\ }\bibfield  {title} {\bibinfo {title} {Lead tungstate (pbwo4) scintillators for lhc em calorimetry},\ }\href {https://doi.org/https://doi.org/10.1016/0168-9002(95)00589-7} {\bibfield  {journal} {\bibinfo  {journal} {Nuclear Instruments and Methods in Physics Research Section A: Accelerators, Spectrometers, Detectors and Associated Equipment}\ }\textbf {\bibinfo {volume} {365}},\ \bibinfo {pages} {291} (\bibinfo {year} {1995})}\BibitemShut {NoStop}%
\bibitem [{\citenamefont {Bandiera}\ \emph {et~al.}(2023)\citenamefont {Bandiera} \emph {et~al.}}]{pwo_highly_compact_oriented_10.3389/fphy.2023.1254020}%
  \BibitemOpen
  \bibfield  {author} {\bibinfo {author} {\bibfnamefont {L.}~\bibnamefont {Bandiera}} \emph {et~al.},\ }\bibfield  {title} {\bibinfo {title} {A highly-compact and ultra-fast homogeneous electromagnetic calorimeter based on oriented lead tungstate crystals},\ }\bibfield  {journal} {\bibinfo  {journal} {Frontiers in Physics}\ }\textbf {\bibinfo {volume} {11}},\ \href {https://doi.org/10.3389/fphy.2023.1254020} {10.3389/fphy.2023.1254020} (\bibinfo {year} {2023})\BibitemShut {NoStop}%
\bibitem [{\citenamefont {Sorensen}(1982)}]{trust_region_doi:10.1137/0719026}%
  \BibitemOpen
  \bibfield  {author} {\bibinfo {author} {\bibfnamefont {D.~C.}\ \bibnamefont {Sorensen}},\ }\bibfield  {title} {\bibinfo {title} {Newton’s method with a model trust region modification},\ }\href {https://doi.org/10.1137/0719026} {\bibfield  {journal} {\bibinfo  {journal} {SIAM Journal on Numerical Analysis}\ }\textbf {\bibinfo {volume} {19}},\ \bibinfo {pages} {409} (\bibinfo {year} {1982})},\ \Eprint {https://arxiv.org/abs/https://doi.org/10.1137/0719026} {https://doi.org/10.1137/0719026} \BibitemShut {NoStop}%
\bibitem [{\citenamefont {Kraskov}\ \emph {et~al.}(2004)\citenamefont {Kraskov}, \citenamefont {St\"ogbauer},\ and\ \citenamefont {Grassberger}}]{kraskov2004estimatingMI}%
  \BibitemOpen
  \bibfield  {author} {\bibinfo {author} {\bibfnamefont {A.}~\bibnamefont {Kraskov}}, \bibinfo {author} {\bibfnamefont {H.}~\bibnamefont {St\"ogbauer}},\ and\ \bibinfo {author} {\bibfnamefont {P.}~\bibnamefont {Grassberger}},\ }\bibfield  {title} {\bibinfo {title} {Estimating mutual information},\ }\href {https://doi.org/10.1103/PhysRevE.69.066138} {\bibfield  {journal} {\bibinfo  {journal} {Phys. Rev. E}\ }\textbf {\bibinfo {volume} {69}},\ \bibinfo {pages} {066138} (\bibinfo {year} {2004})}\BibitemShut {NoStop}%
\bibitem [{\citenamefont {Rezende}\ and\ \citenamefont {Mohamed}(2016)}]{normalising_flows_0_rezende2016variationalinferencenormalizingflows}%
  \BibitemOpen
  \bibfield  {author} {\bibinfo {author} {\bibfnamefont {D.~J.}\ \bibnamefont {Rezende}}\ and\ \bibinfo {author} {\bibfnamefont {S.}~\bibnamefont {Mohamed}},\ }\href {https://arxiv.org/abs/1505.05770} {\bibinfo {title} {Variational inference with normalizing flows}} (\bibinfo {year} {2016}),\ \Eprint {https://arxiv.org/abs/1505.05770} {arXiv:1505.05770 [stat.ML]} \BibitemShut {NoStop}%
\bibitem [{\citenamefont {Kobyzev}\ \emph {et~al.}(2021)\citenamefont {Kobyzev}, \citenamefont {Prince},\ and\ \citenamefont {Brubaker}}]{normalising_flows_1_Kobyzev_2021}%
  \BibitemOpen
  \bibfield  {author} {\bibinfo {author} {\bibfnamefont {I.}~\bibnamefont {Kobyzev}}, \bibinfo {author} {\bibfnamefont {S.~J.}\ \bibnamefont {Prince}},\ and\ \bibinfo {author} {\bibfnamefont {M.~A.}\ \bibnamefont {Brubaker}},\ }\bibfield  {title} {\bibinfo {title} {Normalizing flows: An introduction and review of current methods},\ }\href {https://doi.org/10.1109/tpami.2020.2992934} {\bibfield  {journal} {\bibinfo  {journal} {IEEE Transactions on Pattern Analysis and Machine Intelligence}\ }\textbf {\bibinfo {volume} {43}},\ \bibinfo {pages} {3964–3979} (\bibinfo {year} {2021})}\BibitemShut {NoStop}%
\bibitem [{\citenamefont {Papamakarios}\ \emph {et~al.}(2021)\citenamefont {Papamakarios}, \citenamefont {Nalisnick}, \citenamefont {Rezende}, \citenamefont {Mohamed},\ and\ \citenamefont {Lakshminarayanan}}]{normalizing_flows_papamakarios_JMLR:v22:19-1028}%
  \BibitemOpen
  \bibfield  {author} {\bibinfo {author} {\bibfnamefont {G.}~\bibnamefont {Papamakarios}}, \bibinfo {author} {\bibfnamefont {E.}~\bibnamefont {Nalisnick}}, \bibinfo {author} {\bibfnamefont {D.~J.}\ \bibnamefont {Rezende}}, \bibinfo {author} {\bibfnamefont {S.}~\bibnamefont {Mohamed}},\ and\ \bibinfo {author} {\bibfnamefont {B.}~\bibnamefont {Lakshminarayanan}},\ }\bibfield  {title} {\bibinfo {title} {Normalizing flows for probabilistic modeling and inference},\ }\href {http://jmlr.org/papers/v22/19-1028.html} {\bibfield  {journal} {\bibinfo  {journal} {Journal of Machine Learning Research}\ }\textbf {\bibinfo {volume} {22}},\ \bibinfo {pages} {1} (\bibinfo {year} {2021})}\BibitemShut {NoStop}%
\bibitem [{\citenamefont {Winkler}\ \emph {et~al.}(2023)\citenamefont {Winkler}, \citenamefont {Worrall}, \citenamefont {Hoogeboom},\ and\ \citenamefont {Welling}}]{conditional_nf_winkler2023learninglikelihoodsconditionalnormalizing}%
  \BibitemOpen
  \bibfield  {author} {\bibinfo {author} {\bibfnamefont {C.}~\bibnamefont {Winkler}}, \bibinfo {author} {\bibfnamefont {D.}~\bibnamefont {Worrall}}, \bibinfo {author} {\bibfnamefont {E.}~\bibnamefont {Hoogeboom}},\ and\ \bibinfo {author} {\bibfnamefont {M.}~\bibnamefont {Welling}},\ }\href {https://arxiv.org/abs/1912.00042} {\bibinfo {title} {Learning likelihoods with conditional normalizing flows}} (\bibinfo {year} {2023}),\ \Eprint {https://arxiv.org/abs/1912.00042} {arXiv:1912.00042 [cs.LG]} \BibitemShut {NoStop}%
\bibitem [{\citenamefont {Müller}\ \emph {et~al.}(2019)\citenamefont {Müller}, \citenamefont {McWilliams}, \citenamefont {Rousselle}, \citenamefont {Gross},\ and\ \citenamefont {Novák}}]{linear_splines_müller2019neuralimportancesampling}%
  \BibitemOpen
  \bibfield  {author} {\bibinfo {author} {\bibfnamefont {T.}~\bibnamefont {Müller}}, \bibinfo {author} {\bibfnamefont {B.}~\bibnamefont {McWilliams}}, \bibinfo {author} {\bibfnamefont {F.}~\bibnamefont {Rousselle}}, \bibinfo {author} {\bibfnamefont {M.}~\bibnamefont {Gross}},\ and\ \bibinfo {author} {\bibfnamefont {J.}~\bibnamefont {Novák}},\ }\href {https://arxiv.org/abs/1808.03856} {\bibinfo {title} {Neural importance sampling}} (\bibinfo {year} {2019}),\ \Eprint {https://arxiv.org/abs/1808.03856} {arXiv:1808.03856 [cs.LG]} \BibitemShut {NoStop}%
\bibitem [{\citenamefont {Belghazi}\ \emph {et~al.}(2018)\citenamefont {Belghazi} \emph {et~al.}}]{belghazi2018mutual}%
  \BibitemOpen
  \bibfield  {author} {\bibinfo {author} {\bibfnamefont {M.~I.}\ \bibnamefont {Belghazi}} \emph {et~al.},\ }\bibfield  {title} {\bibinfo {title} {Mutual information neural estimation},\ }in\ \href {https://proceedings.mlr.press/v80/belghazi18a.html} {\emph {\bibinfo {booktitle} {Proceedings of the 35th International Conference on Machine Learning}}},\ \bibinfo {series} {Proceedings of Machine Learning Research}, Vol.~\bibinfo {volume} {80},\ \bibinfo {editor} {edited by\ \bibinfo {editor} {\bibfnamefont {J.}~\bibnamefont {Dy}}\ and\ \bibinfo {editor} {\bibfnamefont {A.}~\bibnamefont {Krause}}}\ (\bibinfo  {publisher} {PMLR},\ \bibinfo {year} {2018})\ pp.\ \bibinfo {pages} {531--540}\BibitemShut {NoStop}%
\bibitem [{\citenamefont {Hjelm}\ \emph {et~al.}(2019)\citenamefont {Hjelm}, \citenamefont {Fedorov}, \citenamefont {Lavoie-Marchildon}, \citenamefont {Grewal}, \citenamefont {Bachman}, \citenamefont {Trischler},\ and\ \citenamefont {Bengio}}]{hjelm2018learning}%
  \BibitemOpen
  \bibfield  {author} {\bibinfo {author} {\bibfnamefont {R.~D.}\ \bibnamefont {Hjelm}}, \bibinfo {author} {\bibfnamefont {A.}~\bibnamefont {Fedorov}}, \bibinfo {author} {\bibfnamefont {S.}~\bibnamefont {Lavoie-Marchildon}}, \bibinfo {author} {\bibfnamefont {K.}~\bibnamefont {Grewal}}, \bibinfo {author} {\bibfnamefont {P.}~\bibnamefont {Bachman}}, \bibinfo {author} {\bibfnamefont {A.}~\bibnamefont {Trischler}},\ and\ \bibinfo {author} {\bibfnamefont {Y.}~\bibnamefont {Bengio}},\ }\bibfield  {title} {\bibinfo {title} {Learning deep representations by mutual information estimation and maximization},\ }in\ \href@noop {} {\emph {\bibinfo {booktitle} {International Conference on Learning Representations}}}\ (\bibinfo {year} {2019})\BibitemShut {NoStop}%
\bibitem [{\citenamefont {Zhuang}\ \emph {et~al.}(2019)\citenamefont {Zhuang} \emph {et~al.}}]{transfer_learning_DBLP:journals/corr/abs-1911-02685}%
  \BibitemOpen
  \bibfield  {author} {\bibinfo {author} {\bibfnamefont {F.}~\bibnamefont {Zhuang}} \emph {et~al.},\ }\bibfield  {title} {\bibinfo {title} {A comprehensive survey on transfer learning},\ }\href {http://arxiv.org/abs/1911.02685} {\bibfield  {journal} {\bibinfo  {journal} {CoRR}\ }\textbf {\bibinfo {volume} {abs/1911.02685}} (\bibinfo {year} {2019})},\ \Eprint {https://arxiv.org/abs/1911.02685} {1911.02685} \BibitemShut {NoStop}%
\bibitem [{\citenamefont {Mokhtar}\ \emph {et~al.}(2025)\citenamefont {Mokhtar}, \citenamefont {Pata}, \citenamefont {Kagan}, \citenamefont {Garcia}, \citenamefont {Wulff}, \citenamefont {Zhang},\ and\ \citenamefont {Duarte}}]{mokhtar2025finetuningmachinelearnedparticleflowreconstruction}%
  \BibitemOpen
  \bibfield  {author} {\bibinfo {author} {\bibfnamefont {F.}~\bibnamefont {Mokhtar}}, \bibinfo {author} {\bibfnamefont {J.}~\bibnamefont {Pata}}, \bibinfo {author} {\bibfnamefont {M.}~\bibnamefont {Kagan}}, \bibinfo {author} {\bibfnamefont {D.}~\bibnamefont {Garcia}}, \bibinfo {author} {\bibfnamefont {E.}~\bibnamefont {Wulff}}, \bibinfo {author} {\bibfnamefont {M.}~\bibnamefont {Zhang}},\ and\ \bibinfo {author} {\bibfnamefont {J.}~\bibnamefont {Duarte}},\ }\href {https://arxiv.org/abs/2503.00131} {\bibinfo {title} {Fine-tuning machine-learned particle-flow reconstruction for new detector geometries in future colliders}} (\bibinfo {year} {2025}),\ \Eprint {https://arxiv.org/abs/2503.00131} {arXiv:2503.00131 [hep-ex]} \BibitemShut {NoStop}%
\bibitem [{\citenamefont {de~Fatis}\ and\ \citenamefont {behalf of~the CMS~Collaboration)}(2012)}]{cms_pwo_scint_constraint_de_Fatis_2012}%
  \BibitemOpen
  \bibfield  {author} {\bibinfo {author} {\bibfnamefont {T.~T.}\ \bibnamefont {de~Fatis}}\ and\ \bibinfo {author} {\bibfnamefont {O.}~\bibnamefont {behalf of~the CMS~Collaboration)}},\ }\bibfield  {title} {\bibinfo {title} {Role of the cms electromagnetic calorimeter in the hunt for the higgs boson in the two-gamma channel},\ }\href {https://doi.org/10.1088/1742-6596/404/1/012002} {\bibfield  {journal} {\bibinfo  {journal} {Journal of Physics: Conference Series}\ }\textbf {\bibinfo {volume} {404}},\ \bibinfo {pages} {012002} (\bibinfo {year} {2012})}\BibitemShut {NoStop}%
\bibitem [{\citenamefont {Clevert}\ \emph {et~al.}(2016)\citenamefont {Clevert}, \citenamefont {Unterthiner},\ and\ \citenamefont {Hochreiter}}]{elu_clevert2016fastaccuratedeepnetwork}%
  \BibitemOpen
  \bibfield  {author} {\bibinfo {author} {\bibfnamefont {D.-A.}\ \bibnamefont {Clevert}}, \bibinfo {author} {\bibfnamefont {T.}~\bibnamefont {Unterthiner}},\ and\ \bibinfo {author} {\bibfnamefont {S.}~\bibnamefont {Hochreiter}},\ }\href {https://arxiv.org/abs/1511.07289} {\bibinfo {title} {Fast and accurate deep network learning by exponential linear units (elus)}} (\bibinfo {year} {2016}),\ \Eprint {https://arxiv.org/abs/1511.07289} {arXiv:1511.07289 [cs.LG]} \BibitemShut {NoStop}%
\bibitem [{\citenamefont {Kingma}\ and\ \citenamefont {Ba}(2017)}]{adam_kingma2017adammethodstochasticoptimization}%
  \BibitemOpen
  \bibfield  {author} {\bibinfo {author} {\bibfnamefont {D.~P.}\ \bibnamefont {Kingma}}\ and\ \bibinfo {author} {\bibfnamefont {J.}~\bibnamefont {Ba}},\ }\href {https://arxiv.org/abs/1412.6980} {\bibinfo {title} {Adam: A method for stochastic optimization}} (\bibinfo {year} {2017}),\ \Eprint {https://arxiv.org/abs/1412.6980} {arXiv:1412.6980 [cs.LG]} \BibitemShut {NoStop}%
\end{thebibliography}%

\medskip

\newpage

\appendix

\section{One-Dimensional flow vs Diffusion}
\label{appendix:1d_vs_diff}

\begin{figure*}
    \begin{subfigure}{0.41\textwidth}
        \centering
        \includegraphics[width=\textwidth]{fig/evolution_avg_runs3_layers1_evts700_en20_ini10_tl_reco.png}
            \caption{One-dimensional flow model evolution, 1 layer}
            \label{fig:1dflow_comparison_1pair}
    \end{subfigure}%
    \hfilplus
    \begin{subfigure}{0.41\textwidth}
        \centering
        \includegraphics[width=\textwidth]{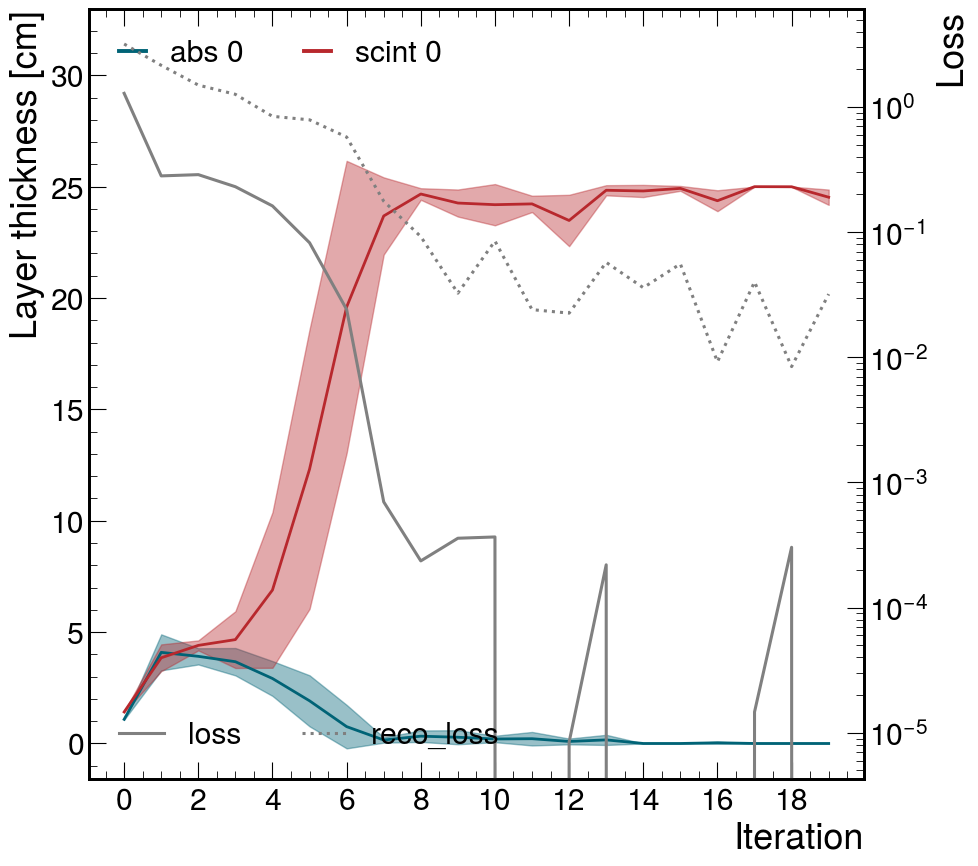}
            \caption{Diffusion model evolution, 1 layer}
            \label{fig:diffusion_comparison_1pair}
    \end{subfigure}
    \begin{subfigure}{0.41\textwidth}
        \centering
        \includegraphics[width=0.99\textwidth]{fig/evolution_avg_runs3_layers3_evts700_en20_ini10_tl_reco.png}
            \caption{One-dimensional flow model evolution, 3 layers}
            \label{fig:1dflow_comparison_3pair}
    \end{subfigure}%
    \hfilplus
    \begin{subfigure}{0.41\textwidth}
        \centering
        \includegraphics[width=\textwidth]{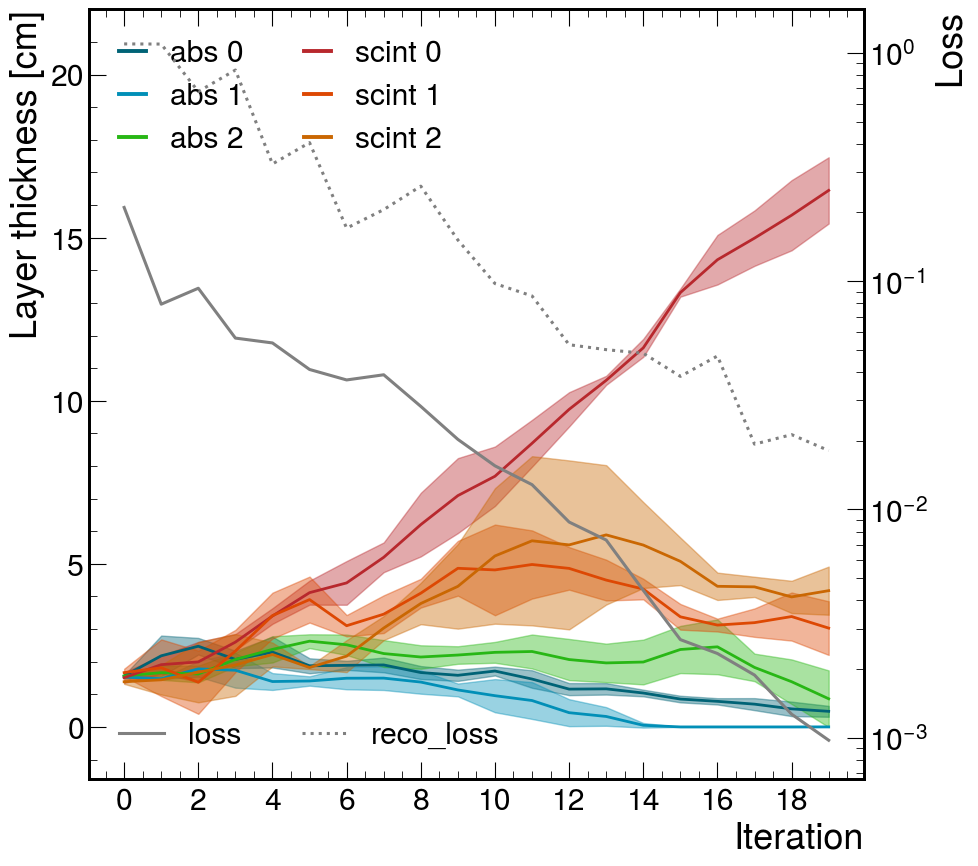}
            \caption{Diffusion model evolution, 3 layers}
            \label{fig:diffusion_comparison_3pair}
    \end{subfigure}
    \caption{Comparison of one-dimensional flow and diffusion model for one and three layers with 700 events and photons between \SI{1} and \SI{20}{GeV}.}
    \label{fig:1dflow_vs_diff_1pair}
\end{figure*}

While the diffusion models generally in this case have lower run-to-run variation, the \ac{1D NF} provides gains in terms of runtime due to significantly simpler architecture and training procedure. For a comparison of the evolution of a single layer optimization see Figure~\ref{fig:1dflow_comparison_1pair}. 

\section{Summation of Layer Thickness}
\label{appendix:summation}

\begin{figure*}
\centering
	\begin{subfigure}{0.41\textwidth}
	    \centering
		      \includegraphics[width=0.98\textwidth]{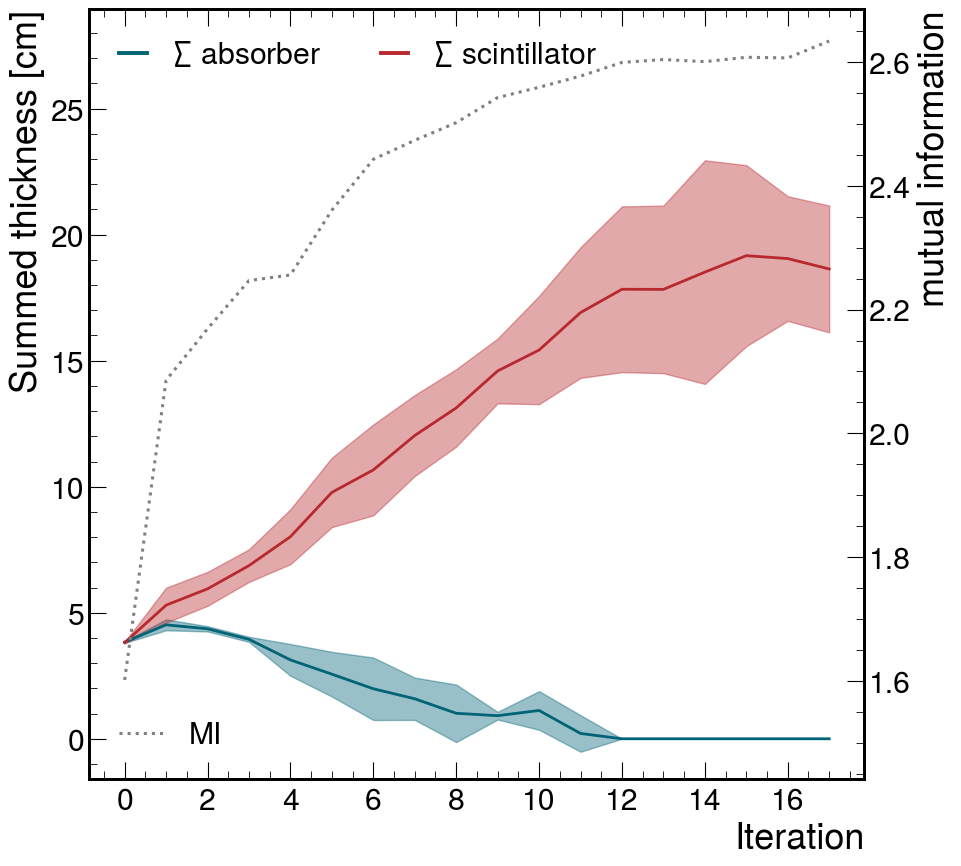}
            \caption{Sum 2 layers, MI-OPT}
            \label{fig:summed_reco_layer2_mi}
	\end{subfigure}%
	\hfilplus
	\begin{subfigure}{0.41\textwidth}
	    \centering
		      \includegraphics[width=\textwidth]{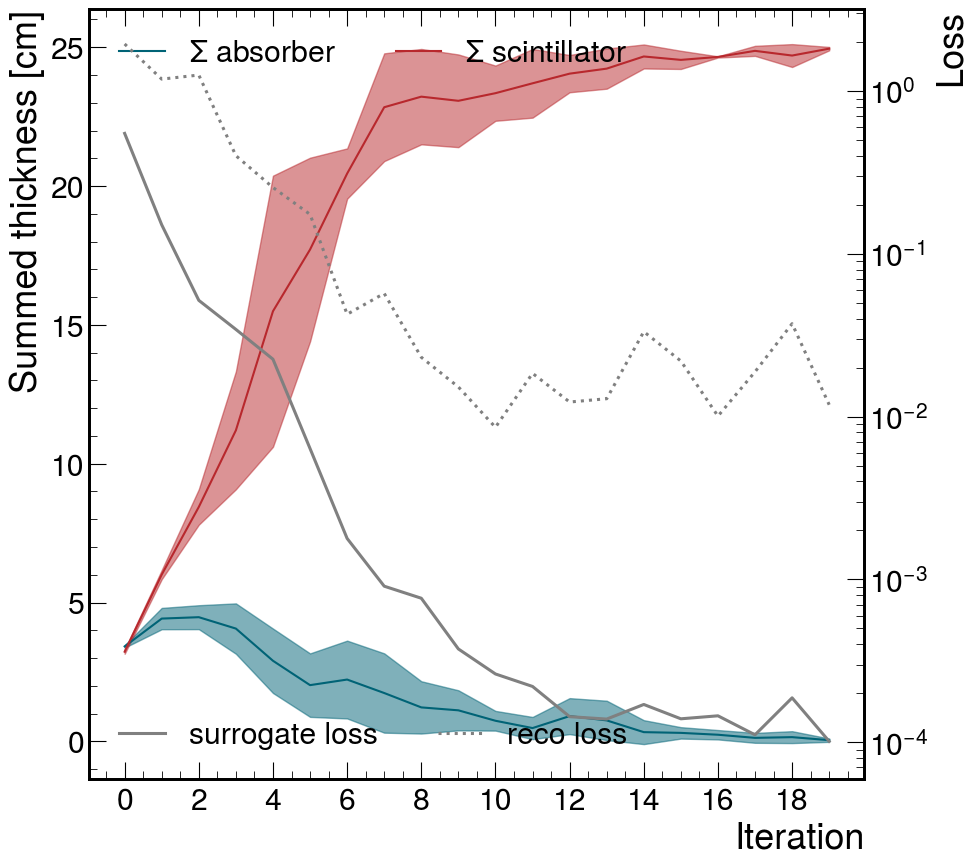}
            \caption{Sum 2 layers, RECO-OPT}
            \label{fig:summed_reco_layer2_reco}
	\end{subfigure}
    \begin{subfigure}{0.41\textwidth}
	    \centering
		      \includegraphics[width=0.98\textwidth]{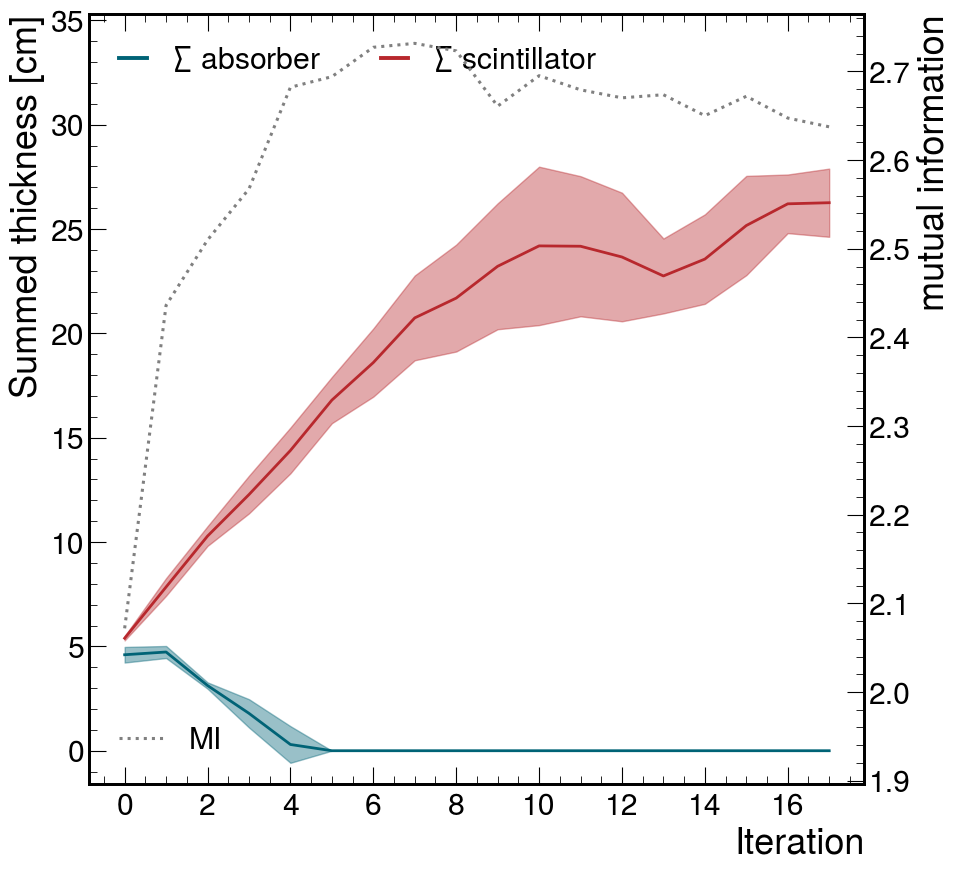}
            \caption{Sum 3 layers, MI-OPT}
            \label{fig:summed_reco_layer3_mi}
	\end{subfigure}%
	\hfilplus
    \begin{subfigure}{0.41\textwidth}
	    \centering
		      \includegraphics[width=\textwidth]{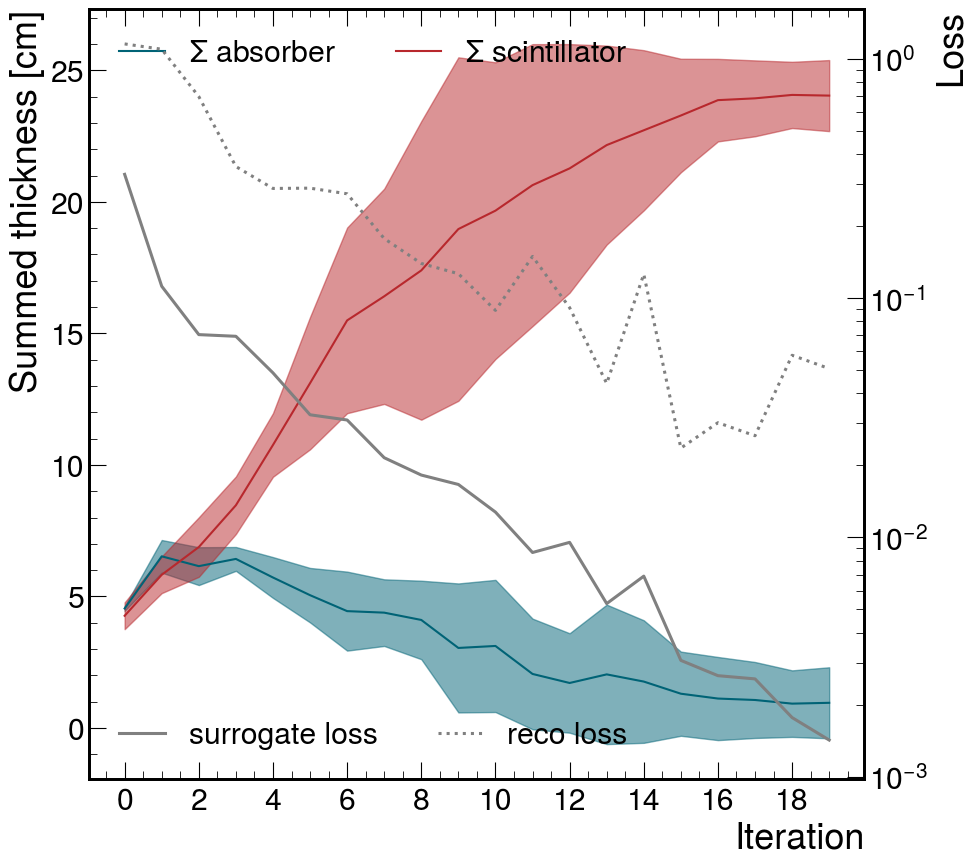}
            \caption{Sum 3 layers, RECO-OPT}
            \label{fig:summed_reco_layer3_reco}
	\end{subfigure}
\caption{Summation of layer thickness evolution for MI-OPT (left) and RECO-OPT (right), averaged over three runs for the 2 layer (top row), and 3 layer (bottom row) calorimeters. All runs use 700 events with incident energy ranging between \qtyrange[range-units=single,range-phrase=-]{1}{20}{\GeV}.
\label{fig:evo_summation}}
\end{figure*}

When summing total scintillator and absorber thickness we see that generally the total scintillator thickness approaches the constraint of \SI{25}{cm} while the total absorber thickness approaches \SI{0}{cm}, shown in Figure~\ref{fig:evo_summation}. 

\section{Failure Cases without Transfer Learning}
\label{appendix:notl_failure}

\begin{figure*}
\centering
    \begin{subfigure}{0.3\textwidth}
        \centering
            \includegraphics[width=\textwidth]{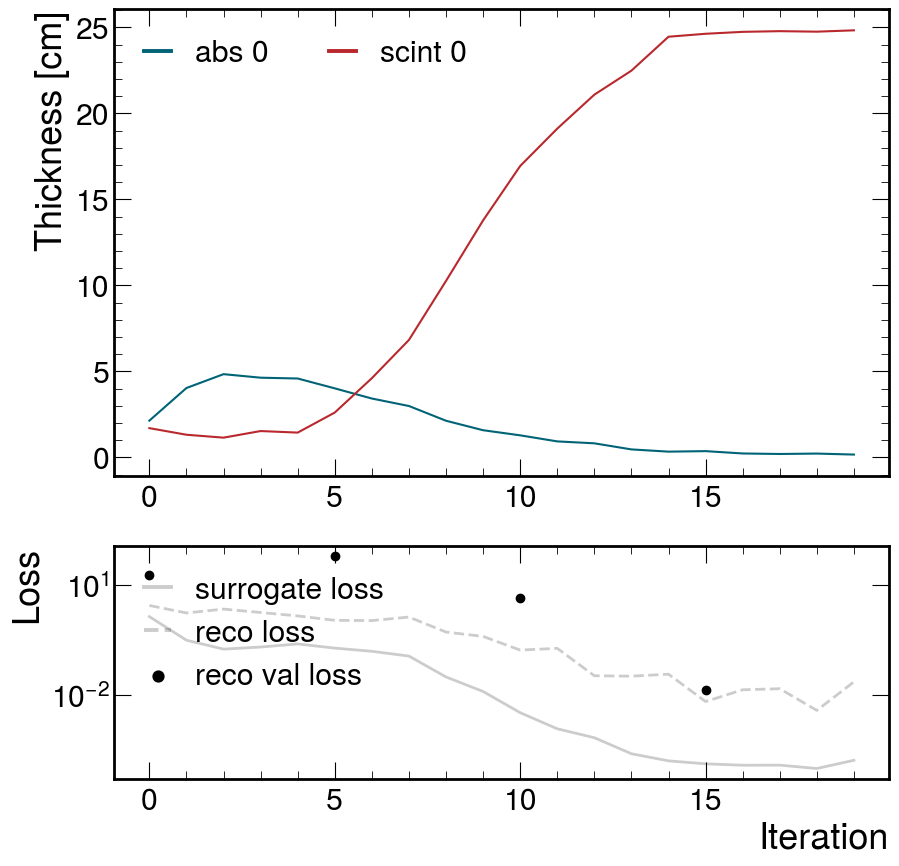}
            \caption{700 events, run 0}
            \label{fig:tl_failure_700evts_run0}
    \end{subfigure}
    \hfill
    \begin{subfigure}{0.3\textwidth}
        \centering
            \includegraphics[width=\textwidth]{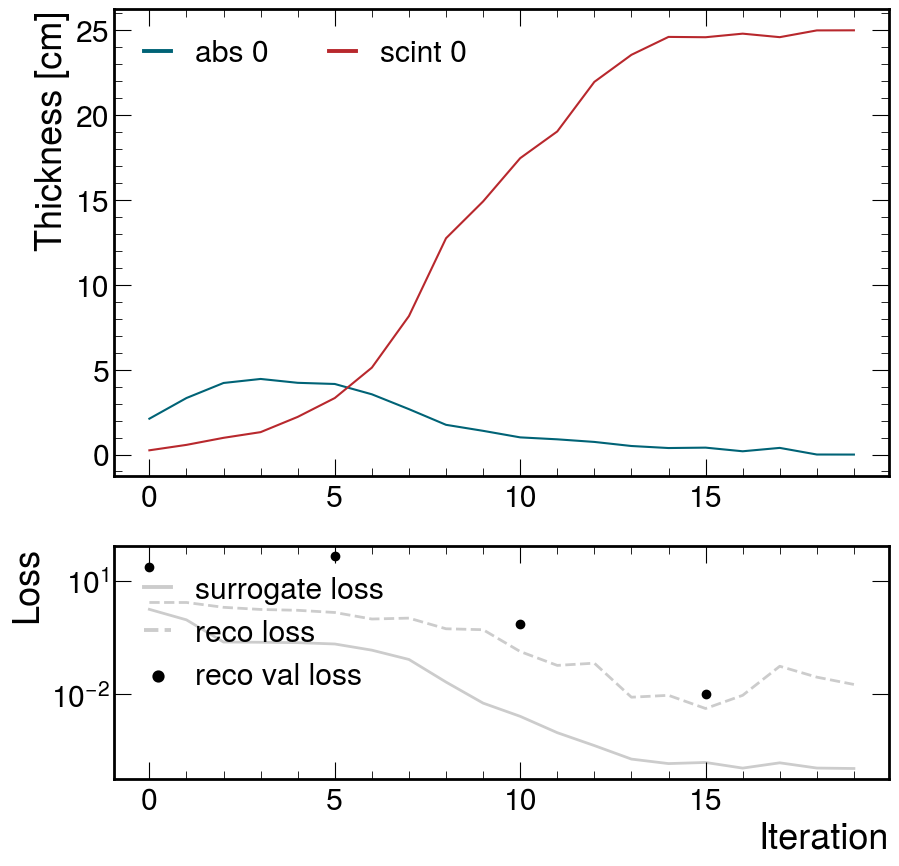}
            \caption{700 events, run 1}
            \label{fig:tl_failure_700evts_run1}
    \end{subfigure}
    \hfill
    \begin{subfigure}{0.3\textwidth}
        \centering
            \includegraphics[width=\textwidth]{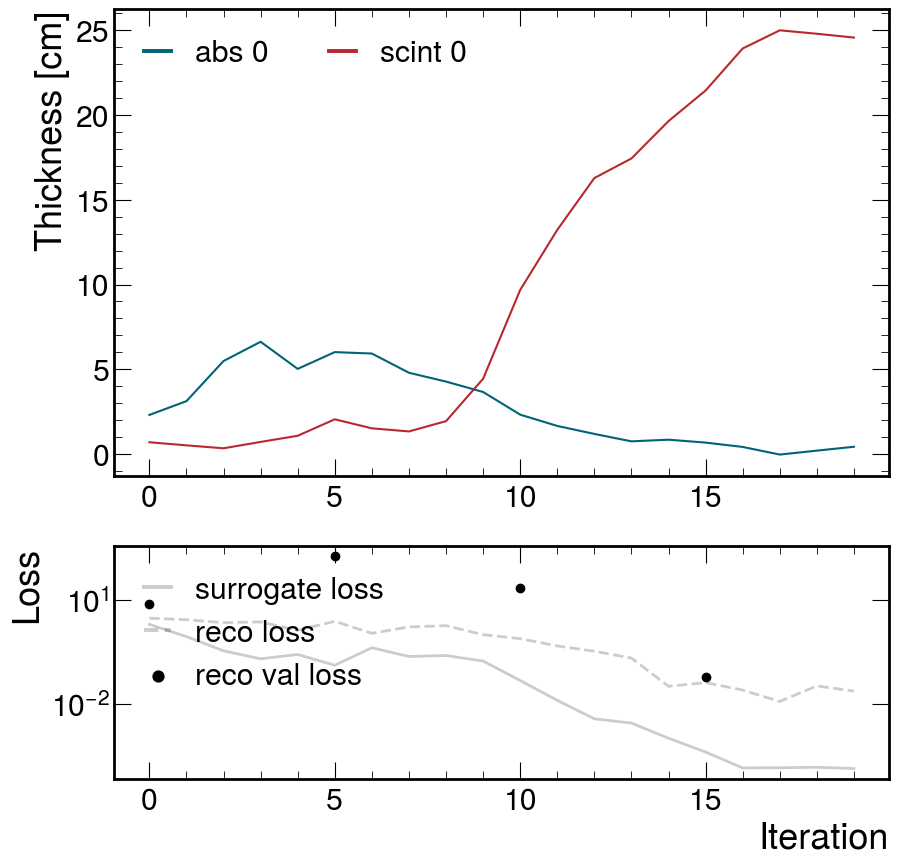}
            \caption{700 events, run 2}
            \label{fig:tl_failure_700evts_run2}
    \end{subfigure}
    \hfill
        \begin{subfigure}{0.3\textwidth}
        \centering
            \includegraphics[width=\textwidth]{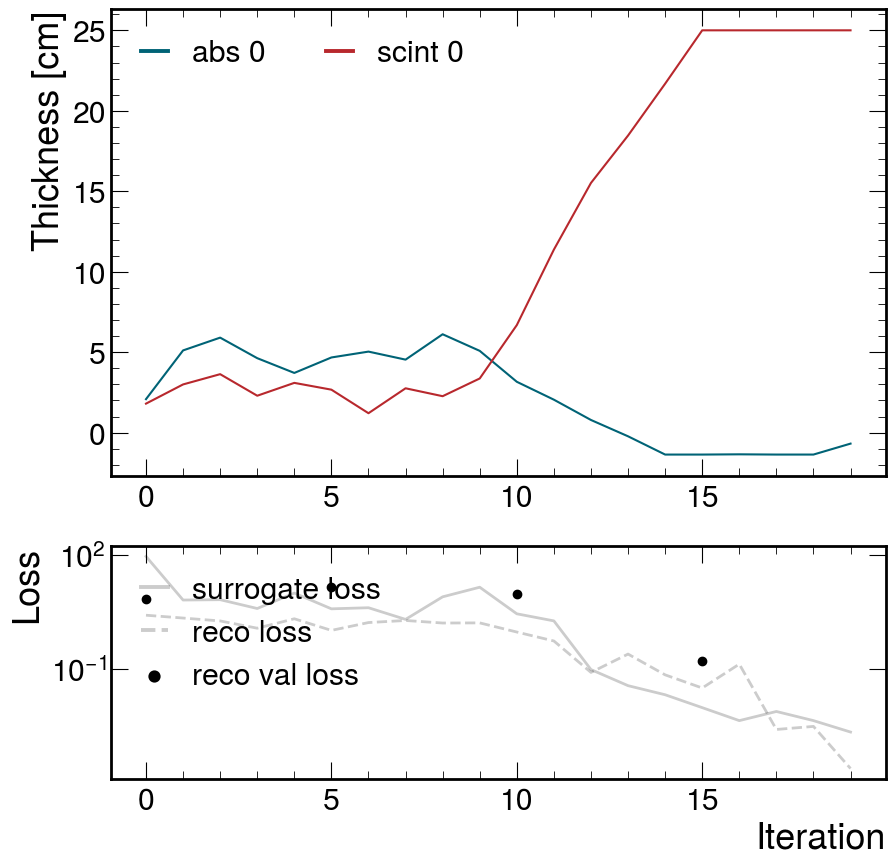}
            \caption{50 events, run 0}
            \label{fig:tl_failure_50evts_run0}
    \end{subfigure}
    \hfill
    \begin{subfigure}{0.3\textwidth}
        \centering
            \includegraphics[width=\textwidth]{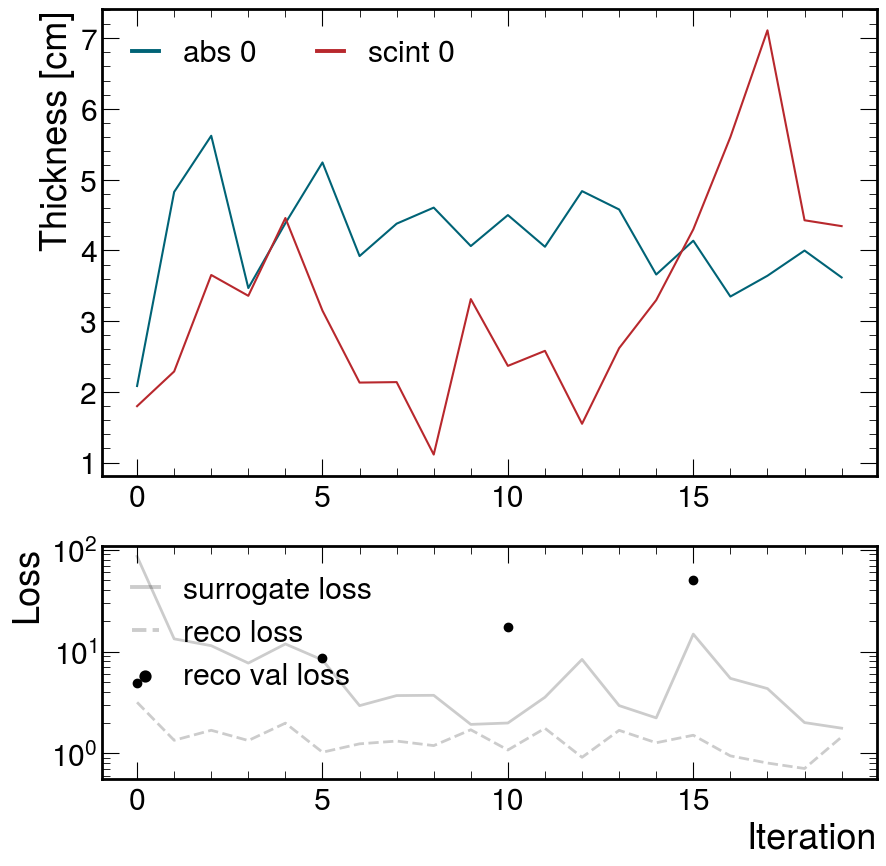}
            \caption{50 events, run 1}
            \label{fig:tl_failure_50evts_run1}
    \end{subfigure}
    \hfill
    \begin{subfigure}{0.3\textwidth}
        \centering
            \includegraphics[width=\textwidth]{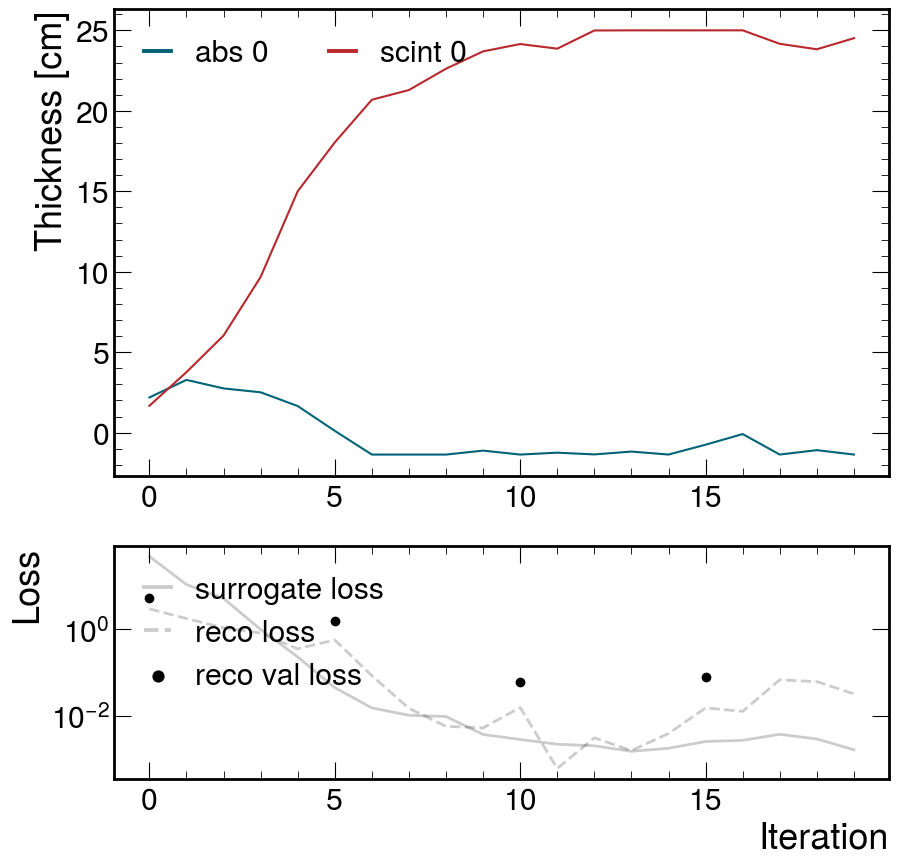}
            \caption{50 events, run 2}
            \label{fig:tl_failure_50evts_run2}
    \end{subfigure}
    \hfill
    \begin{subfigure}{0.3\textwidth}
        \centering
            \includegraphics[width=\textwidth]{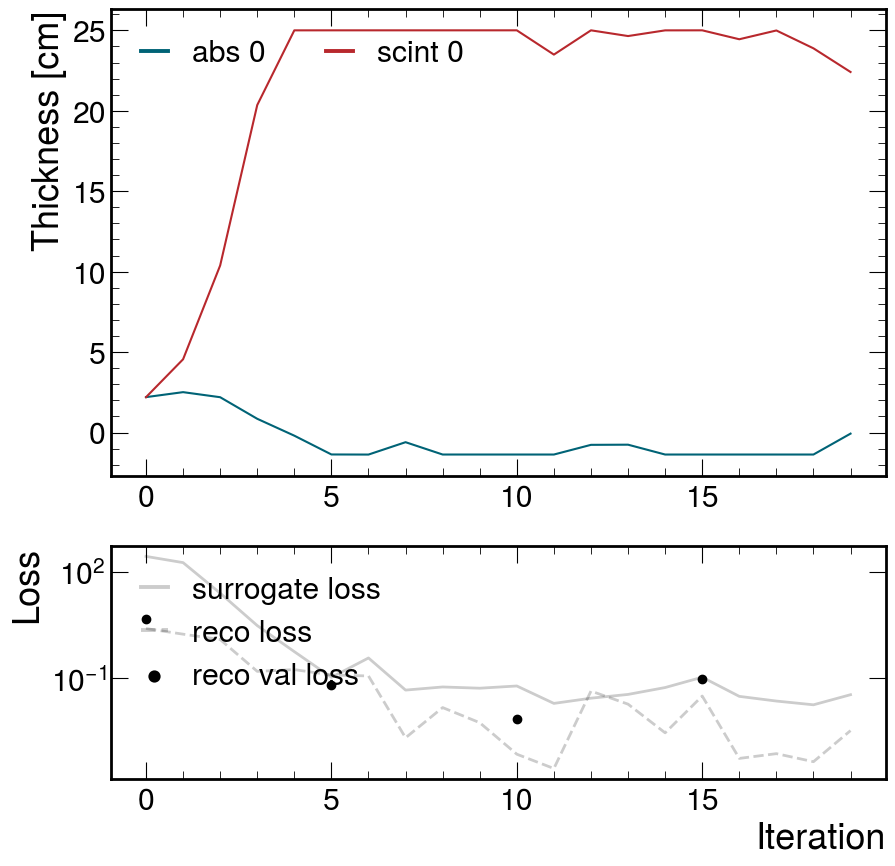}
            \caption{5 events, run 0}
            \label{fig:tl_failure_5evts_run0}
    \end{subfigure}
    \hfill
    \begin{subfigure}{0.3\textwidth}
        \centering
            \includegraphics[width=\textwidth]{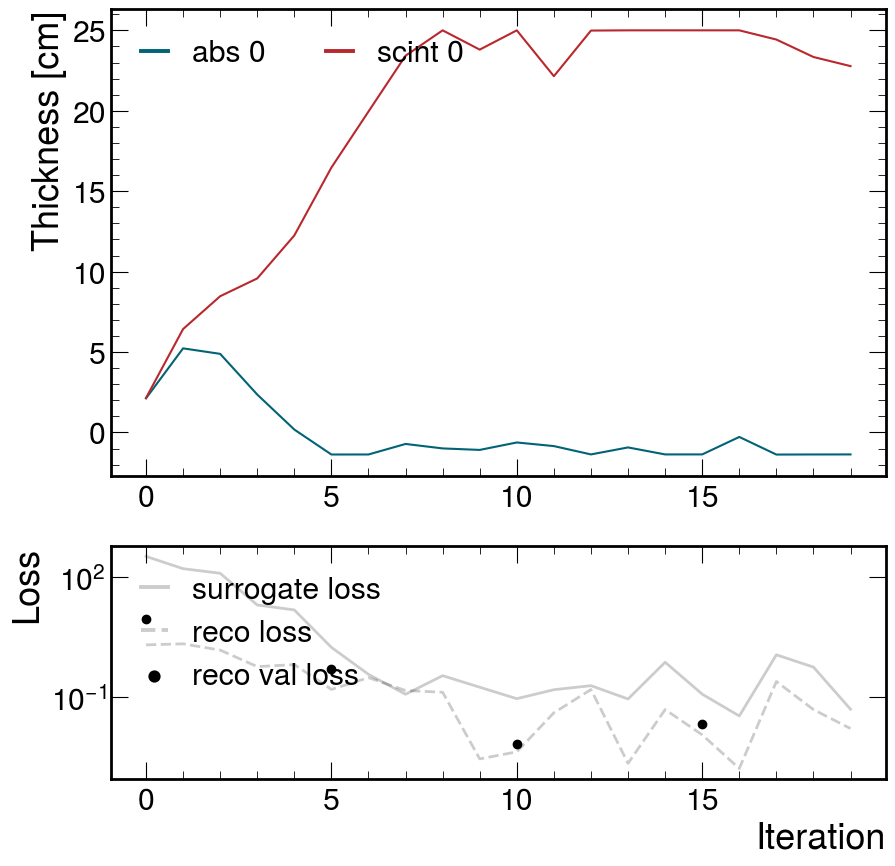}
            \caption{5 events, run 1}
            \label{fig:tl_failure_5evts_run1}
    \end{subfigure}
    \hfill
    \begin{subfigure}{0.3\textwidth}
        \centering
            \includegraphics[width=\textwidth]{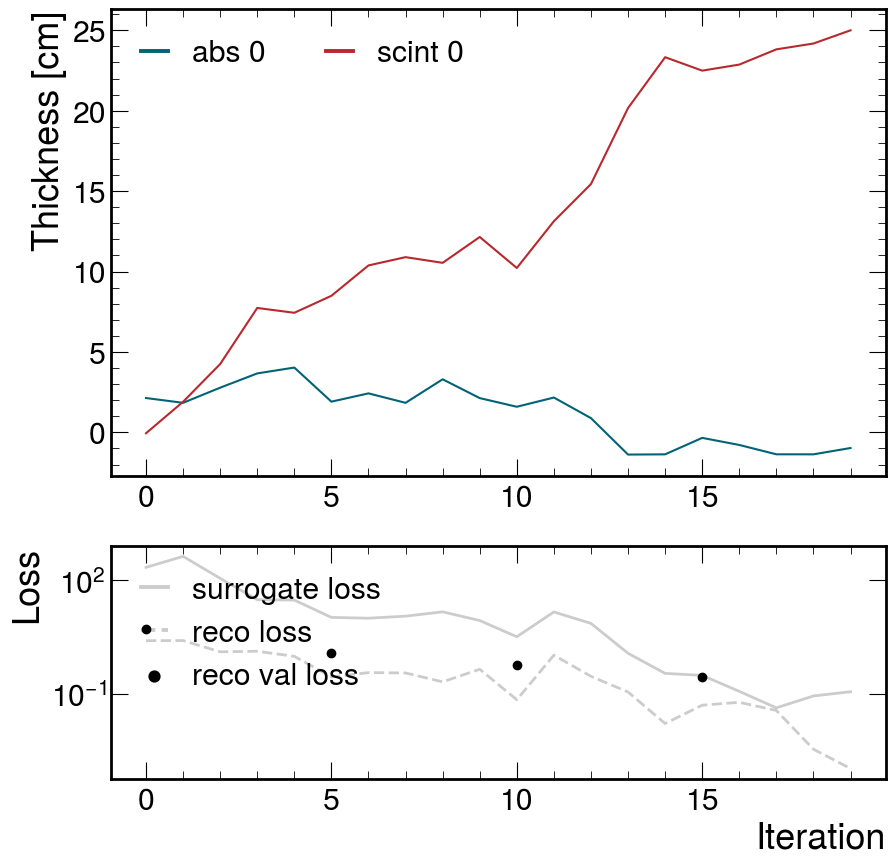}
            \caption{5 events, run 2}
            \label{fig:tl_failure_5evts_run2}
    \end{subfigure}
    \hfill
\caption{RECO-OPT: Individual runs with \ac{TL}. Top row shows 700 events, middle shows 50 events, bottom shows 5 events.}
\label{fig:tl_failure_single_runs}
\end{figure*}

\begin{figure*}
\centering
    \begin{subfigure}{0.3\textwidth}
        \centering
            \includegraphics[width=\textwidth]{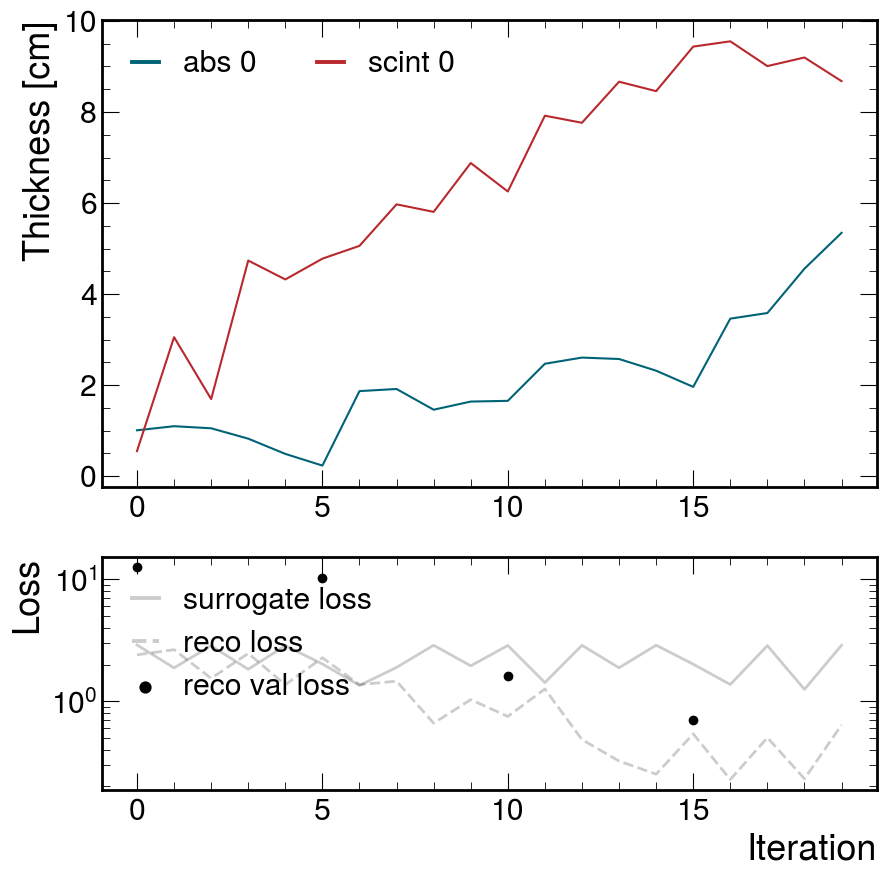}
            \caption{700 events, run 0}
            \label{fig:notl_failure_700evts_run0}
    \end{subfigure}
    \hfill
    \begin{subfigure}{0.3\textwidth}
        \centering
            \includegraphics[width=\textwidth]{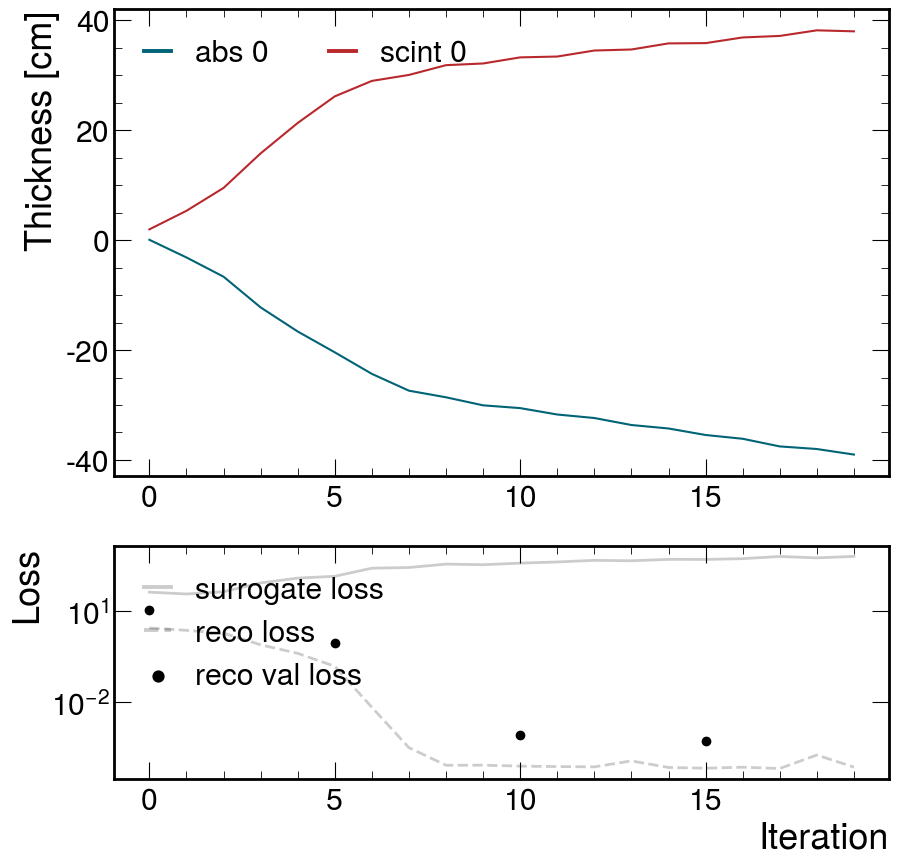}
            \caption{700 events, run 1}
            \label{fig:notl_failure_700evts_run1}
    \end{subfigure}
    \hfill
    \begin{subfigure}{0.3\textwidth}
        \centering
            \includegraphics[width=\textwidth]{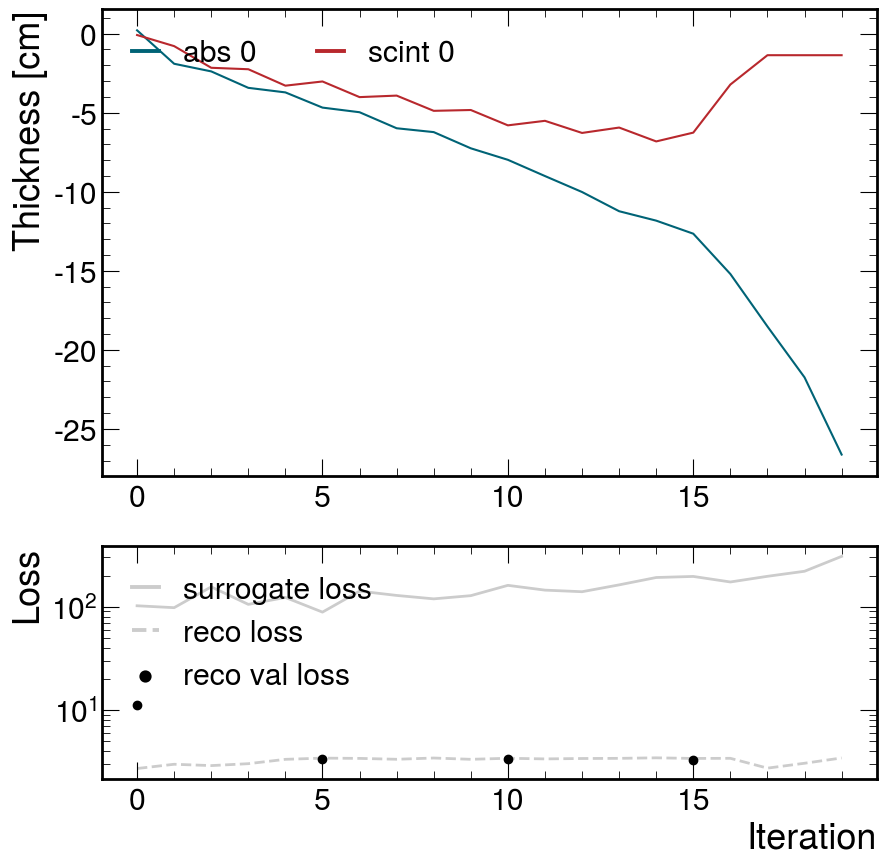}
            \caption{700 events, run 2}
            \label{fig:notl_failure_700evts_run2}
    \end{subfigure}
    \hfill
        \begin{subfigure}{0.3\textwidth}
        \centering
            \includegraphics[width=\textwidth]{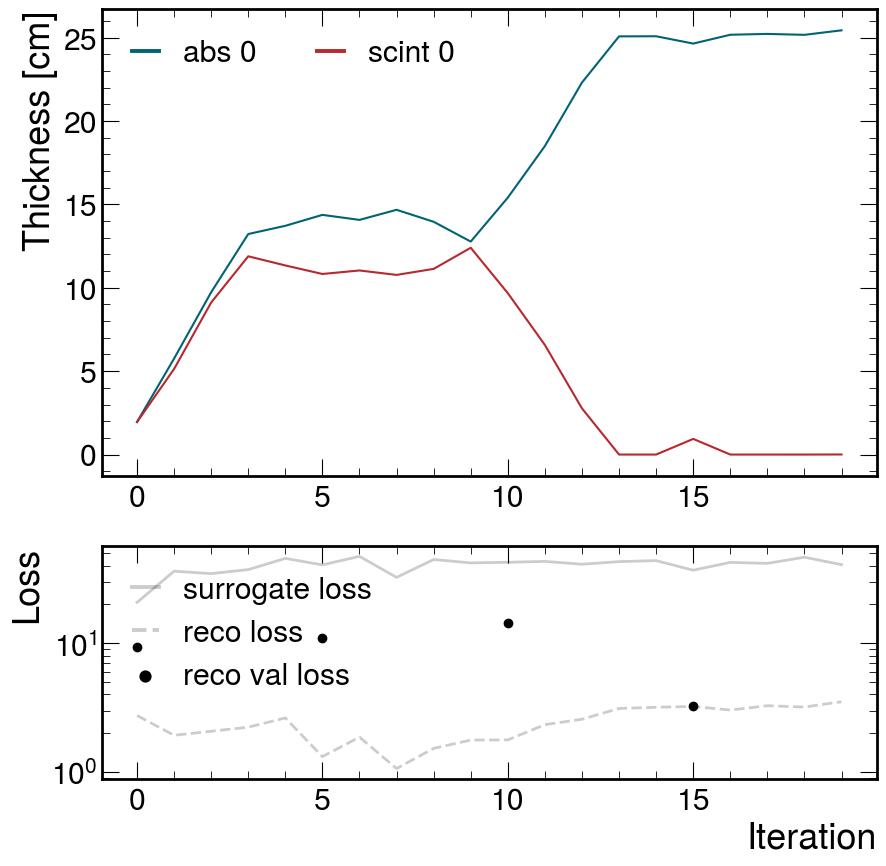}
            \caption{50 events, run 0}
            \label{fig:notl_failure_50evts_run0}
    \end{subfigure}
    \hfill
    \begin{subfigure}{0.3\textwidth}
        \centering
            \includegraphics[width=\textwidth]{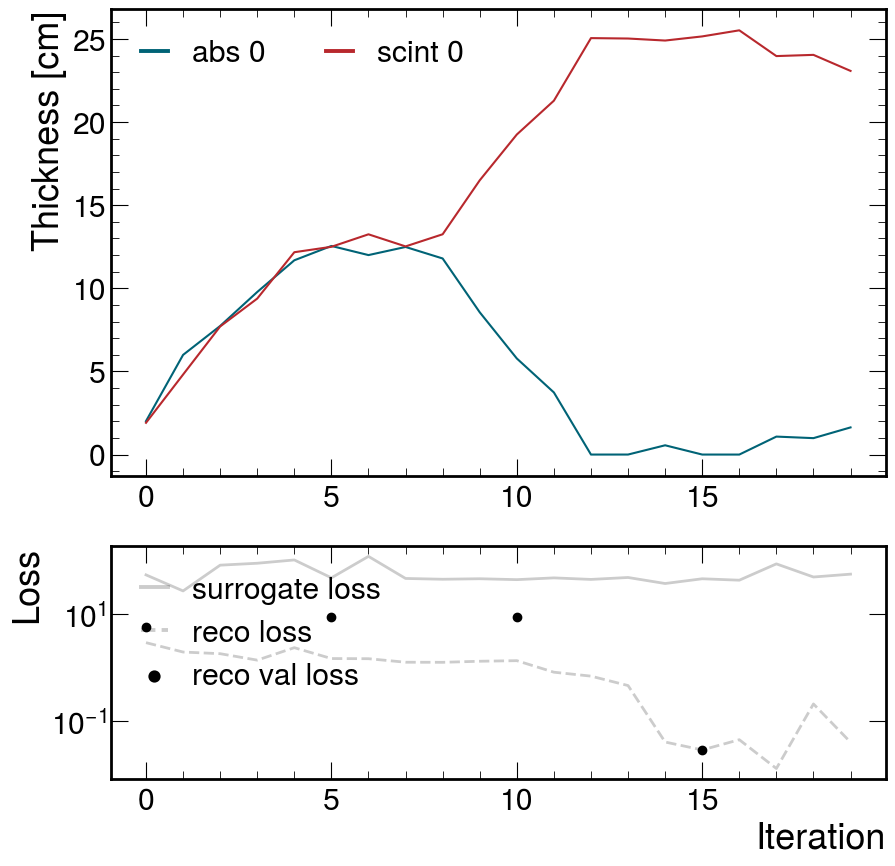}
            \caption{50 events, run 1}
            \label{fig:notl_failure_50evts_run1}
    \end{subfigure}
    \hfill
    \begin{subfigure}{0.3\textwidth}
        \centering
            \includegraphics[width=\textwidth]{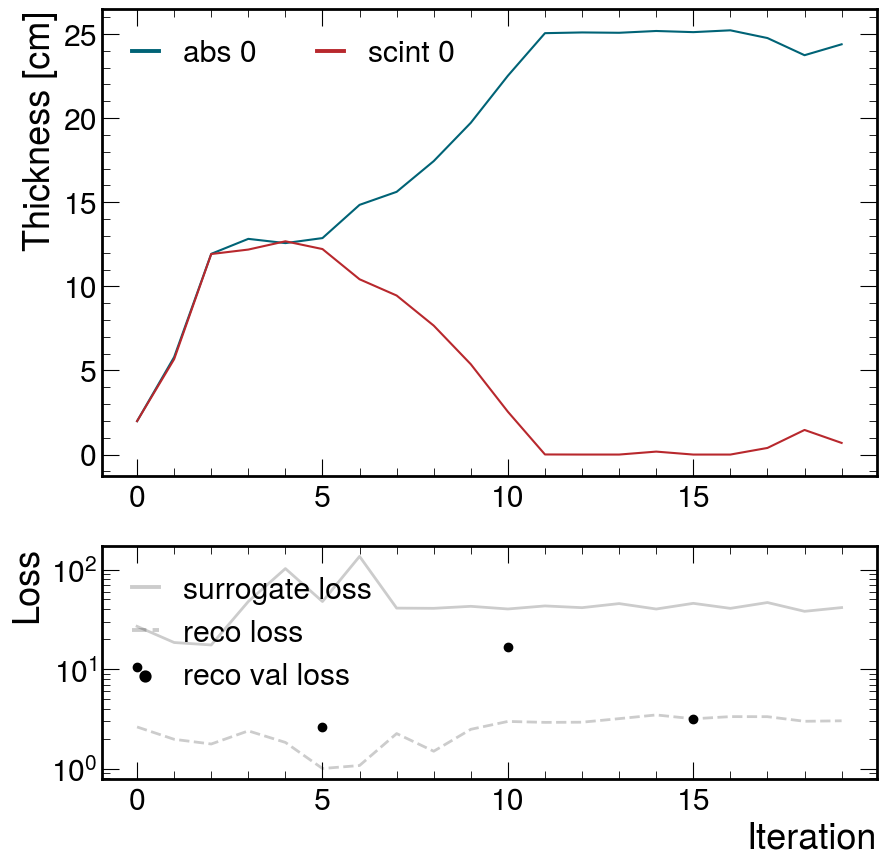}
            \caption{50 events, run 2}
            \label{fig:notl_failure_50evts_run2}
    \end{subfigure}
    \hfill
    \begin{subfigure}{0.3\textwidth}
        \centering
            \includegraphics[width=\textwidth]{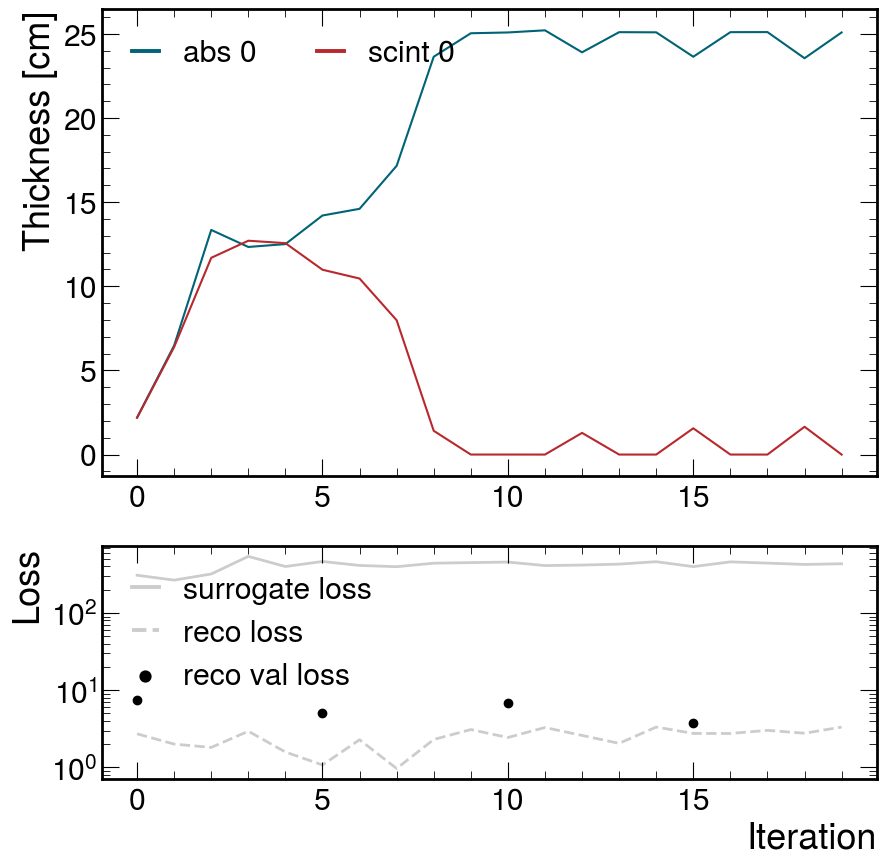}
            \caption{5 events, run 0}
            \label{fig:notl_failure_5evts_run0}
    \end{subfigure}
    \hfill
    \begin{subfigure}{0.3\textwidth}
        \centering
            \includegraphics[width=\textwidth]{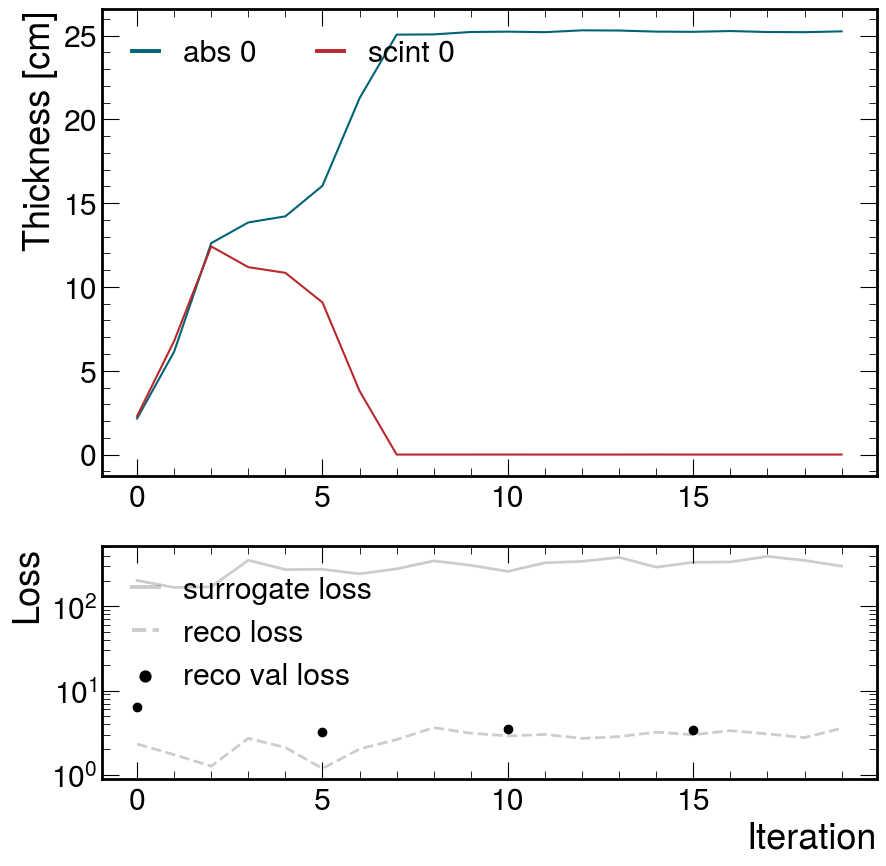}
            \caption{5 events, run 1}
            \label{fig:notl_failure_5evts_run1}
    \end{subfigure}
    \hfill
    \begin{subfigure}{0.3\textwidth}
        \centering
            \includegraphics[width=\textwidth]{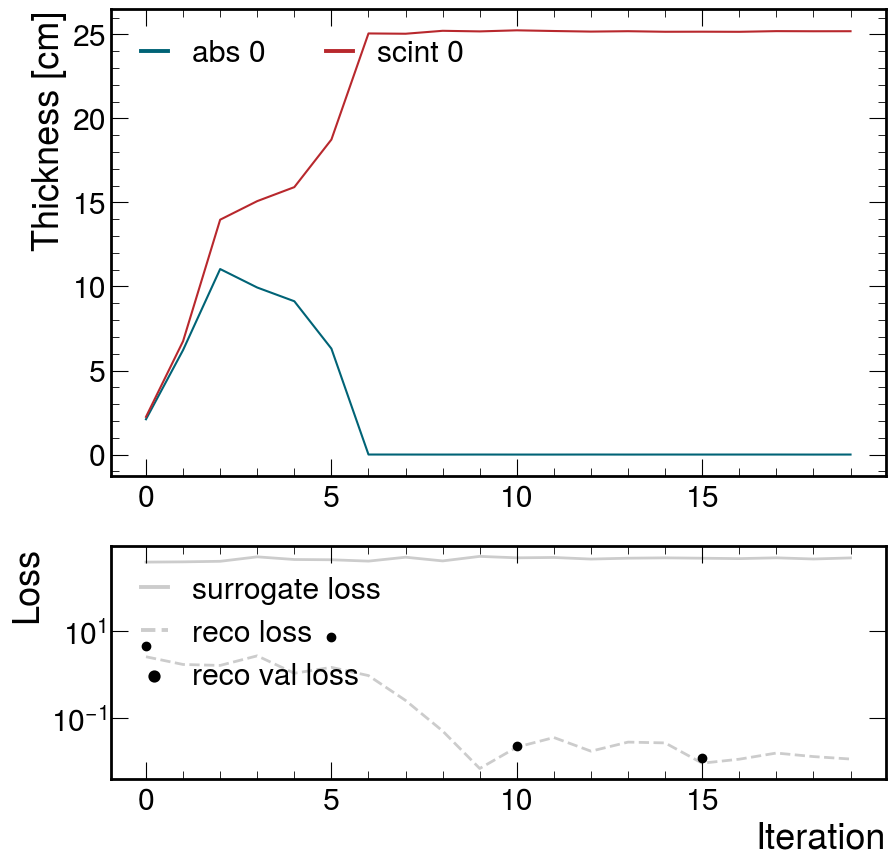}
            \caption{5 events, run 2}
            \label{fig:notl_failure_5evts_run2}
    \end{subfigure}
    \hfill
\caption{RECO-OPT: Individual runs without \ac{TL}. Top row shows 700 events, middle shows 50 events, bottom shows 5 events.}
\label{fig:notl_failure_single_runs}
\end{figure*}

For RECO-OPT, the use of \ac{TL} seems to be essential for reasonable and consistent performance. Without \ac{TL}, even for relatively large event numbers, a breakdown in the optimization process is observed. Individual runs fail in distinct ways, which can make the run-averaged plots for these cases difficult to interpret. Here the evolutions of the individual runs are presented for reference. Individual runs of cases using \ac{TL} are displayed in Figure \ref{fig:tl_failure_single_runs} and without \ac{TL} in Figure \ref{fig:notl_failure_single_runs}. 

Using \ac{TL}, the evolutions regardless of the number of simulated events are very similar, all beginning with some small increase in absorber thickness before beginning to increase scintillator thickness.

Without \ac{TL}, the behaviour is quite different. For 700 events, we see very inconsistent behaviour, often achieving large and nonphysical values. When reducing to 50 and 5 events, we see a repeated pattern of failure where both segments are increased uniformly until $\sim$ \SI{13}{cm}, when either the scintillator or absorber will continue to increase or begin to decrease. These failure cases are not well understood but are clearly linked to the lack of \ac{TL}.

\section{Experiment and Network Hyperparameters}
\label{appendix:experiment_network_hyper}

The hyperparameters used by the reconstruction and surrogate models of each framework are presented in Table~\ref{tab:network_hyper}. The networks in RECO-OPT go through different training stages with reduced learning rates (LR). The reconstruction network is trained for 200 epochs with LR $4 \times 10^{-4}$ and then 200 epochs at $1 \times 10^{-5}$, while the surrogate network starts with $3 \times 10^{-4}$, then $1 \times 10^{-4}$, and finally $1 \times 10^{-5}$, each for 200 epochs.
The \ac{1D NF} uses an MLP to predict the piecewise linear parameters based on the conditional parameters. The number of MLP layers and activation function listed in the table for RECO-OPT Surr are the hyperparameters of this MLP.

\begin{table*}[t]
    \centering
    \renewcommand{\arraystretch}{1.2}
    \begin{tabular}{lccccc}
         \toprule
         \multicolumn{6}{c}{\textbf{\textcolor{gray}{Hyperparameters}}} \\
         \textbf{Networks} & N MLP Layers & Activation & LR & Epochs & Optimizer \\
         \midrule
         \textbf{RECO-OPT Reco} & \multirow{2}{*}{3} & \multirow{4}{*}{ELU \cite{elu_clevert2016fastaccuratedeepnetwork}} & $4 \times 10^{-4}$ / $1 \times 10^{-5}$ & 400 & \multirow{4}{*}{ADAM \cite{adam_kingma2017adammethodstochasticoptimization}} \\
         \textbf{RECO-OPT Surr} & & & $3 \cdot 10^{-4}$ / $1 \times 10^{-4}$ / $1 \times 10^{-5}$ & 600 & \\
         \textbf{MI-OPT Reco} & encoder + decoder: 3 + MLP: 2 & & scheduler (init $1\cdot 10^{-2}$) & 2000 & \\
         \textbf{MI-OPT Surr} & 4 & & scheduler (init $5\cdot 10^{-2}$) & 150 & \\
        \bottomrule
    \end{tabular}
    \caption{Network hyperparameters}
    \label{tab:network_hyper}
\end{table*}

\end{document}